\documentclass[journal]{IEEEtran}

\usepackage{multirow}

\usepackage{ifpdf}
\usepackage{epsfig}
\usepackage{subcaption}
\usepackage[utf8]{inputenc}
\usepackage{cite}
\usepackage{graphicx}

\usepackage{amssymb}

\usepackage{hyperref}

\usepackage{color}
\definecolor{veg}{RGB}{0, 251, 20}
\definecolor{sap}{RGB}{226, 175, 165}
\definecolor{che}{RGB}{222, 150, 200}
\definecolor{sul}{RGB}{209, 109, 106}
\definecolor{sha}{RGB}{212, 212, 212}
\definecolor{psha}{RGB}{178, 167, 211}
\definecolor{maf}{RGB}{212, 231, 205}
\definecolor{mafb}{RGB}{50, 205, 50}
\definecolor{fela}{RGB}{175, 238, 238}
\definecolor{felb}{RGB}{244, 199, 131}
\definecolor{fel}{RGB}{252, 225, 198}

\definecolor{Class_1!}{RGB}{226, 175, 165}
\definecolor{Class_2!}{RGB}{222, 150, 200}
\definecolor{Class_3!}{RGB}{209, 109, 106}
\definecolor{Class_4!}{RGB}{212, 212, 212}
\definecolor{Class_5!}{RGB}{178, 167, 211}
\definecolor{Class_6!}{RGB}{212, 231, 205}
\definecolor{Class_7!}{RGB}{50, 205, 50}
\definecolor{Class_8!}{RGB}{175, 238, 238}
\definecolor{Class_9!}{RGB}{244, 199, 131}
\definecolor{Class_10!}{RGB}{252, 225, 198}

\ifCLASSINFOpdf
\else
\fi
\usepackage[cmex10]{amsmath}
\usepackage{algorithmic}
\usepackage{array}
\usepackage{stfloats}
\usepackage{url}
\usepackage{array}
    
\begin{document}
\bstctlcite{IEEEexample:BSTcontrol}

%\title{HyperPointFormer: A Dual-branch Transformer with Cross Attention for Multimodal 3D Point Clouds}
\title{HyperPointFormer: Multimodal Fusion in 3D Space with Dual-Branch Cross-Attention Transformers}

%\author{Michael~Shell,~\IEEEmembership{Member,~IEEE,}
%        John~Doe,~\IEEEmembership{Fellow,~OSA,}
%        and~Jane~Doe,~\IEEEmembership{Life~Fellow,~IEEE}% <-this % stops a space
%\thanks{M. Shell is with the Department
%of Electrical and Computer Engineering, Georgia Institute of Technology, Atlanta,
%\thanks{Manuscript received April 19, 2005; revised September 17, 2014.}}

\author{Aldino~Rizaldy, Richard~Gloaguen, Fabian~Ewald~Fassnacht, and Pedram~Ghamisi,~\IEEEmembership{Senior Member,~IEEE}% <-this % stops a space
\thanks{Aldino Rizaldy is with the Helmholtz-Zentrum Dresden-Rossendorf (HZDR), Helmholtz Institute Freiberg for Resource Technology (HIF), 09599 Freiberg, Germany, and also with the Freie Universität Berlin, Department of Remote Sensing and Geoinformation, 12249 Berlin, Germany.}%
\thanks{Richard Gloaguen and Pedram Ghamisi are with the Helmholtz-Zentrum Dresden-Rossendorf (HZDR), Helmholtz Institute Freiberg for Resource Technology (HIF), 09599 Freiberg, Germany.}%
\thanks{Fabian Ewald Fassnacht is with the Freie Universität Berlin, Department of Remote Sensing and Geoinformation, 12249 Berlin, Germany.}
%\thanks{M. Shell is with the Department
%of Electrical and Computer Engineering, Georgia Institute of Technology, Atlanta,
%\thanks{Manuscript received April 19, 2005; revised September 17, 2014.}
}

\maketitle

\begin{abstract}

Multimodal remote sensing data, including spectral and lidar or photogrammetry, is crucial for achieving satisfactory land-use / land-cover classification results in urban scenes. So far, most studies have been conducted in a 2D context. When 3D information is available in the dataset, it is typically integrated with the 2D data by rasterizing the 3D data into 2D formats. Although this method yields satisfactory classification results, it falls short in fully exploiting the potential of 3D data by restricting the model's ability to learn 3D spatial features directly from raw point clouds. Additionally, it limits the generation of 3D predictions, as the dimensionality of the input data has been reduced. In this study, we propose a fully 3D-based method that fuses all modalities within the 3D point cloud and employs a dedicated dual-branch Transformer model to simultaneously learn geometric and spectral features. To enhance the fusion process, we introduce a cross-attention-based mechanism that fully operates on 3D points, effectively integrating features from various modalities across multiple scales. The purpose of cross-attention is to allow one modality to assess the importance of another by weighing the relevant features. We evaluated our method by comparing it against both 3D and 2D methods using the 2018 IEEE GRSS Data Fusion Contest (DFC2018) dataset. Our findings indicate that 3D fusion delivers competitive results compared to 2D methods and offers more flexibility by providing 3D predictions. These predictions can be projected onto 2D maps, a capability that is not feasible in reverse. Additionally, we evaluated our method on different datasets, specifically the ISPRS Vaihingen 3D and the IEEE 2019 Data Fusion Contest. Our code will be published here: https://github.com/aldinorizaldy/hyperpointformer.

\end{abstract}


% \fi
\IEEEpeerreviewmaketitle

\section{Introduction} 
\label{sec:intro}

\IEEEPARstart{T}{he} rapid development of new types of remote sensing sensors has led to the availability of new data modalities over many areas on Earth \cite{7786851,LI2022102926}. This has led to a recent trend in the utilization of multimodal data \cite{KAHRAMAN2021236,rs13173393}, which led to improved performance in numerous geoscience applications, including, for example, land cover classification.

However, the majority of studies on remote sensing data fusion are conducted in a 2D fashion \cite{10147258,10153685, 10462184}. We argue that this approach is only appropriate when the data are all in 2D format. With the increasing availability of 3D point cloud data \cite{rs12111877}, obtained globally from sources such as lidar or photogrammetry, this approach becomes inadequate, particularly when 3D labeling of point clouds is desired. The main disadvantages of 2D-based fusion are twofold. First, rasterizing 3D point clouds into 2D images prevents the model from learning potentially valuable 3D geometric features, which could hypothetically hinder the model’s full performance. Second, 2D-based fusion inherently produces 2D labels, which may be insufficient in scenarios where semantic objects are vertically defined. For instance, vegetated areas often contain points belonging to different classes within the same horizontal region, such as tree leaves, branches, stems and ground surface. Furthermore, since most 2D-based fusion approaches combine lidar and imaging data from a nadir perspective, underlying objects in the vertical structure cannot be effectively identified. Additionally, projecting labeled 3D points onto a 2D plane is straightforward, whereas the reverse process is theoretically impossible.

In our recent work \cite{10641140}, we showcase the inherent advantages of 3D methodologies over the prevalent 2D approaches in the context of land cover classification in urban areas. Despite achieving commendable results, the neural networks utilized in our study remain insufficient to effectively handle multimodal data since they simply concatenate multiple modalities as input data. The proposed architecture does not adequately consider interactions between the diverse modalities in the dataset.

Despite some prior works on deep learning with multispectral lidar data having been conducted \cite{rs13132516,rs14092113,decker2023hyperspectral,rs15184417}, the interaction among modalities at different scales has yet to be considered by the community.

In this work, we push the boundaries of the state of the art by presenting a neural network explicitly crafted for multimodal 3D point cloud data. This is achieved via the integration of a transformer encoder within a dual-branch network architecture. Addressing multimodal data necessitates distinct sub-networks for each modality to capture precise representations effectively. Moreover, we introduce CrossPointAttention, which not only efficiently fuses different modalities but also enables the network to interact between sub-networks across multiple scales of the encoded features. 

\begin{figure}
    \centering
    \includegraphics[clip=true, trim = 180 50 210 50, width=0.98\linewidth]{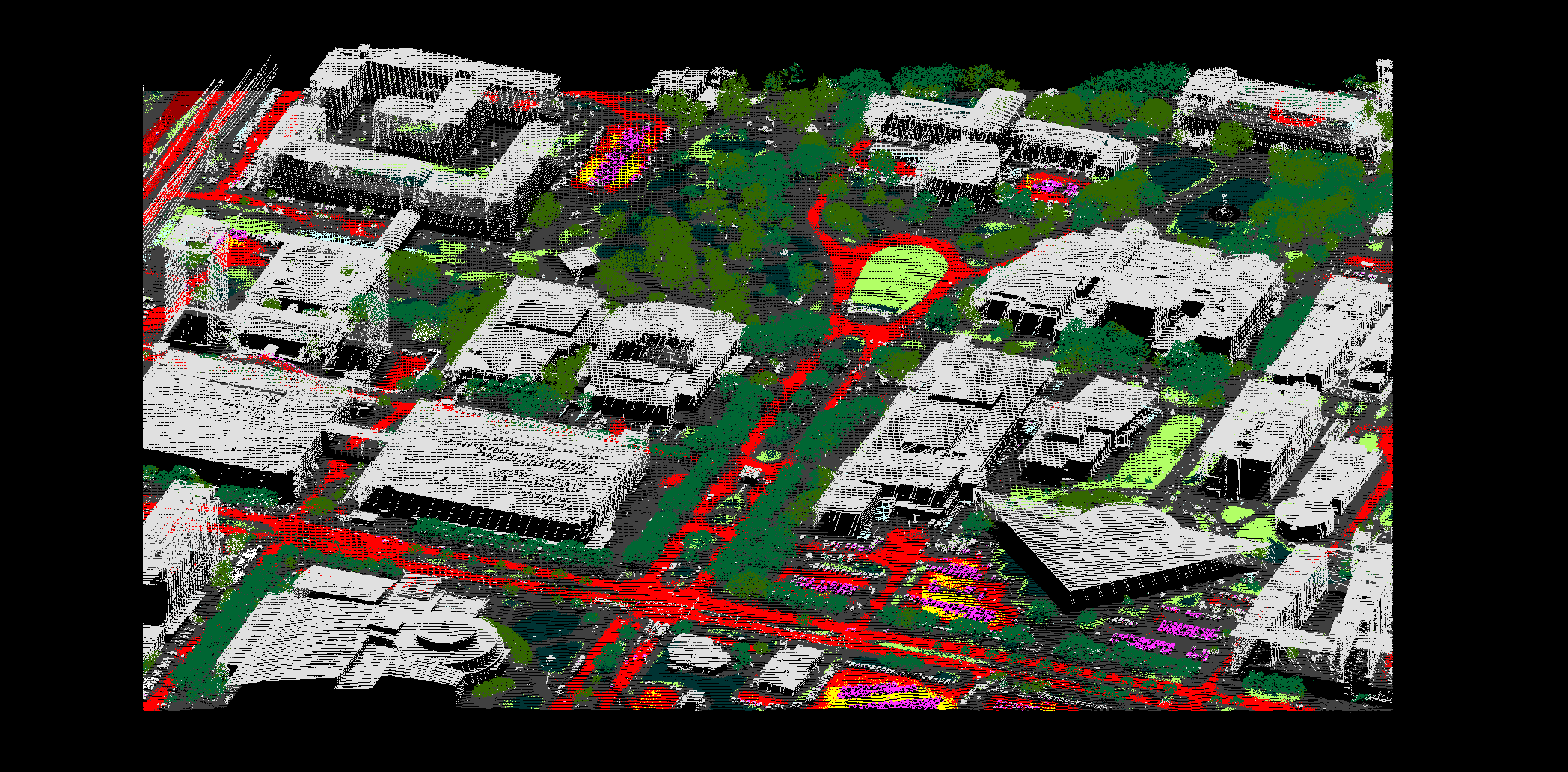}%trim - left, bottom, right, top
    \caption{3D prediction result of the multimodal point clouds.}
    \label{fig:3dhspc2}
\end{figure}

Our method ultimately produces 3D predictions, as illustrated in Fig. \ref{fig:3dhspc2}, using lidar point clouds and hyperspectral data from the 2018 IEEE GRSS Data Fusion Contest (DFC2018) dataset. In this challenging dataset, we show that 3D approaches are competitive with well-established 2D approaches for data fusion. Moreover, 3D approaches can produce predictions at both the 3D points and the 2D maps, whereas 2D approaches can only generate 2D prediction maps.

Our experiments demonstrate that the architecture of our network, coupled with the fusion mechanism we employ, results in improved performance in terms of the accuracy of multimodal remote sensing data classification. Additionally, our design exhibits flexibility and can be readily adapted to accommodate additional modalities from diverse datasets.

The key contributions of our work are as follows.
\begin{itemize}
    \item We present a novel neural network, \textbf{HyperPointFormer}, designed for 3D semantic segmentation by fusing 3D point cloud and 2D imaging data in a fully 3D manner. Unlike commonly used 2D-based fusion methods, our approach employs a complete 3D pipeline, preserving the full 3D spatial and spectral information inherent in the data. This design minimizes information loss and facilitates more accurate segmentation through 3D predictions.
    \item Additionally, we advocate for the adoption of CrossPointAttention, a bidirectional cross-attention mechanism tailored for point cloud data, to enhance the integration of these diverse data modalities at different feature scales.
    \item Our findings indicate that 3D approaches produce results that are competitive with those of 2D approaches. %However, the superiority of 3D methodologies lies in their greater flexibility, allowing for the generation of labels in 3D points and the subsequent projection of these labels onto 2D maps. 
    However, 3D methodologies offer greater flexibility, allowing for the generation of labels in 3D points and the subsequent projection of these labels onto 2D maps.
    \item We demonstrate the flexibility and generalizability of our framework by deploying the network on datasets featuring various modalities and semantic information.
    \item We enhance the lidar and hyperspectral data fusion dataset for the purpose of 3D points learning. The improved dataset will be made publicly available to the community, contributing to the broader research efforts in the field.
\end{itemize}

%The rest of this paper is organized as follows. In Section \ref{sec:related}, we review existing methodologies for multimodal data fusion within the context of 2D approaches. Section \ref{sec:approach} provides an in-depth description of the dual-attention network design, highlighting the integration of the cross-attention mechanism for effective fusion. In Section \ref{sec:exp}, we present the outcomes of our experiments, accompanied by an investigation into hyperparameter sensitivity. In Section \ref{sec:con}, we draw conclusions and delineate potential directions for future research.
The rest of this paper is organized as follows. In Section \ref{sec:related}, we review existing methodologies for multimodal data fusion within the context of 2D approaches. Section \ref{sec:approach} provides an in-depth description of the dual-attention network design, highlighting the integration of the cross-attention mechanism for effective fusion. Section \ref{sec:exp} presents the experimental results. In Section \ref{sec:discussion}, we provide ablation studies, interpret the findings, and discuss the limitations. Finally, Section \ref{sec:con} concludes the paper.

\section{Related work}
\label{sec:related}

In this section, we discuss existing studies on applying deep learning to remote sensing data fusion in a 2D format. Additionally, we briefly review current deep learning models for 3D point cloud processing.

\subsection{2D Fusion Approaches of Multimodal Remote Sensing Data}

Hyperspectral data have been utilized as input to deep learning algorithms for the classification of land cover in urban areas. Hyperspectral data, whether from spaceborne or airborne sensors, is characterized by its high spectral dimensions but typically limited spatial sampling, in contrast to the high spatial resolution of RGB images. Initial approaches involved 1D CNNs in the spectral dimension of the hyperspectral data, leading to a pixel-based classification. This approach has now been improved by incorporating 2D CNNs to learn spatial patterns and ultimately by utilizing 3D CNNs for the simultaneous learning of spectral and spatial patterns \cite{chen, rs9010067, Audebert2019DeepLF}.

With the increasing availability of remote sensing data, the fusion of hyperspectral data with other remote sensing data has been extensively studied in recent years \cite{KAHRAMAN2021236}. Combining hyperspectral and lidar data is particularly appealing, as lidar provides strong 3D spatial features that complement the rich spectral features of hyperspectral data.

The fusion of hyperspectral and lidar data using machine learning has been investigated in many studies and can be categorized into different types \cite{KAHRAMAN2021236}. The first type is data-level fusion, which is achieved by concatenating lidar and hyperspectral data \cite{7729651,7729650,8127075}. A method based on Markov Random Fields (MRF) was proposed, combining hyperspectral and lidar data with a new energy function \cite{7729651}. This method employs a Support Vector Machine (SVM) as the classifier.

The second category is feature-level fusion, which was included in several earlier presented methods \cite{7924601,doi:10.1080/22797254.2017.1314179,7115053}. Multiple levels of features used for fusing hyperspectral and lidar data in urban areas have been investigated \cite{7924601}. The study shows that mid-level morphological features outperform high-level deep learning features. In contrast, a different fusion strategy that avoids concatenating extracted features was proposed to prevent redundant information and excessively high-dimensional features \cite{7115053}. The method adaptively utilizes information from both spatial and spectral features.

Data fusion can also be performed at the object level for classification tasks \cite{4234444,8075399}. Typical methods for object-level classification involve segmenting homogeneous objects, extracting object-based features, and then classifying those features. Random Forests have been used to perform object-based classification of fused hyperspectral and lidar data in urban areas \cite{8075399}. Image segmentation is conducted to obtain objects at multiple scales using hyperspectral data, while lidar data is added as an additional feature for classification.

%Other methods for classifying fused hyperspectral and lidar data include graph-based fusion \cite{6891148,6946657}, kernel-based fusion \cite{7325697,8476167}, and ensemble-based fusion \cite{8517333,8127173}.

Other methods for classifying fused hyperspectral and lidar data include graph-based fusion \cite{6891148,6946657}, which captures similarity relationships between pixels from different modalities. Another approach is kernel-based fusion \cite{7325697,8476167}, which exploits various composite kernel algorithms for feature integration. Additionally, ensemble-based fusion \cite{8517333,8127173} combines multiple classifiers to enhance classification performance.

With the rapid development of analytical methods over the last decade, the fusion of hyperspectral and lidar data has also been achieved by incorporating deep learning models. Currently, studies on data fusion are largely dominated by Convolutional Neural Networks (CNNs). A notable example is a two-tunnel CNN framework with cascade blocks designed to extract spatial-spectral features from hyperspectral and lidar data \cite{8068943}. More recently, a CNN-based model, FusAtNet \cite{9150738}, has been proposed to leverage self-attention and cross-attention mechanisms, utilizing the lidar-derived attention map to enhance the spatial features of hyperspectral data. This trend continues with the development of typical two-stream CNN models, as evidenced in several other works \cite{7786851,9179756,8985546,9057518,9583936,9494718,9598903,AKWENSI2023103302}.

While many studies tend to favor designing two-stream CNNs, one study proposes the use of a single-stream CNN to efficiently learn semantic features from the fusion of hyperspectral and lidar data \cite{9761218}. This was achieved by incorporating dynamic grouping convolutions on top of the CNN backbone model, such as U-Net.

The fusion of hyperspectral and lidar data has also been explored using Transformer models \cite{9999457,9749821,10314566}. Using a Transformer allows the model to learn relationships between features, in contrast to CNNs, which primarily encode local features. A joint classification approach combining CNNs and Transformers was proposed by \cite{9999457, 9749821}, where spectral and spatial features are first encoded using CNNs and then passed to a Transformer encoder for feature fusion. Another approach utilizes a combination of Transformers and CNNs, employing a masked autoencoder to reconstruct both missing spatial pixels and spectral channels during the pretraining stage \cite{10314566}.

The Vision Transformer (ViT) \cite{dosovitskiy2021an} was introduced to the computer vision community as a Transformer-based model for image classification. The adaptation of ViT to hyperspectral data has recently been explored, with the introduction of SpectralFormer \cite{9627165}. Following this trend, several fusion models have been developed using ViT as the base architecture \cite{10147258,10153685,10462184}. For instance, a multimodal fusion transformer (MFT) \cite{10153685} employs ViT as the backbone and incorporates cross-attention to fuse the classification (CLS) token of lidar data with the patch token from hyperspectral data. This approach is similar to Cross-ViT \cite{9711309}, which uses cross-attention to fuse multi-scale images for general image classification. The MFT model was further improved by its successor, Cross-HL \cite{10462184}, which introduces an extended self-attention mechanism to capture complex spectral-spatial relationships.

Despite using ViT as the backbone, these models still rely on CNNs to extract local features in the earlier stages. More combinations of CNN-Transformer models for lidar and hyperspectral fusion can be seen in NCGLF$^2$ \cite{TU2024102192}, MBFormer \cite{10122197}, GLT-Net \cite{9926173}, and DF2NCECT \cite{10439261}. These models employ dual-branch architectures with dedicated encoders for each modality. While NCGLF$^2$ and GLT-Net adopt identical encoders for both modalities, MBFormer and DF2NCECT utilize unique encoders for each modality, specifically spectral-spatial attention for hyperspectral data and convolutional layers for lidar data. In addition, NCGLF$^2$ and GLT-Net emphasize the significance of combining global and local attention.

Multilevel fusion has gained significant attention for its ability to integrate features at various levels using different types of encoders. FTransUNet \cite{10458980} was introduced to address the segmentation of multimodal remote sensing data, specifically rasterized lidar data and optical images. This model leverages CNNs to extract shallow-level features and employs ViTs to capture deep semantic features. Another multilevel fusion model, SoftFormer \cite{LIU2024277}, was proposed for land use and land cover classification by fusing SAR and optical satellite images. SoftFormer combines the strengths of CNNs and Transformers as local feature extractors, with Transformers mimicking the receptive fields of CNNs to effectively extract local features. The final step involves joint learning by integrating CNNs and Transformers.

\subsection{Deep learning models for point cloud classification and segmentation}

Deep learning models for point cloud data have been developed following similar trends to their counterparts in image data. However, the irregularity and unstructured nature of 3D point clouds are key aspects that differentiate point-based processing and image-based processing \cite{Guo2020,9857927}. Many studies have explored approaches where 3D points are projected into multi-view images \cite{7410471,8265294} or occupied voxels \cite{7353481,8579057}. This allows for the use of 2D and 3D CNNs to encode point features. However, these approaches are not always efficient, which led to the introduction of sparse convolution methods to efficiently learn from sparse grid data \cite{8579059,8953494}. Due to their efficiency, these sparse convolution-based models are frequently utilized as the backbone for more complex architectures \cite{9156675,9880069}.

On the contrary, pure point-based models have been developed exclusively to handle point cloud data. The MLP-based architecture was initially proposed to process raw 3D points \cite{pointnet,pointnet2}. Graph-based neural networks are also favored for use with unstructured data such as point clouds \cite{dgcnn,ldgcnn}. Point convolution operates similarly to image convolution but replaces 2D kernels with point kernels. Although convolutional filters are designed for grid-like data, adaptation of convolutional filters has been successfully applied to point clouds \cite{pointcnn,convpoint,kpconv,pointconv}. 

%Lastly, with the advancement of Transformers \cite{point_transformer,NEURIPS2022_d78ece66,pct}, it is possible to use recent deep learning models for point clouds with a Transformer as their sole feature encoder.

Recently, transformer-based models have gained traction in the point cloud processing domain, inspired by their success in natural language processing \cite{transformer} and computer vision tasks \cite{dosovitskiy2021an}. The Point Transformer \cite{point_transformer} and Point Cloud Transformer (PCT) \cite{pct} represent pioneering work in this area. These models employ self-attention mechanisms to effectively capture complex spatial relationships, overcoming the limitations of traditional point-based models.

The development of transformer-based models for point clouds has continued, as evidenced by their consistent outperformance of non-transformer models in 3D vision benchmark datasets \cite{scannet}. Point Transformer V2 \cite{NEURIPS2022_d78ece66} enhances the original Point Transformer by introducing group vector attention, an improvement on the vector attention used in the original model. Meanwhile, Stratified Transformer \cite{9879705} employs standard multi-head self-attention modules, with the key innovation lying in its sampling strategy for preparing key points for the self-attention module. It densely samples nearby points and sparsely samples distant points, effectively capturing long-range features.

Another notable advancement is PointConvFormer \cite{10203572}, an improved version of PointConv \cite{pointconv}. The main improvement involves the integration of an attention mechanism that computes attention of individual points in relation to their surrounding points. This attention is then used to modify the convolutional weights derived from convolutional operations. OneFormer3D \cite{10655551} aims to unify various 3D vision tasks, including semantic, instance, and panoptic segmentation, into a single model. Built on top of 3D MinkUnet \cite{8953494}, which uses sparse convolution to encode rich features, the model employs a Transformer decoder to perform both semantic and instance segmentation tasks.

To address the high computational costs associated with point cloud processing, several recent models have focused on reducing these costs, as computing attention results in quadratic complexity. ConDaFormer \cite{duan2023condaformer} achieves this by breaking a cubic window into three orthogonal 2D planes, enabling a larger attention range without increasing computational expenses. It also employs depth-wise convolution to capture local geometric representations. OctFormer \cite{10.1145/3592131}, as its name suggests, utilizes efficient octree-based operations instead of the regular fixed-size voxel grid format for input to the attention mechanism. The use of octrees is more suitable for point clouds, given the irregular distribution of points, and allows for flexible grid sizes while maintaining a consistent number of points per grid.

While most models are designed to process points individually, Superpoint Transformer \cite{10376484} operates on a hierarchical superpoint structure, which is more efficient for processing large-scale point cloud data. The model improves upon the previous SuperPoint Graph \cite{superpointgraph} model by partitioning points into hierarchical superpoints, segmenting nearby points with similar geometric structures. Self-attention is then used to model relationships between superpoints at different levels.

Swin3D \cite{9879123} leverages sparse voxels to represent point clouds and introduces a novel 3D self-attention mechanism that reduces attention complexity from quadratic to linear by using local window self-attention. Drawing inspiration from the Swin Transformer \cite{9710580}, Swin3D operates on both regular and shifted windows in 3D space. Finally, Point Transformer V3 \cite{10658198} prioritizes simplicity and efficiency by introducing point cloud serialization, which replaces the widely used $k$-nearest neighbors query for neighbor search when formulating local structures. It also employs efficient patch attention, which works on top of point cloud serialization to capture interactions between points across different patches.

Despite the significant improvement in performance, these models were developed to encode geometric features within point cloud data. Building upon these models, some works have investigated the fusion of spectral data and lidar point clouds \cite{rs13132516,rs14092113,decker2023hyperspectral,rs15184417}. A neural network was proposed \cite{rs13132516} by combining PointNet++ \cite{pointnet2} and SENet \cite{8701503} to process multispectral lidar data. Meanwhile, comprehensive studies \cite{rs14092113,decker2023hyperspectral} investigated the utilization of KPConv \cite{kpconv} as the backbone for deep learning models used in the classification of hyperspectral and lidar data. In a different study \cite{rs15184417}, a graph-based neural network is chosen to fully exploit joint spatial-spectral information.

Our approach builds upon the advancements made in prior works, leveraging deep learning models designed for 3D point clouds and incorporating fusion techniques for hyperspectral and lidar data.

\section{Methodology}
\label{sec:approach}
In this section, we begin by introducing the overall structure of the proposed HyperPointFormer network. Subsequently, we present CrossPointAttention as a crucial module for learning fused features from the 3D multimodal point cloud data.

\subsection{Architecture}

\textbf{Early and late fusion.} Classic fusion strategies involve early fusion and late fusion. Before exploring our approach, we briefly describe these two classic fusion strategies to establish the rationale behind our method and highlight the limitations posed by these strategies.

\begin{figure}[!t]
\centering
\subfloat[]{\includegraphics[clip=true, trim = 50 130 50 130, width=0.98\linewidth]{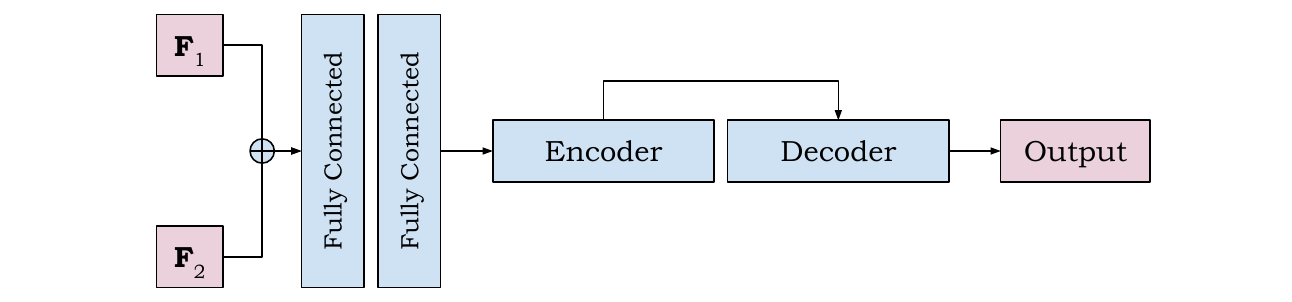}%trim - left, bottom, right, top
\label{fig:early}}
\hfil
\subfloat[]{\includegraphics[clip=true, trim = 50 100 50 100, width=0.98\linewidth]{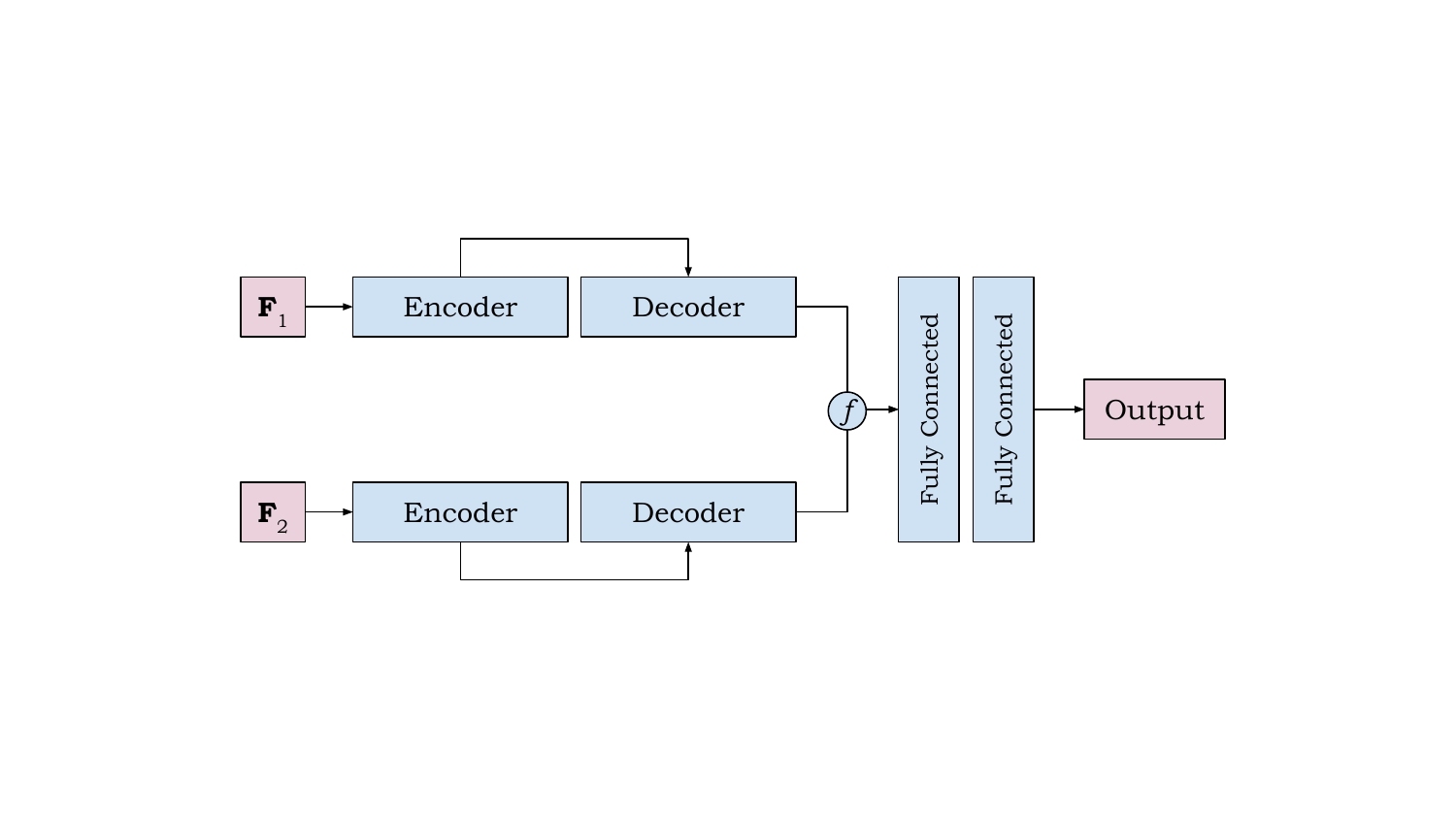}%trim - left, bottom, right, top
\label{fig:late}}
\hfil
\subfloat[]{\includegraphics[clip=true, trim = 50 100 50 100, width=0.98\linewidth]{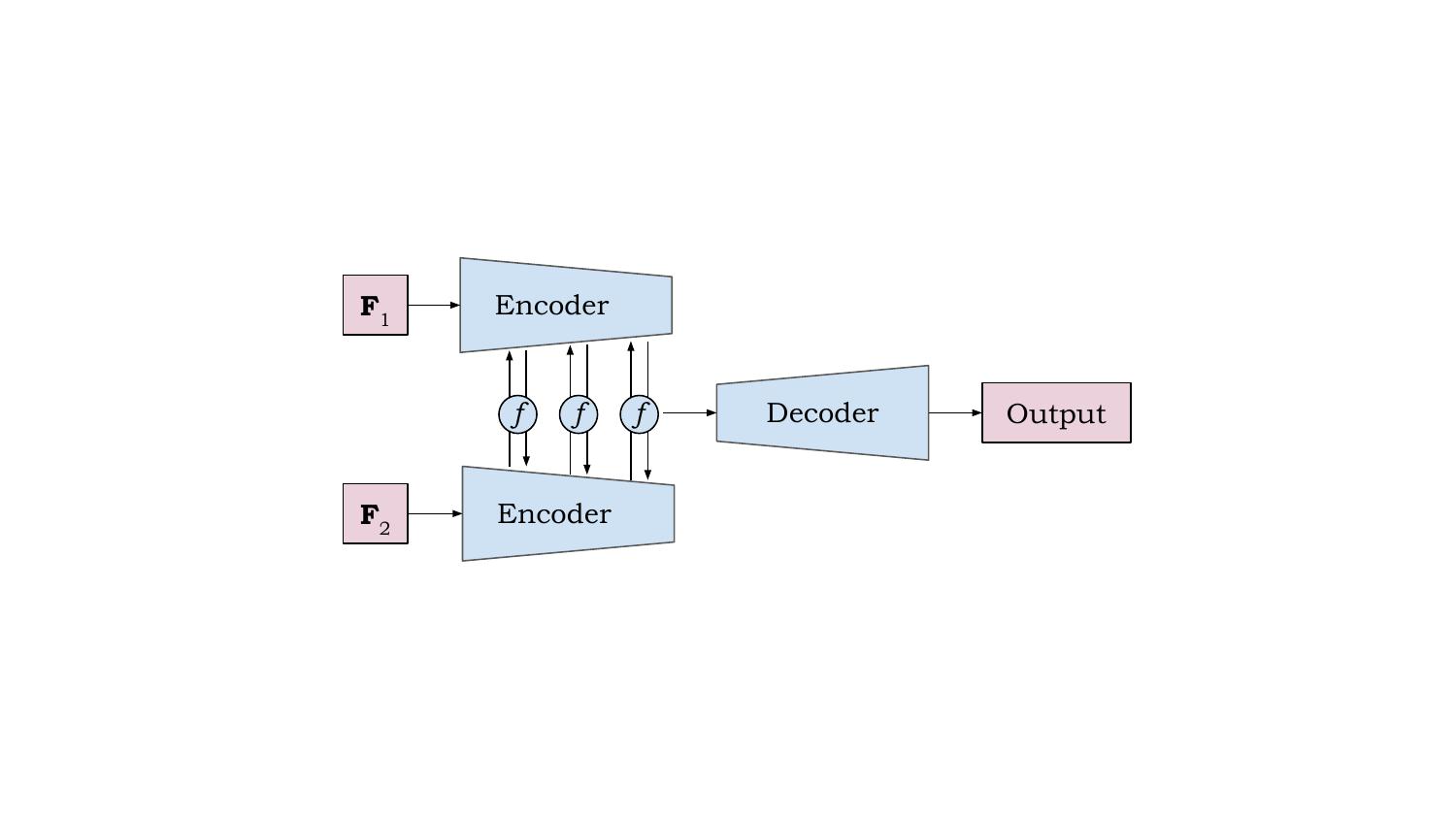}%trim - left, bottom, right, top
\label{fig:multi-scale}}
\caption{Classical fusion methods: (a) early, (b) late fusion, and (c) our multi-scale features fusion.}
\label{fig:earlylate}
\end{figure}

%A straightforward solution for a dataset with different modalities is early fusion, also known as data-level fusion. Early fusion, as illustrated in Fig. \ref{fig:early}, involves stacking all modalities into a feature set, which is then fed to the machine learning models. Assuming $\mathbf{F}_{1}$ and $\mathbf{F}_{2}$ represent distinct modalities of remote sensing data within the same spatial area, early fusion involves a straightforward combination of $\mathbf{F}_{1}$ and $\mathbf{F}_{2}$ through channel-wise concatenation, as illustrated in Eq. \ref{eq:concat}. 
A common solution for multimodal datasets is early fusion, also known as data-level fusion. Early fusion, as illustrated in Fig. \ref{fig:early}, involves stacking all modalities into a combined feature set that is then fed into machine learning models. Assuming $\mathbf{F}_{1}$ and $\mathbf{F}_{2}$ represent different modalities of remote sensing data, early fusion combines them through channel-wise concatenation, as shown in Eq. \ref{eq:concat}.

Limitations of early fusion can lead to sub-optimal performance in tasks where modality-specific features are crucial for distinguishing between classes. For example, in remote sensing data, integrating hyperspectral and lidar information at an early stage may obscure valuable distinctions between material properties and spatial structures. Additionally, this early fusion approach can lead to an increased risk of feature interference, where the contributions of each modality may overshadow one another, resulting in less effective feature representation.

% The advantage of early fusion is a lighter neural network and faster computation time. However, it does not treat different modalities as separate representations, thus failing to capture the unique characteristics of each modality.

% Assuming $\mathbf{F}_{1}$ and $\mathbf{F}_{2}$ represent distinct modalities of remote sensing data within the same spatial area, early fusion involves a straightforward combination of $\mathbf{F}_{1}$ and $\mathbf{F}_{2}$ through channel-wise concatenation, as illustrated in Eq. \ref{eq:concat}. In neural network architectures, this results in feature interactions occurring primarily in the first layer of the network. The consequences are twofold: first, the features passed to subsequent layers are already fused, hindering the extraction of the unique characteristics of each modality separately. Second, this implies that the fused features are primarily learned at a low level of abstraction, leaving the ability to learn fused features at a higher level of abstraction.

The contrasting approach to early fusion is late fusion. In this strategy, as illustrated in Fig. \ref{fig:late}, each modality has its own dedicated sub-network. The features learned by each sub-network are subsequently concatenated in the final layer. 

With $f$ as the fusion function of the given $\mathbf{F}_{1}$ and $\mathbf{F}_{2}$ features from different modalities, and \texttt{concat} representing channel-wise concatenation, features fusion of $\mathbf{F}_{1}$ and $\mathbf{F}_{2}$ at any arbitrary layer with the same number of samples can be mathematically expressed as:
\begin{equation}
\label{eq:concat}
    %f \left ( \mathbf{F}_{1}, \mathbf{F}_{2} \right ) = \mathbf{F}_{1} \mathbin\Vert \mathbf{F}_{2}
    \mathbf{F}_{\text{fuse}} = f \left ( \mathbf{F}_{1}, \mathbf{F}_{2} \right ) = \texttt{concat} \left ( \mathbf{F}_{1}, \mathbf{F}_{2} \right )
\end{equation}

Here, $\mathbf{F}_{\text{fuse}}$ represents the concatenated feature map, combining the two modalities along the channel dimension.

As an alternative to features concatenation, late fusion can also be done during the computation of the loss function by combining losses from the separate branches as the final loss $\mathcal{L}$, as shown in Eq. (\ref{eq:late2}). Typically, a weighting factor $\alpha$ is introduced to control the contribution of each modality to the total loss. This ensures a balanced influence from both branches during optimization.

% However, late fusion through loss function combination might lead to issues with balancing the contributions of different modalities to the final loss.

\begin{equation}
\label{eq:late2}
    \mathcal{L} = \alpha \mathcal{L}_{1} + (1 - \alpha) \mathcal{L}_{2} 
\end{equation}

Despite each modality having its own dedicated feature encoder, late fusion may lead to issues with balancing the contributions of each modality, especially if the modalities have vastly different scales or if one modality dominates the learning process. Additionally, combining losses from separate branches might require careful tuning to ensure that the model does not bias towards modalities with smaller losses.

% In contrast to early fusion, late fusion typically results in a heavier model, as the overall architecture consists of two or more sub-networks with their own parameters. The advantage of this strategy lies in each modality having a dedicated feature encoder, enabling the network to extract different features from different modalities. This can be understood as each modality corresponding to specific semantic information. For instance, in remote sensing data, hyperspectral sensors are more inclined to detect the material of objects, while lidar sensors are more suitable for detecting the shape of objects.

% Both classic fusion strategies have their own set of advantages and disadvantages. Early fusion, while simple, falls short in capturing the unique representation of each modality. On the other hand, late fusion is accommodating but tends to result in a heavier model. More importantly, both designs lack continuous interaction between different modalities at different scales.

Regardless of the fusion technique employed, normalization plays a crucial role in ensuring that all data sources contribute fairly to the model. In data fusion, features from different modalities often have varying scales and ranges. Normalization methods standardize these features to a consistent scale, preventing any single modality from dominating the fusion process and ensuring a balanced contribution from each modality. For example, when fusing optical and SAR data, feature normalization is essential due to the differing imaging mechanisms of the two sensors \cite{KAHRAMAN2021236,7004791}.

\textbf{HyperPointFormer.} To address the aforementioned challenges, we introduce HyperPointFormer — a dual-branch Transformer with encoder-decoder architecture and a fusion module that integrates multi-scale features, as illustrated in Fig. \ref{fig:multi-scale}. This specialized network is designed exclusively for learning from multimodal 3D point clouds. Please note that multimodal 3D point cloud data is a fusion of 3D spatial information (i.e., XYZ coordinates) with associated spectral attributes from other sensors. Our method purely processes point clouds in their original 3D point format, distinguishing it from previous methods that use rasterized point clouds \cite{9999457,9749821,10314566}.

\begin{figure*}
    \centering
    \includegraphics[clip=true, trim = 0 0 0 0, width=0.98\linewidth]{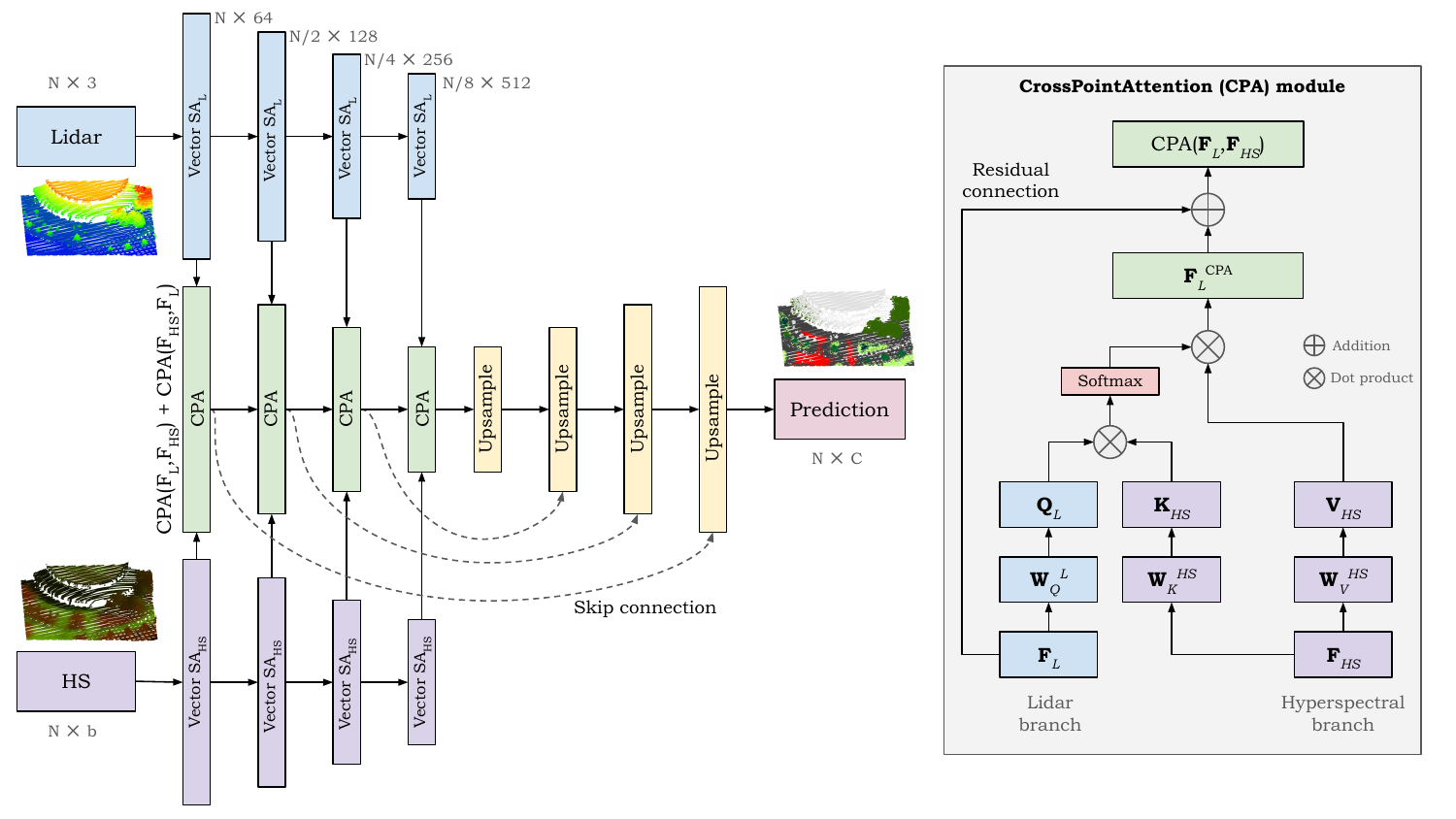}%trim - left, bottom, right, top
    \caption{The overall architecture of HyperPointFormer and the CrossPointAttention (\texttt{CPA}) module. HyperPointFormer takes XYZ coordinates ($N \times 3$) and spectral information ($N \times b$) of multimodal lidar point clouds at different branches, where $N$ is the number of points and $b$ is the number of spectral bands. It utilizes Vector Self-Attention as the backbone encoders, while Farthest Point Sampling (FPS) is used to downsample the number of points at each layer. The \texttt{CPA} module, which is explained in more detail in Section \ref{sec:fusion}, takes encoded features from different modalities at various scales and fuses them using cross-attention mechanism. Additionally, skip connections are used in the upsampling stages to enhance the detail in the segmentation outputs. The output is a predicted label for each 3D point ($N \times C$), where $C$ represents the number of classes.}
    \label{fig:arch}
\end{figure*}

In terms of modality, our method was initially developed to address the fusion of lidar and hyperspectral data in a complex classification problem, as posed in the DFC2018 dataset. Subsequently, we evaluated our method on various datasets where we integrated point cloud data with additional spectral information, such as near-infrared and RGB images. It could also be used with photogrammetric information or textures. 

The number of branches in HyperPointFormer is arbitrary and depends on the modalities present in the given dataset. However, since most modalities in remote sensing data can be categorized into geometrical and spectral representations, we intentionally designed HyperPointFormer as a dual-branch neural network for simplicity. This design choice balances the model complexity with its ability to handle diverse data types effectively.

Let us consider a dataset containing two different modalities: lidar and hyperspectral data, represented as $\mathbf{X}_{L}\in \mathbb{R}^{N \times 3}$ and $\mathbf{X}_{HS}\in \mathbb{R}^{N \times b}$, respectively. Here, $N$ denotes the number of points and $b$ represents the number of hyperspectral bands. The lidar data $\mathbf{X}_{L}$ exclusively contains XYZ coordinates captured by the lidar sensors, while $\mathbf{X}_{HS}$ contains the corresponding hyperspectral features. In the HyperPointFormer architecture, these modalities are processed independently within separate branches of the encoder section, resulting in latent features $\mathbf{F}_{L}\in \mathbb{R}^{N \times d}$ and $\mathbf{F}_{HS}\in \mathbb{R}^{N \times d}$ with \textit{d}-dimensional feature description.

Our proposed architecture is based on the assumptions that lidar data primarily provide geometrical features, which correspond to different semantic information found in the spectral features captured by hyperspectral data. Moreover, the Transformer architecture allows the network to learn interactions among features \cite{dosovitskiy2021an}. We therefore designed a dual-branch neural network with Transformer encoders to effectively integrate these complementary modalities.

Nevertheless, the process is not completely separable since our proposed fusion mechanism actively integrates the output feature maps from the two branches at different scales. It is important to note that the cardinality of $N$ changes by half in the subsequent layers. Further explanation on the fusion process is provided in Section \ref{sec:fusion}.

To ensure that the features are sufficiently prepared for fusion, our approach focuses on feature fusion in the latent space to enable effective integration of data from multiple modalities. Prior to the fusion process, features from distinct sources, such as 3D point clouds and 2D spectral images, are encoded separately to generate a unified representation. This encoding step transforms raw, modality-specific data into a common representational form, facilitating more meaningful fusion. By performing the fusion in the latent space, rather than directly on raw features, we mitigate challenges arising from the heterogeneous nature of the input data.

In each branch, following \cite{pct, point_transformer, NEURIPS2022_d78ece66}, HyperPointFormer is composed of four encoder blocks. Each block contains a transformer layer followed by a downsample layer. Together, these layers extract local attention features and reduce the number of points, leading to a higher level of abstraction in subsequent layers. The decoder, employing skip connections, then upscales the learned features to restore the original point resolution, akin to the operation of the Unet architecture. The overall architecture of our proposed model is illustrated in Fig. \ref{fig:arch}.

\textbf{Selection of self-attention mechanism.} Initially, we assessed several alternatives for crafting the transformer layer. Over the past few years, researchers have introduced at least three distinct attention modules tailored for capturing point features. These comprise Scalar Self-Attention (SSA), Offset Self-Attention (OSA), and Vector Self-Attention (VSA) \cite{lu2022transformers3dpointclouds}.

The conventional and widely adopted attention mechanism is the Scalar Self-Attention (SSA) module, as found in the original ViT implementation \cite{dosovitskiy2021an}. Given the input feature $\mathbf{F}_{in}\in \mathbb{R}^{N \times d}$, SSA transforms $\mathbf{F}_{in}$ into $\mathbf{Q}, \mathbf{K}, \mathbf{V}$ matrices through linear transformation with learnable weights $\mathbf{W}_{Q}, \mathbf{W}_{K}, \mathbf{W}_{V}$. 

SSA first computes attention scores $\mathbf{A}_{\text{SSA}}\in \mathbb{R}^{N \times N}$ as the dot product between $\mathbf{Q}$ and $\mathbf{K}$, as shown in Eq. \ref{eq:ssa}. These attention scores illustrate the semantic similarity among input features. The final output features $\mathbf{F}_{\text{SSA}}$ are then computed by performing a second dot product between $\mathbf{A}_{\text{SSA}}$ and $\mathbf{V}$, as depicted in Eq. \ref{eq:ssa_out}

\begin{equation}
\label{eq:ssa}
    \mathbf{A}_{\text{SSA}} = \texttt{softmax} \left ( \frac{\mathbf{Q}\cdot \mathbf{K}^{\text{T}}}{\sqrt{d}} \right ) 
\end{equation}

\begin{equation}
\label{eq:ssa_out}
    \mathbf{F}_{\text{SSA}} = \mathbf{A}_{\text{SSA}} \cdot \mathbf{V}
\end{equation}

Recently, Point Cloud Transformer (PCT) \cite{pct} enhanced the original SSA into Offset Self-Attention (OSA). OSA subtracts the input features from the SSA features to obtain the offset (difference) features, inspired by the Laplacian matrix. This simple operation enhances the network performance in point cloud classification tasks. However, its performance in segmentation tasks is poor, as the attention design is applied without considering hierarchical features.

\begin{equation}
\label{eq:osa}
    \mathbf{F}_{\text{OSA}} = \mathbf{F}_{in} - \left ( \mathbf{A}_{\text{SSA}} \cdot \mathbf{V} \right )
\end{equation}

Point Transformer \cite{point_transformer, NEURIPS2022_d78ece66}, on the other hand, proposed a different neural network design for point cloud data by replacing scalar (dot product) attention with vector attention. This mechanism, first introduced by \cite{9156532}, adapts attention weights at different channels by employing adaptive weight vectors. In contrast, the dot product operation produces a scalar output that is shared across all channels. The operations to obtain VSA features are formulated in Eq. \ref{eq:vsa} and \ref{eq:vsa_out}:

\begin{equation}
\label{eq:vsa}
    \mathbf{A}_{\text{VSA}} = \texttt{softmax} \left ( \frac{\beta \left (\mathbf{Q}, \mathbf{K} \right )}{\sqrt{d}} \right ), 
\end{equation}

\begin{equation}
\label{eq:vsa_out}
    \mathbf{F}_{\text{VSA}} = \mathbf{A}_{\text{VSA}} \odot \mathbf{V}
\end{equation}
where possible functions for $\beta$ to produce vector output include summation, subtraction, and Hadamard product \cite{9156532}.

%We conducted an experiment, as depicted in Table \ref{tab:relational_fn}, to compare the three relational functions and found that subtraction performs better for our task. \colorbox{yellow}{explain why}. It aligns with the study \cite{point_transformer} that suggests subtraction as the selected relational function. 

To select the most suitable attention mechanism, we conducted a series of experiments. These experiments were based on the 2018 IEEE GRSS Data Fusion Contest dataset, which we further processed and modified to create a 3D point cloud dataset enriched with hyperspectral features and accurate 3D labels. Additional details about the dataset can be found in Section \ref{sec:grss}. This dataset was used to validate our 3D approach for multimodal data fusion. Specifically, we trained models using Area 1 and evaluated them on Area 2, as illustrated in Fig. \ref{fig:2fold}.

Our first experiment focuses on utilizing VSA and comparing different relational functions, as suggested in a prior study \cite{9156532}. The dual-branch architecture described in the previous subsection serves as the basis for this experiment. To compute attention, we utilize pairwise attention, which calculates the attention of data points based on their local neighboring points. In this setup, $\mathbf{Q}$ represents the transformed features of the central points, while $\mathbf{K}$ and $\mathbf{V}$ are the transformed features of the neighboring points. Finally, the relational function $\beta$ in Eq. \ref{eq:vsa} is replaced with the Hadamard product, summation, and subtraction to evaluate their impact on performance.

Table \ref{tab:relational_fn} presents the results of using different relational functions. Our findings indicate that the subtraction function significantly outperforms the other relational functions. A possible explanation for this is that subtraction, unlike summation and the Hadamard product, is a noncommutative operation, meaning the order of the operation is important and switching the order produces different results. Since the goal is to capture the relationship between central points and their surroundings, the order is crucial. Let $\textbf{x}_{i}$ represent the central point and $\textbf{x}_{j}$ represent the surrounding points, if $\textbf{x}_{i} - \textbf{x}_{j} = a$, then $\textbf{x}_{j} - \textbf{x}_{i} \neq a$, maintaining the relative position of $\textbf{x}_{i}$ and $\textbf{x}_{j}$, and preserving the directional information. In contrast, summation is commutative $(\textbf{x}_{i} + \textbf{x}_{j} = \textbf{x}_{j} + \textbf{x}_{i})$, which can obscure the relative position of points.

\begin{table}[ht]
    \centering
    \caption{Selection of Relational Function $\beta$}
    \label{tab:relational_fn}
    \begin{tabular}{llll}
        \hline
        \textbf{Relational function} & \textbf{Precision (\%)} & \textbf{Recall (\%)} & \textbf{F1 (\%)} \\ 
        \hline
        Hadamard product & 56.82 & 48.44 & 48.36 \\
        Summation & 56.42 & 44.50 & 45.48 \\
        Subtraction & 69.05 & 53.40 & 55.54 \\
        \hline
    \end{tabular}
\end{table}

Additionally, summation is inherently an aggregation function, which does not effectively define the geometry of points in relation to their surroundings. This makes it less suitable for capturing spatial structures and is more appropriate for feature fusion tasks.

Subtraction also exhibits translation invariance, a critical property for maintaining the geometrical structure of points. For instance, if $\textbf{x}_{i} - \textbf{x}_{j} = a$ and we add a translation of $b$ to both points, then $(\textbf{x}_{i}+b) - (\textbf{x}_{j}+b) = a$, ensuring consistent results regardless of translation. In contrast, summation is not translation invariant. For example, if $\textbf{x}_{i} + \textbf{x}_{j} = a$, then $(\textbf{x}_{i}+b) + (\textbf{x}_{j}+b) \neq a$. Translation invariance is vital as it allows similar objects located in different positions to yield consistent geometric features.

Moreover, this approach aligns with the design of edge graph operators \cite{dgcnn}. These operators compute features by considering the differences between points and their neighbors, effectively leveraging subtraction to encode the local structures of point clouds.

We then conducted additional experiments to validate our choice by comparing different self-attention mechanisms for point cloud data. We follow our previous experiments by computing attention within local structure and incorporating SSA, OSA, and VSA, inspired by \cite{pct,point_transformer}, into our dual-branch architecture.

%Table \ref{tab:SAM} indicates that VSA outperformed SSA and OSA. \colorbox{yellow}{explain why}. This result confirms previous reports \cite{point_transformer, NEURIPS2022_d78ece66}, which suggested to use vector attention over scalar attention for point cloud data. 

Table \ref{tab:SAM} presents the results of different self-attention mechanisms. We observed that VSA outperforms both SSA and OSA. We argue that this is due to the commutative nature of the dot product operation used in both SSA and OSA, leading to less sensitivity of the relative positions. On the other hand, VSA preserves the directional and relational information between the central and surrounding points.

\begin{table}[ht]
    \centering
    \caption{Selection of Self-Attention}
    \label{tab:SAM}
    \begin{tabular}{llll}
        \hline
        \textbf{Attention} & \textbf{Precision (\%)} & \textbf{Recall (\%)} & \textbf{F1 (\%)} \\ 
        \hline
        Scalar Self-Attention & 65.60 & 51.52 & 53.02 \\
        Offset Self-Attention & 66.52 & 52.15 & 53.82 \\
        Vector Self-Attention & 69.05 & 53.40 & 55.54 \\
        \hline
    \end{tabular}
\end{table}

However, the differences are not as significant as the previous experiment, because SSA and OSA employ self-attention operations rather than simple summation or multiplication. Despite the commutative nature of the dot product, these mechanisms still provide some level of flexibility in encoding local structure.

%\textbf{Selection of global and local attentions.} Finally, we assessed the decision to utilize attention mechanisms for computing either local attention features and/or global attention features. Local attention excels at refining local structures and modeling interactions among neighboring points. Conversely, global attention captures long-range dependencies among all input points but requires heavy computation on large-scale datasets.

%Local attention is effectively utilized in \cite{point_transformer, NEURIPS2022_d78ece66}, which follows a Unet-like network structure, while global attention is successfully employed in \cite{8578582, pct}, which follows a PointNet-like network structure.

%To capture finer details in the segmentation task, we chose to utilize attention mechanisms as local feature extractors. However, local attention is limited to small spatial areas. To incorporate global contexts, a downsampling mechanism is integrated similar to those used in Unet, aggregating features in the subsequent layers.

%The downsampling mechanism gradually reduces the number of points in subsequent layers using the Farthest Point Sampling (FPS) method. Our FPS implementation leverages the Deep Graph Library (DGL) instead of using a custom CUDA kernel, as introduced by PointNet++. This approach ensures seamless and efficient Python integration.

\textbf{Selection of global and local attentions.} Recent studies in remote sensing data fusion \cite{TU2024102192,9926173} highlight the importance of integrating local and global attention mechanisms. Given that segmentation tasks demand strong per-point features \cite{pointnet2} and considering memory efficiency constraints, our design relies solely on local attention. We conducted additional experiments to validate this choice by comparing local and global attention mechanisms.

It is important to acknowledge, however, that local attention or other localized operations alone may not effectively capture the broader context of the data. This limitation is often addressed in point clouds models by incorporating downsampling layers to capture global context \cite{pointnet2,kpconv,point_transformer}. Meanwhile, global attention for point clouds is also widely used in recent point cloud models as well \cite{8578582, pct}.

In our approach, we utilize local attention combined with downsample layers. To reconstruct fine details, we also include upsampling layers. The downsampling mechanism gradually reduces the number of points in subsequent layers using the Farthest Point Sampling (FPS) method. Our FPS implementation leverages the Deep Graph Library (DGL) instead of using a custom CUDA kernel, ensuring seamless and efficient Python integration.

To validate our design, we performed experiments comparing our local attention-based model with one using global attention. The fundamental difference between local and global attention lies in how the $\mathbf{Q}$, $\mathbf{K}$, and $\mathbf{V}$ are computed. In local attention, $\mathbf{Q}$ is derived from the points, while $\mathbf{K}$ and $\mathbf{V}$ are obtained from neighboring points. In contrast, global attention mechanisms compute $\mathbf{K}$ and $\mathbf{V}$ from all points, as proposed in \cite{pct}, enabling the attention to reflect relationships between each point and the entire dataset.

The results, shown in Table \ref{tab:global_local}, present that local attention paired with downsampling mechanisms significantly outperforms global attention. We attribute this to the segmentation task’s need for strong per-point information, which the global attention mechanism fails to deliver. Instead, global attention is better suited for object classification tasks, as evidenced in \cite{pct}, where it provides holistic representation of the entire point cloud.

%Table \ref{tab:global_local} presents the quantitative comparison between employing global and local self attentions. 

\begin{table}[ht]
    \centering
    \caption{Selection of Global and Local Attention Features}
    \label{tab:global_local}
    \begin{tabular}{llll}
        \hline
        \textbf{Attention} & \textbf{Precision (\%)} & \textbf{Recall (\%)} & \textbf{F1 (\%)} \\ 
        \hline
        Global attention & 47.53 & 39.69 & 38.91 \\
        Local attention & 69.05 & 53.40 & 55.54 \\
        \hline
    \end{tabular}
\end{table}

To demonstrate how downsampling layers contribute to achieving more global context, we provide the following explanation. In our model, points are downsampled in four stages, with each stage reducing the number of points by half. This design allows the model to progressively learn attention over increasingly larger areas in subsequent layers. By the final layer in the series of downsampling stages, the model captures much broader global contexts without increasing computational cost, as attention is still computed over the same number of $k$-neighboring points.

Specifically, the model begins with an input of $4096$ points. At each stage, the attention for each point is calculated over $k$ neighboring points, where $k$ is set to $16$. By the fourth stage, the model effectively learns attention within an area encompassing $128$ points under regular distribution conditions, covering $1/32$ of the total points. This is a significant increase compared to the initial $1/256$ coverage in the first stage, enabling the model to learn attention over a much larger context.

Table \ref{tab:local_attention_coverage} illustrates the progression of local attention coverage at each stage of the model. It is important to note that the "Coverage" and "Percentage" columns are based on the assumption of perfectly regular point distributions.

\begin{table}[ht]
    \centering
    \caption{Coverage of Local Attention in Each Stage}
    \label{tab:local_attention_coverage}
    \begin{tabular}{lllll}
        \hline
        \textbf{Stage} & \textbf{Number of Points} & \textbf{$k$} & \textbf{Coverage} & \textbf{{Percentage}}\\
        \hline
        1 & 4096 & 16 & 16 & 1/256 \\
        2 & 2048 & 16 & 32 & 1/128 \\
        3 & 1024 & 16 & 64 & 1/64 \\
        4 & 512 & 16 & 128 & 1/32 \\
        \hline
    \end{tabular}
\end{table}

In terms of computational complexity, a global attention layer would have a complexity of in $O(N^{2})$, while a local attention layer with $k$-neighboring points would have a complexity of $O(k.N)$.

\subsection{Fusion Modules}
\label{sec:fusion}

Given $\mathbf{F}_{L}$ and $\mathbf{F}_{HS}$ as encoded features in each layer at different scale of a dual-branch network, one can define fusion function $f\left(\mathbf{F}_{L}, \mathbf{F}_{HS}\right)$ to obtain $\mathbf{F}_{\text{fuse}}$. The straightforward choices for fusion function $f$ are:

$\textbf{Summation: } f\left(\mathbf{F}_{L}, \mathbf{F}_{HS}\right) = \mathbf{F}_{L} + \mathbf{F}_{HS}$

% $\textbf{Concatenation: } f\left(\mathbf{F}_{L}, \mathbf{F}_{HS}\right) = \mathbf{F}_{L} \mathbin\Vert \mathbf{F}_{HS}$
$\textbf{Concatenation: } f\left(\mathbf{F}_{L}, \mathbf{F}_{HS}\right) = \texttt{concat} \left ( \mathbf{F}_{L}, \mathbf{F}_{HS} \right )$

$\textbf{Average pooling: } f\left(\mathbf{F}_{L}, \mathbf{F}_{HS}\right) = \texttt{average} \left ( \mathbf{F}_{L}, \mathbf{F}_{HS} \right )$

$\textbf{Max pooling: } f\left(\mathbf{F}_{L}, \mathbf{F}_{HS}\right) = \texttt{maxpooling} \left ( \mathbf{F}_{L}, \mathbf{F}_{HS} \right )$

Despite the availability of straightforward solutions, the drawback of these classical fusion methods is that it neglects the interaction between the learned features. We address this limitation by proposing a fusion technique, which enables features from one branch to interact with features from another branch in a query-key-value manner.

\textbf{Cross attention fusion:} Motivated by Cross-ViT \cite{9711309}, which uses cross attention to learn multi-scale patches of image data, we investigated the use of cross attention for fusing multimodal point cloud data. CrossViT achieved superior performance compared to vanilla ViT by learning multi-scale features through a cross-attention mechanism on inputs of different scales. In CrossViT, the CLS (classification) token from one branch serves as the query, while patch tokens from another branch act as the key and value. Cross-attention features are then computed using query, key, and value matrices. We then design a fusion function, which takes encoded features $\mathbf{F}_{L}$ and $ \mathbf{F}_{HS}$ as input and produces the fused features $\mathbf{F}_{\text{fuse}}$ using a cross-attention mechanism. We refer to our fusion operation as CrossPointAttention (\texttt{CPA}). 

\begin{equation}
\label{eq:cpa_all}
    \mathbf{F}_{\text{fuse}} = f\left(\mathbf{F}_{L}, \mathbf{F}_{HS}\right) = \texttt{CPA}\left ( \mathbf{F}_{L}, \mathbf{F}_{HS} \right ) + \texttt{CPA}\left ( \mathbf{F}_{HS}, \mathbf{F}_{L} \right )
\end{equation}

In Cross-ViT, the cross-attention features are learned from the CLS and patch tokens, which works exceptionally well for classification tasks. However, since we designed our fusion for a segmentation task, \texttt{CPA} differs from CrossViT in the way cross-attention features are computed. In \texttt{CPA}, the query is derived from an entire patch (e.g., neighboring points) from one branch, while the key and value are taken from an entire patch from another branch.

More specifically, the idea is to use the embedding features from one branch as the query and find the key from the embedding features in the other branch. The query and the key then produce attention scores to compute the weighted sum of the value. We argue that this operation measures the relevance of features from one modality to another. In short, the fusion is accomplished by computing the cross-attention features for both branches, taking the features of the other branch into account.

Suppose $\texttt{CPA}\left ( \mathbf{F}_{L}, \mathbf{F}_{HS} \right )$ as the fused features at lidar branch, first $\mathbf{F}_{L}$ and $\mathbf{F}_{HS}$ are transformed into $\mathbf{Q}_{L}, \mathbf{K}_{L}, \mathbf{V}_{L},$ and $\mathbf{Q}_{HS}, \mathbf{K}_{HS}, \mathbf{V}_{HS}$, respectively. Subsequently, $\mathbf{Q}_{L}$ is taken as the query matrix and $\mathbf{K}_{HS}$ and $\mathbf{V}_{HS}$ as the key and value matrices, and the relevance is measured using the scaled dot product. This outputs cross attention scores $\mathbf{A}^{\texttt{CPA}}_{L}$.

\begin{equation}
\label{eq:cross_attn_map}
    \mathbf{A}^{\texttt{CPA}}_{L} = \texttt{softmax}\left ( \frac{\mathbf{Q}_{L}\cdot \mathbf{K}_{HS}^{\text{T}}}{\sqrt{d_{e}}} \right ),
\end{equation}

where $d_{e}$ is the embedding dimension. The attention score is then multiplied with the value $\mathbf{V}_{HS}$ to obtain the weighted sum of the embedding features from another branch as the fused features $\mathbf{F}^\texttt{CPA}_{L}$. Finally, residual connection is used by adding the computed $\mathbf{F}^\texttt{CPA}_{L}$ with its input features of lidar branch $\mathbf{F}_{L}$. Additionally, inspired by Dual Attention Networks (DANet) \cite{8953974}, $\gamma$ is added as a learnable parameter to modulate the impact of $\mathbf{F}^\texttt{CPA}_{L}$.

\begin{equation}
\label{eq:cross_attn}
    \mathbf{F}^\texttt{CPA}_{L} = \mathbf{A}^{\texttt{CPA}}_{L}\cdot \mathbf{V}_{HS}
\end{equation}

\begin{equation}
\label{eq:res_cross}
    \texttt{CPA}\left ( \mathbf{F}_{L}, \mathbf{F}_{HS} \right ) = \mathbf{F}_{L} + \gamma \mathbf{F}_{L}^\texttt{CPA}
\end{equation}

The computation of fused features for the hyperspectral branch $\texttt{CPA}\left ( \mathbf{F}_{HS}, \mathbf{F}_{L} \right )$ is simply done by swapping the index L and HS in Eq. \ref{eq:cross_attn_map}, \ref{eq:cross_attn}, and \ref{eq:res_cross}. Finally, the pairwise fusion $\mathbf{F}_{\text{fuse}}$ involves a bidirectonal cross-attention mechanism. This allows the lidar branch to attend hyperspectral branch to learn spectral information, and the hyperspectral branch to attend lidar branch to learn geometric details. The fused features from the bidirectional cross-attention mechanism of both branches can be computed as referenced in Eq. \ref{eq:cpa_all}.

Despite recent studies have incorporating the cross-attention mechanism to learn features from the fusion of lidar and hyperspectral \cite{10153685, 10462184}, our method differs significantly in that our cross-attention is built within a fully 3D pipeline, unlike their reliance on the commonly used 2D-fusion approach. Our method also utilizes a bidirectional cross-attention mechanism, instead of unidirectional as presented in these studies, where only the lidar branch attends to the hyperspectral branch. Additionally, our cross-attention mechanism facilitates feature fusion across multiple scales, whereas these methods perform feature fusion at a single scale.

%\section{Experimental Results and Discussion} 
\section{Experimental Results} 
\label{sec:exp}

\subsection{2018 IEEE Data Fusion Contest dataset}
\label{sec:grss}

\textbf{Description.} 2018 IEEE GRSS Data Fusion Contest (DFC2018) dataset \cite{8727489} focuses on multimodal remote sensing data, including:

\begin{itemize}
    \item 3D point clouds from the Optech Titan MW lidar sensor,
    \item hyperspectral data (1-m resolution) from an ITRES CASI 1500 sensor, and
    \item very high resolution RGB images (5-cm resolution) from a DiMAC ULTRALIGHT+ camera.
\end{itemize}

As our approach relies solely on 3D point clouds, we exclude lidar-derived raster products such as Digital Surface Models (DSMs) and Digital Terrain Models (DTMs). The point clouds, with density of 10-15 points/m$^{2}$, provide strong potential fusion for fusion with spectral data.

\begin{figure}[ht]
    \centering
    \includegraphics[clip=true, trim = 105 250 105 250, width=0.98\linewidth]{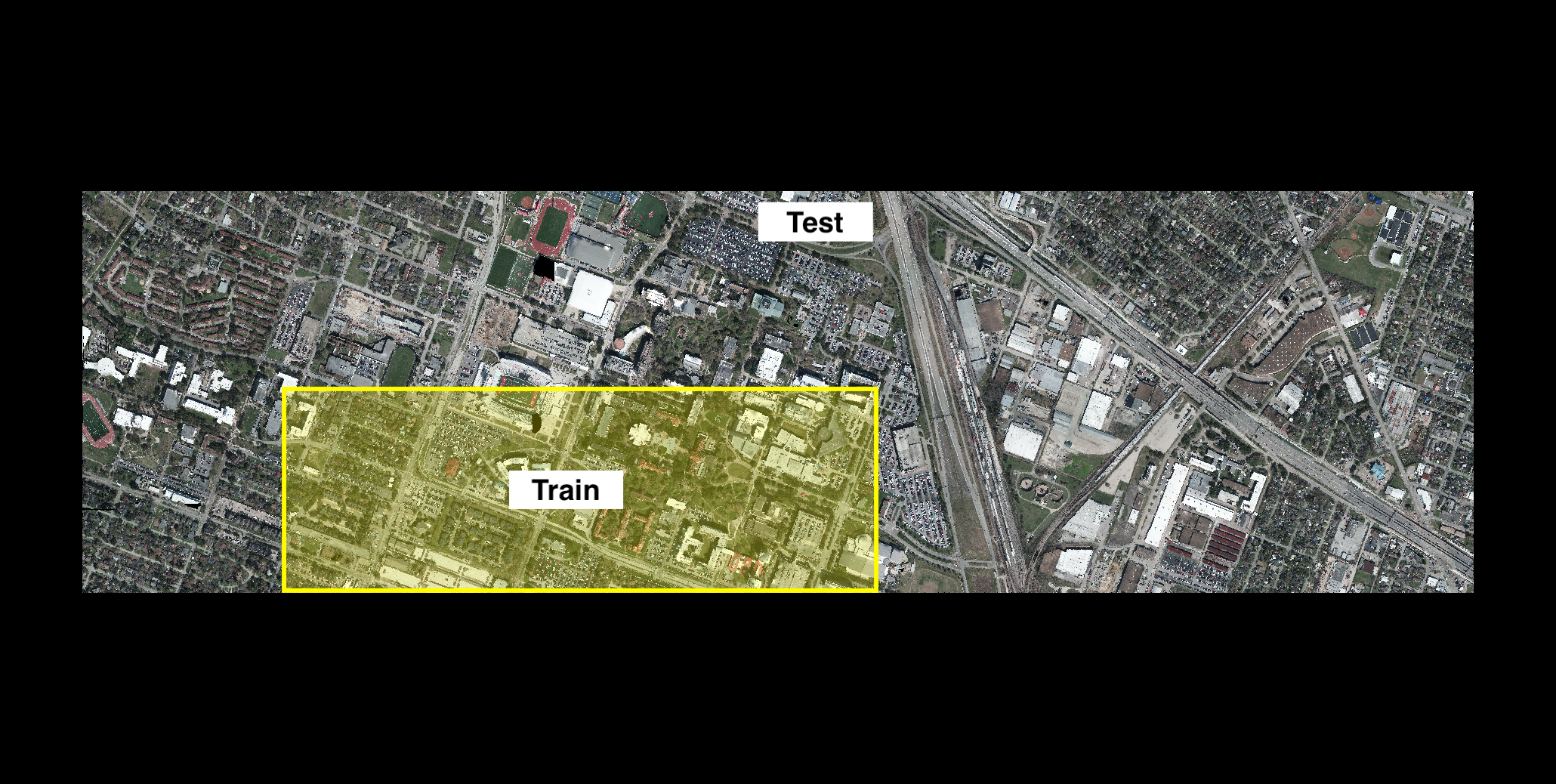}%trim - left, bottom, right, top
    \caption{The actual Train and Test split of the DFC2018 dataset.}
    \label{fig:dfc2018}
\end{figure}

The dataset consists of 20 classes, posing a challenge due to closely related subclasses, such as healthy and stressed grass, which form a continuum without clear separation. Ground truth is provided as a 2D map at 0.5m resolution. Table \ref{tab:dfc2018} presents the class distribution, including Class 0 (unlabeled). Unlike its predecessor, the IEEE 2013 Data Fusion Contest dataset, the DFC2018 dataset features completely separated Train and Test areas, illustrated in Fig. \ref{fig:dfc2018}.

%Due to our purely 3D point cloud-based approach, we exclude lidar-derived products such as raster Digital Surface Models (DSMs) and Digital Terrain Models (DTMs) data. The point clouds have a high point density at approximately 10-15 points/m$^{2}$, offer excellent potential for fusion with the spectral data. 

%The DFC2018 dataset consists of 20 classes, presenting a challenge due to the presence of classes that can be considered as sub-classes covering a continuum, such as healthy grass and stressed grass which represent different stages of the same class with no clear separation. 

%The ground truth is provided as a 2D map at 0.5m spatial sampling. The statistics of the 20-class labels are provided in Table \ref{tab:dfc2018} with an addition of Class 0 as unlabeled or background label. In terms of the Train and Test split, unlike its predecessor, the IEEE 2013 Data Fusion Contest dataset, the DFC2018 dataset is divided into fully separated Train and Test areas, as illustrated in Fig. \ref{fig:dfc2018}. 

\begin{table}[ht]
    \centering
    \caption{Semantic Classes of The DFC2018 Dataset}
    \label{tab:dfc2018}
    \begin{tabular}{llll}
        \hline
        \textbf{No} & \textbf{Class} & \multicolumn{2}{c}{\textbf{Number of samples}} \\
        & & \textbf{Train} & \textbf{Test} \\
        \hline
        \textbf{1} & Healthy grass & 39196 & 20000 \\
        \textbf{2} & Stressed grass & 130008 & 20000 \\
        \textbf{3} & Artificial turf & 2736 & 20000 \\
        \textbf{4} & Evergreen trees & 54322 & 20000 \\
        \textbf{5} & Deciduous trees & 20172 & 20000 \\
        \textbf{6} & Bare earth & 18064 & 20000 \\
        \textbf{7} & Water & 1064 & 20000 \\
        \textbf{8} & Residential buildings & 158995 & 1628 \\
        \textbf{9} & Non-residential buildings & 894769 & 20000 \\
        \textbf{10} & Roads & 183283 & 20000 \\
        \textbf{11} & Sidewalks & 136035 & 20000 \\
        \textbf{12} & Crosswalks & 6059 & 5345 \\
        \textbf{13} & Major thoroughfares & 185438 & 20000 \\
        \textbf{14} & Highways & 39438 & 20000 \\
        \textbf{15} & Railways & 27748 & 11232 \\
        \textbf{16} & Paved parking lots & 45932 & 20000 \\
        \textbf{17} & Unpaved parking lots & 587 & 3524 \\
        \textbf{18} & Cars & 26289 & 20000 \\
        \textbf{19} & Trains & 21479 & 20000 \\
        \textbf{20} & Stadium seats & 27296 & 20000 \\
        \hline
    \end{tabular}
\end{table}

Our experimental pipeline consists of:
%The general steps of our experiments using DFC2018 dataset are as follows and will be briefly described in the subsequent section.
\begin{enumerate}
    \item Generation of multimodal 3D point cloud data.
    \item 3D labels preparation.
    \item 3D evaluation.
    \item 2D evaluation.
    %\item Generation of multimodal 3D point cloud data: Projecting spectral information from 2D data to 3D point clouds.
    %\item 3D labels preparation: Projecting 2D labels to 3D points on the Train split and manually refining the labels.
    %\item 3D evaluation: Training the models and evaluating with 2-fold cross validation.
    %\item 2D evaluation: Training the models using 3D multimodal point cloud data on the Train split, then projecting 3D predictions on the Test split to 2D maps.
\end{enumerate}

In addition, we did not evaluate or investigate whether any mismatch between modalities exists. Since the dataset serves as benchmark data for data fusion, we assume that all modalities within the dataset have been carefully processed during georeferencing to ensure they are perfectly aligned.

\textbf{Generation of multimodal 3D point cloud data.} Generating multimodal 3D point cloud data by fusing 3D points with image data from other sensors can be approached in two distinct manners. The first involves associating spectral values of pixels with the points that belong to them. However, this method suffers from the irregularity and unordered nature of point cloud data, requiring an exhaustive search to establish the pixel-to-point relationship. 

The second approach, which we prefer due to its efficiency, involves finding the nearest pixel for each 3D point. Inspired by \cite{10347274, THIELE2021104252}, we converted 2D images to 2D points with their associated attributes, then utilized a kd-Tree to find the k-nearest neighbors (k-NN) of the 3D points to the 2D points with $k = 1$. Subsequently, we assigned the attributes of 2D points to their corresponding 3D points. This process results in 3D points enriched with hyperspectral and RGB information as additional attributes. Furthermore, this approach facilitates fusion without the need to upsample or downsample the hyperspectral and RGB images, instead relying on finding the nearest pixels to the 3D points. 

\begin{figure}[!t]
\centering
% \subfloat[]{\includegraphics[clip=true, trim = 200 400 200 200, width=0.98\linewidth]{img/DFC2018/multimodalpc/rgb.png}%trim - left, bottom, right, top
\subfloat[]{\includegraphics[clip=true, trim = 50 100 50 100, width=0.98\linewidth]{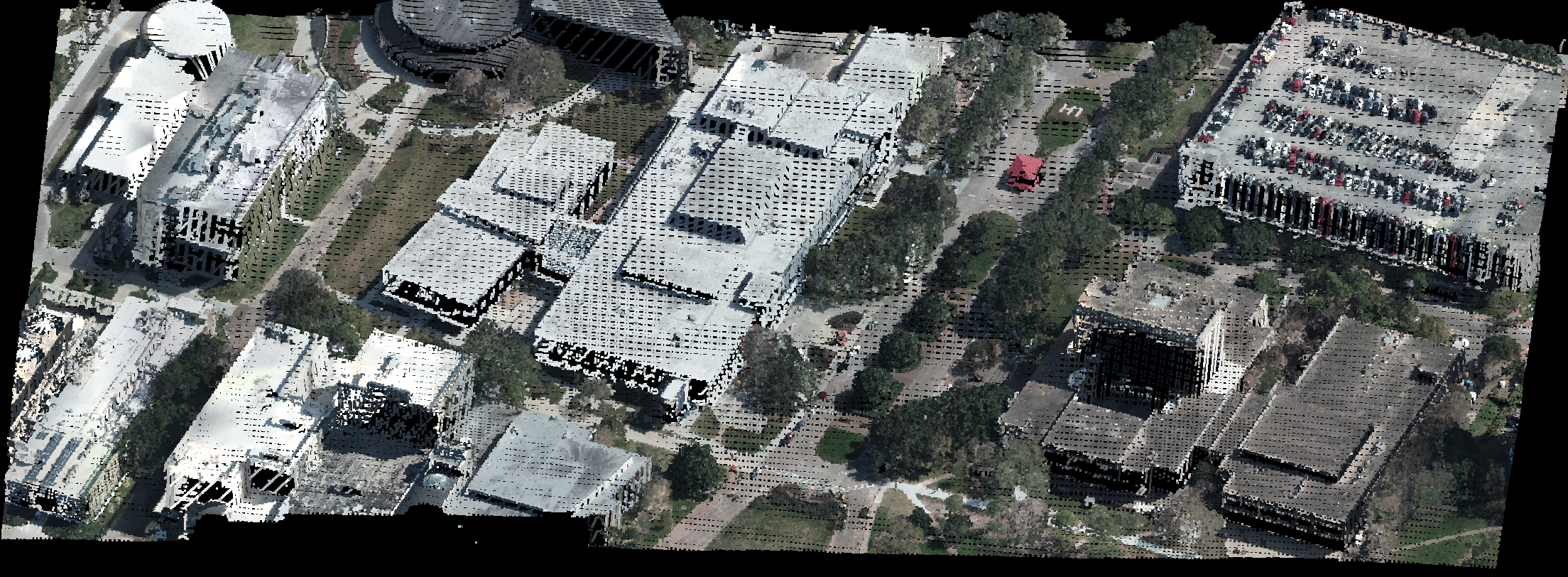}%trim - left, bottom, right, top
\label{fig:multimodalpc_rgb}}
\hfil
% \subfloat[]{\includegraphics[clip=true, trim = 200 400 200 200, width=0.98\linewidth]{img/DFC2018/multimodalpc/hs.png}%trim - left, bottom, right, top
\subfloat[]{\includegraphics[clip=true, trim = 50 100 50 100, width=0.98\linewidth]{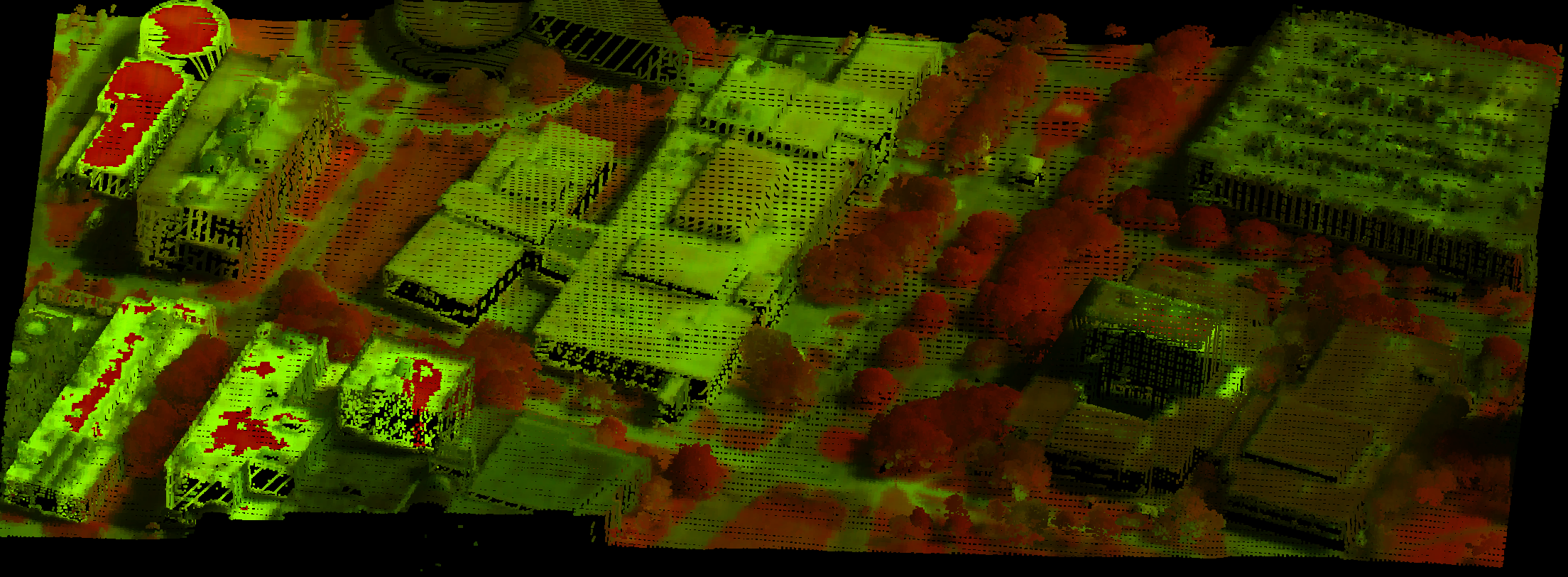}%trim - left, bottom, right, top
\label{fig:multimodalpc_hs}}
\hfil
% \subfloat[]{\includegraphics[clip=true, trim = 200 400 200 200, width=0.98\linewidth]{img/DFC2018/multimodalpc/lidar.png}%trim - left, bottom, right, top
\subfloat[]{\includegraphics[clip=true, trim = 50 100 50 100, width=0.98\linewidth]{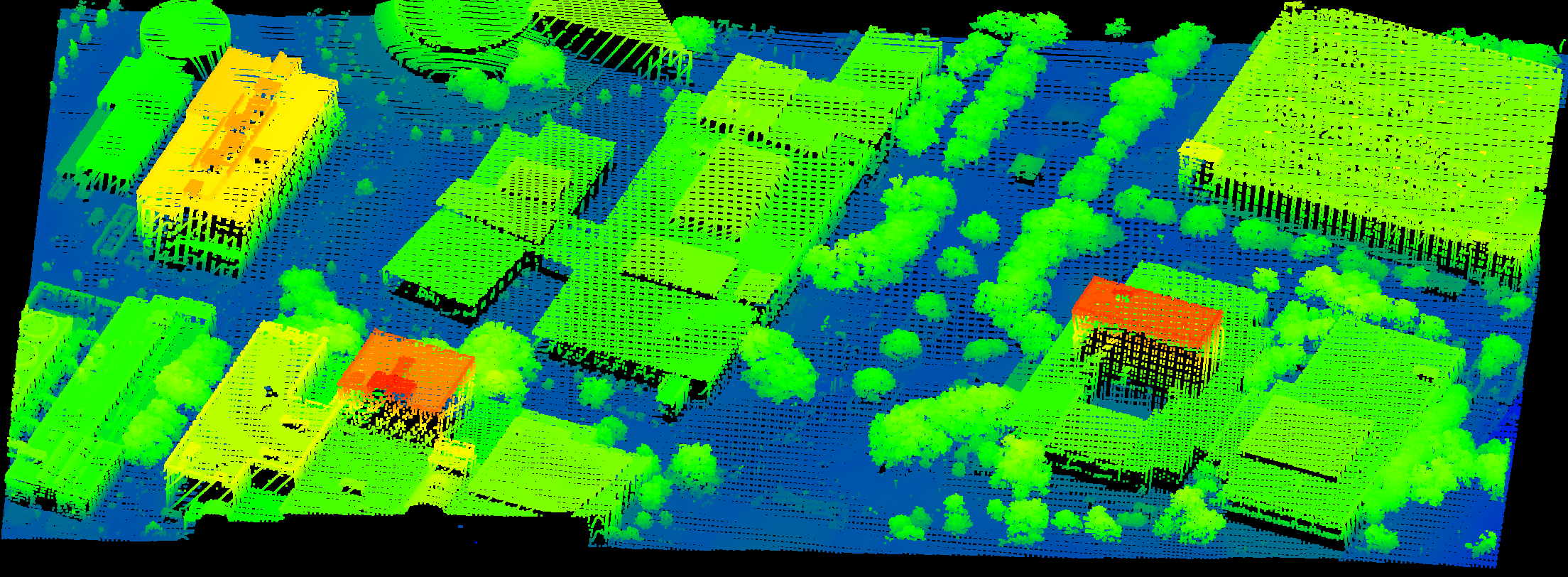}%trim - left, bottom, right, top
\label{fig:multimodalpc_lidar}}
\caption{Multimodal 3D point clouds data, rendered with different features: (a) RGB, (b) hyperspectral, and (c) elevation.}
\label{fig:multimodalpc}
\end{figure}

Fig. \ref{fig:multimodalpc} shows the generated multimodal 3D point clouds rendered with RGB, hyperspectral and elevation attributes.

\begin{figure}[h]
    \centering
    \includegraphics[clip=true, trim = 0 0 0 0, width=0.98\linewidth]{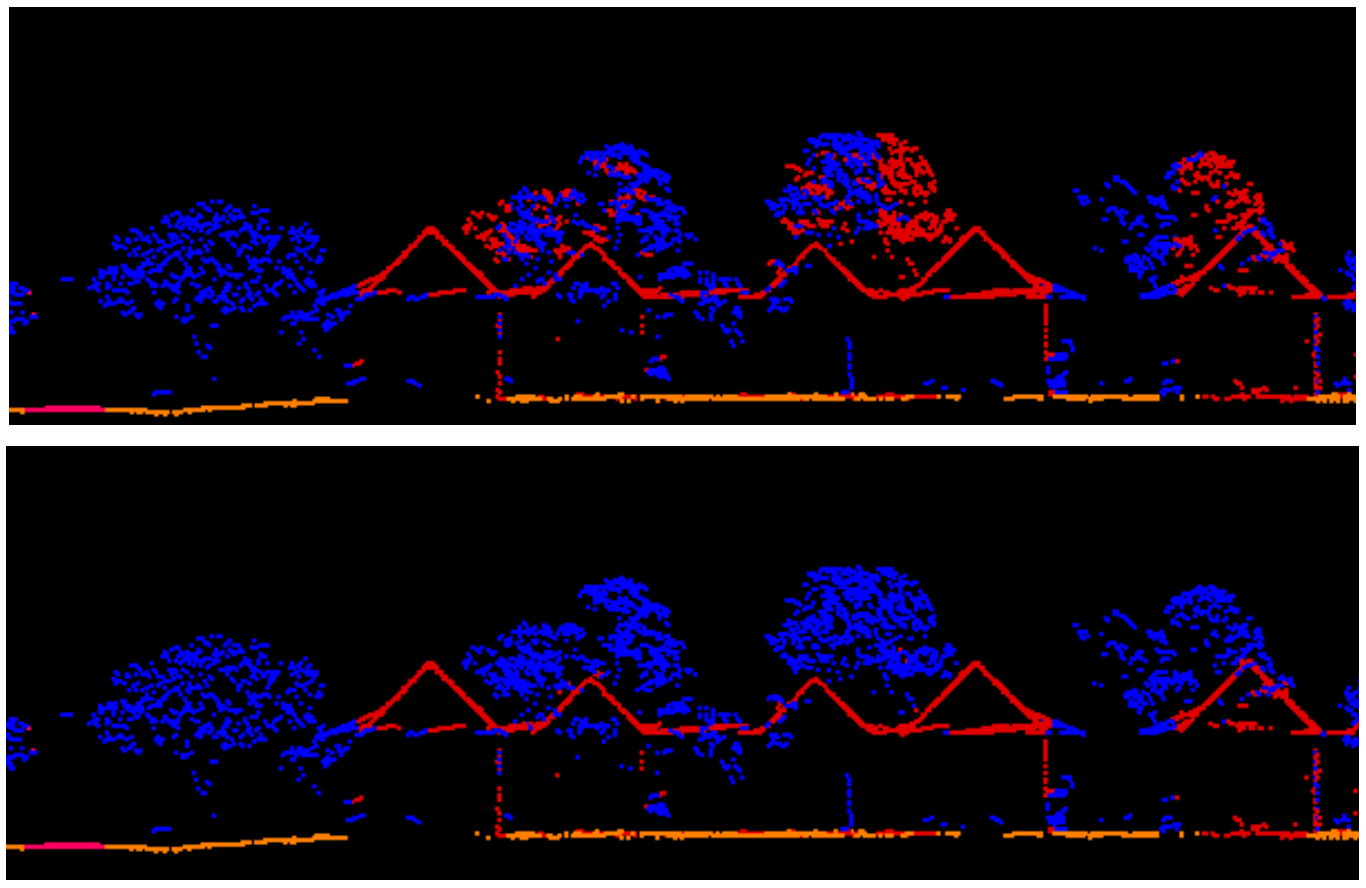}%trim - left, bottom, right, top
    \caption{The result after manual relabeling on the Train split. Top: errors on some points due to 2D to 3D labels projection. Bottom: corrected labels.}
    \label{fig:manual_label}
\end{figure}

\textbf{3D labels preparation.} The original ground truth was provided in the form of a 2D map. To train the models effectively, the labels needed to be transferred from the 2D map to the corresponding 3D points. Initially, we employed a straightforward approach by projecting the labels using a nadir projection. While this method successfully labeled many points in open areas, it resulted in numerous errors, particularly in problematic areas such as sections where tree branches overlapped with building roofs, as illustrated in Fig. \ref{fig:manual_label}. These errors on the Train split adversely affected the performance of the models on the Test split. We observed that the 2D labels were not perfectly aligned when transferred to the 3D scenes. To address these issues, we meticulously edited the labels to ensure accuracy. However, we only refined the 3D labels on the Train split, as dense ground truth data was available only for this subset of the dataset. 

\begin{figure}[h]
    \centering
    \includegraphics[clip=true, trim = 40 50 40 50, width=0.98\linewidth]{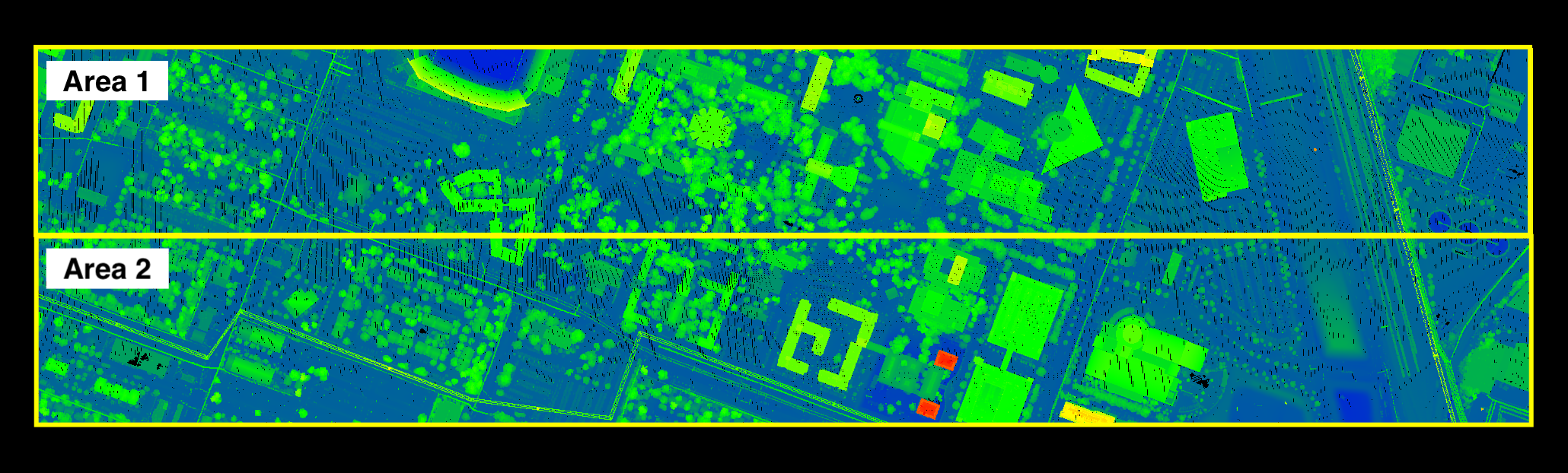}%trim - left, bottom, right, top
    \caption{2-fold areas for evaluation of 3D multimodal point clouds.}
    \label{fig:2fold}
\end{figure}

\textbf{3D evaluation. } We assessed the performance of 3D approaches using 2-fold cross-validation on the Train split. We partitioned the Train split into Fold-1 and Fold-2 areas, as illustrated in Fig. \ref{fig:2fold}. The predictions from each fold were combined to calculate the final accuracy scores. We employed per-class F1-score accuracy, as well as Overall Accuracy (OA), mean F1-score, mean Precision, mean Recall, and mean IoU metrics.

Please note that the 2-fold areas are divided in this manner to ensure the availability of all classes in each fold. However, two classes only appear in one fold due to the location in the case of stadium seat class and limited number of samples in the case of artificial turf class. Therefore, we excluded those classes during the evaluation.

\textbf{2D evaluation.} To enable comparison with 2D approaches, we evaluated several 3D models, including ours, on the actual Test split of the DFC2018 dataset. The models were trained using 3D multimodal point cloud data from the Train split, and evaluated on the actual Test split to generate 3D prediction labels.

Additionally, we projected the 3D prediction labels onto 2D maps not only to obtain an accuracy score in the Test split but also for comparison with the winners of DFC2018 competition and other recent deep learning models for remote sensing data fusion that utilized 2D approaches. While this may seem like a straightforward projection, we encountered challenges that introduced errors into the 2D maps.

Initially, we projected 3D labels onto a 2D plane by determining the nearest 3D labels for each 2D pixel. To mitigate the influence of object height, the search only considered XY coordinates. However, in vegetated areas where lidar penetrates, numerous pixels become mixed between ground classes (such as grass) and non-ground classes (such as trees). We believe that the ground truth for the competition was annotated based on 2D remote sensing images captured from a nadir view. Consequently, non-ground objects are consistently depicted above ground objects in these annotations. However, our 2D prediction map derived from the straightforward projection does not consistently align with that.

To address this issue, we executed the projection of 3D labels in two stages. First, we projected all classes belonging to the ground classes to generate a base prediction map. Subsequently, we projected the remaining classes, which belong to the non-ground classes, on top of the ground classes to replicate the nadir view of the remote sensing images.

Using this approach we generate a homogeneous dataset truly representing the 3D nature of the objects in the scene.

For the 2D evaluation, we employed OA and Kappa value for the metric accuracy according to the DFC2018 competition. 

\textbf{Experimental setup. } Prior training, the dataset is partitioned into blocks with a fixed size of 75x75 m and a stride size of 25 m. Following common practice in point cloud processing, points within each block are randomly sampled to create a subset of 4096 points. Afterwards, the blocks are split with 90\% allocated to the training set and 10\% to the validation set. This preprocessing procedure is applied to all models except for KPConv, for which we follow to its original implementation by creating spheres. For KPConv, we chose a radius of 75 m for the input sphere and 0.5 m for the first subsampling grid parameters.

Subsequently, standard data normalization is applied to each modality to ensure consistency and compatibility during feature encoding. For 2D images, such as hyperspectral data, a standard normalization technique is used to scale pixel intensity values to the range [0, 1]. For 3D point clouds, common normalization practices are followed, such as converting absolute coordinates to relative coordinates by subtracting the minimum and maximum coordinate values and then centering the point cloud based on its geometric center. This ensures that the spatial and spectral features are appropriately scaled and aligned for downstream processing.

We implemented our neural network using PyTorch and utilized the Adam optimizer with a commonly used learning rate of 0.001. A batch size of 16 was used for training, and the model was trained using 4 Nvidia A100 GPUs. We trained the model for 100 epochs and selected the model with the highest mean IoU on the validation set for predicting the Test split.  

\textbf{3D quantitative and qualitative results. } We conducted a thorough evaluation of our method alongside various deep learning models for 3D point cloud data, including PointNet \cite{pointnet}, PointNet++ \cite{pointnet2}, DGCNN \cite{dgcnn}, KPConv \cite{kpconv}, SpUnet \cite{8579059}, Point Transformer \cite{point_transformer}, OctFormer \cite{10.1145/3592131}, Point Transformer V2 \cite{NEURIPS2022_d78ece66}, and Point Transformer V3 \cite{10658198}.

Although these models were not specifically designed to fuse point clouds with complex spectral features, such as hyperspectral data, they are commonly used with RGB as additional point features alongside point coordinates. This flexibility allows them to incorporate other spectral features, including hyperspectral data. Therefore, we included spectral features (hyperspectral and RGB) as point features in these models. This highlights the gap in research, as few studies focus on developing dedicated fusion models that integrate point coordinates with complex spectral features. Additionally, we compared our results with L3D \cite{decker2023hyperspectral}, a KP-Conv-based model specifically designed for data fusion. It is worth noting that we interpreted the implementation of L3D, as the authors did not make the code publicly available.

%All modalities, including XYZ coordinates, hyperspectral, and RGB features, are stacked as input features for these models. 

%PointNet and PointNet++ were implemented using the PyTorch version available at \url{https://github.com/yanx27/Pointnet_Pointnet2_pytorch}. DGCNN was implemented using code from \url{https://github.com/antao97/dgcnn.pytorch}. We utilized the PyTorch version of KPConv from \url{https://github.com/HuguesTHOMAS/KPConv-PyTorch}. Lastly, we used \url{https://github.com/qq456cvb/Point-Transformers} for PCT and Point Transformer. 

In Table \ref{tab:3D_accuracy}, we present the impressive results achieved by HyperPointFormer compared to various deep learning models for 3D point clouds. Relative to our baseline, Point Transformer, we observed an improvement of 4.5 percentage points in the mean F1 score and 2.0 percentage points in the mean IoU. We also achieved a significant improvement over the L3D model, which is specifically designed for the fusion of lidar and hyperspectral data.

\begin{table*}[!h]
    \centering
    \caption{Quantitative Results of 3D Evaluation Using 2-fold Cross Validation. The best results are highlighted in bold and the second-best results are underlined.}
    \label{tab:3D_accuracy}
    %\begin{tabular}{p{0.3cm}p{2.2cm}p{0.9cm}p{1cm}p{0.9cm}p{0.9cm}p{0.9cm}p{0.8cm}p{0.8cm}p{1.2cm}p{0.8cm}p{0.8cm}p{0.8cm}}
    \begin{tabular}{p{0.3cm}p{2.2cm}>{\raggedleft\arraybackslash}p{0.9cm}>{\raggedleft\arraybackslash}p{1cm}>{\raggedleft\arraybackslash}p{0.9cm}>{\raggedleft\arraybackslash}p{0.9cm}>{\raggedleft\arraybackslash}p{0.9cm}>{\raggedleft\arraybackslash}p{0.8cm}>{\raggedleft\arraybackslash}p{0.8cm}>{\raggedleft\arraybackslash}p{1.2cm}>{\raggedleft\arraybackslash}p{0.8cm}>{\raggedleft\arraybackslash}p{0.8cm}>{\raggedleft\arraybackslash}p{0.8cm}}
        \hline
        \textbf{No} & \textbf{Class} & \textbf{PointNet} & \textbf{PointNet++} & \textbf{DGCNN} & \textbf{KPConv} & \textbf{L3D} & \textbf{SpUnet} & \textbf{PT} & \textbf{OctFormer} & \textbf{PTv2} & \textbf{PTv3} & \textbf{Ours} \\
         & & \cite{pointnet} & \cite{pointnet2} & \cite{dgcnn} & \cite{kpconv} & \cite{decker2023hyperspectral} & \cite{8579059} & \cite{point_transformer} & \cite{10.1145/3592131} & \cite{NEURIPS2022_d78ece66} & \cite{10658198} & \\
        \hline
        \textbf{1} & Healthy grass & 47.83 & \textbf{78.44} & \underline{68.70} & 13.42 & 16.43 & 36.99 & 40.78 & 51.77 & 62.18 & 63.59 & 63.46 \\
        \textbf{2} & Stressed grass & 63.03 & \textbf{80.45} & \underline{72.96} & 12.72 & 17.20 & 44.56 & 68.02 & 66.10 & 47.69 & 50.23 & 71.29 \\
        \textbf{3} & Artificial turf & -- & -- & -- & -- & -- & -- & -- & -- & -- & -- & -- \\
        \textbf{4} & Evergreen trees & 68.84 & 83.28 & 91.71 & 66.79 & 79.26 & 93.83 & \underline{91.75} & 88.53 & 91.65 & 91.97 & \textbf{95.58} \\
        \textbf{5} & Deciduous trees & 23.25 & 27.63 & 64.19 & 42.86 & 58.89 & \textbf{79.41} & 70.24 & 57.27 & 76.10 & 76.97 & \underline{78.84} \\
        \textbf{6} & Bare earth & 10.49 & 0.00 & 16.52 & 11.23 & 0.00 & 0.00 & 32.48 & \underline{39.18} & 28.89 & 33.00 & \textbf{61.44} \\
        \textbf{7} & Water & 0.00 & 0.00 & 0.00 & 0.00 & 0.00 & 0.00 & 0.00 & 0.00 & 0.00 & 0.00 & 0.00 \\
        \textbf{8} & Res. buildings & 53.22 & 64.70 & \textbf{84.09} & 61.49 & 65.32 & 57.98 & 69.56 & \underline{80.43} & 68.32 & 69.88 & 54.99 \\
        \textbf{9} & Non-res. buildings & 90.64 & 92.83 & \textbf{95.92} & 84.08 & 87.88 & 77.35 & \underline{95.50} & 91.79 & 84.81 & 85.32 & 94.16 \\
        \textbf{10} & Roads & 40.51 & 54.23 & 52.66 & 41.44 & \textbf{63.96} & 38.35 & \underline{60.27} & 46.28 & 53.15 & 55.16 & 56.27 \\
        \textbf{11} & Sidewalks & 54.75 & \underline{68.45} & 60.81 & 46.54 & 62.73 & 31.73 & \textbf{68.66} & 67.95 & 52.80 & 54.75 & \underline{68.45} \\
        \textbf{12} & Crosswalks & \underline{3.23} & 0.00 & \textbf{4.99} & 0.00 & 0.00 & 0.00 & 0.00 & 0.00 & 0.00 & 0.00 & 0.00 \\
        \textbf{13} & Major thorough. & 31.94 & 55.44 & 35.71 & 40.04 & 64.33 & 2.33 & 58.79 & \underline{59.84} & 37.54 & 41.26 & \textbf{61.58} \\
        \textbf{14} & Highways & 37.32 & 43.26 & 53.86 & 45.20 & 43.17 & 34.18 & 31.38 & 22.97 & \underline{59.06} & \textbf{60.73} & 55.34 \\
        \textbf{15} & Railways & 15.71 & 36.39 & 23.49 & 28.82 & 27.32 & 00.05 & 05.37 & 26.88 & \underline{45.55} & \textbf{48.72} & 06.45 \\
        \textbf{16} & Paved parking lots & 06.89 & 26.85 & 39.22 & 79.81 & 81.31 & 8.55 & 80.17 & \textbf{84.22} & 71.66 & 72.99 & \underline{81.46} \\
        \textbf{17} & Unpaved parking l. & 0.00 & 0.00 & 0.00 & 0.00 & 0.00 & 0.00 & 0.00 & 0.00 & 0.00 & 0.00 & 0.00 \\
        \textbf{18} & Cars & 14.27 & 0.60 & 36.58 & 91.43 & 90.12 & 11.82 & \textbf{93.91} & \underline{92.96} & 84.93 & 85.55 & 89.40 \\
        \textbf{19} & Trains & 81.58 & 70.54 & 63.95 & 96.61 & 91.56 & 93.23 & 84.02 & 90.81 & \underline{95.92} & \textbf{96.08} & 93.80 \\
        \textbf{20} & Stadium seats & -- & -- & -- & -- & -- & -- & -- & -- & -- & -- & --\\
        \hline
         & F1 (\%) & 35.75 & 43.50 & 48.08 & 40.13 & 42.47 & 33.91 & 52.83 & 53.73 & 53.35 & \underline{54.79} & \textbf{57.36} \\
         & Precision (\%) & 50.95 & 46.64 & 61.23 & 48.72 & 48.61 & 52.50 & 60.58 & 61.87 & \underline{67.62} & \textbf{68.38} & 64.19 \\
         & Recall (\%) & 33.51 & 45.59 & 46.68 & 38.31 & 40.47 & 31.66 & 50.12 & \underline{51.94} & 48.36 & 49.69 & \textbf{55.43} \\
         & OA (\%) & 67.87 & 74.98 & 77.74 & 66.41 & 73.87 & 64.50 & \underline{79.88} & 76.21 & 73.94 & 74.95 & \textbf{79.91} \\
         & mIoU (\%) & 25.73 & 33.41 & 36.84 & 30.83 & 33.55 & 26.22 & \underline{44.87} & 43.10 & 41.84 & 43.22 & \textbf{46.94} \\
        \hline
    \end{tabular}
\end{table*}

\begin{figure*}[!t]
\centering
\subfloat[]{\includegraphics[clip=true, trim = 250 0 250 50, width=0.3\linewidth]{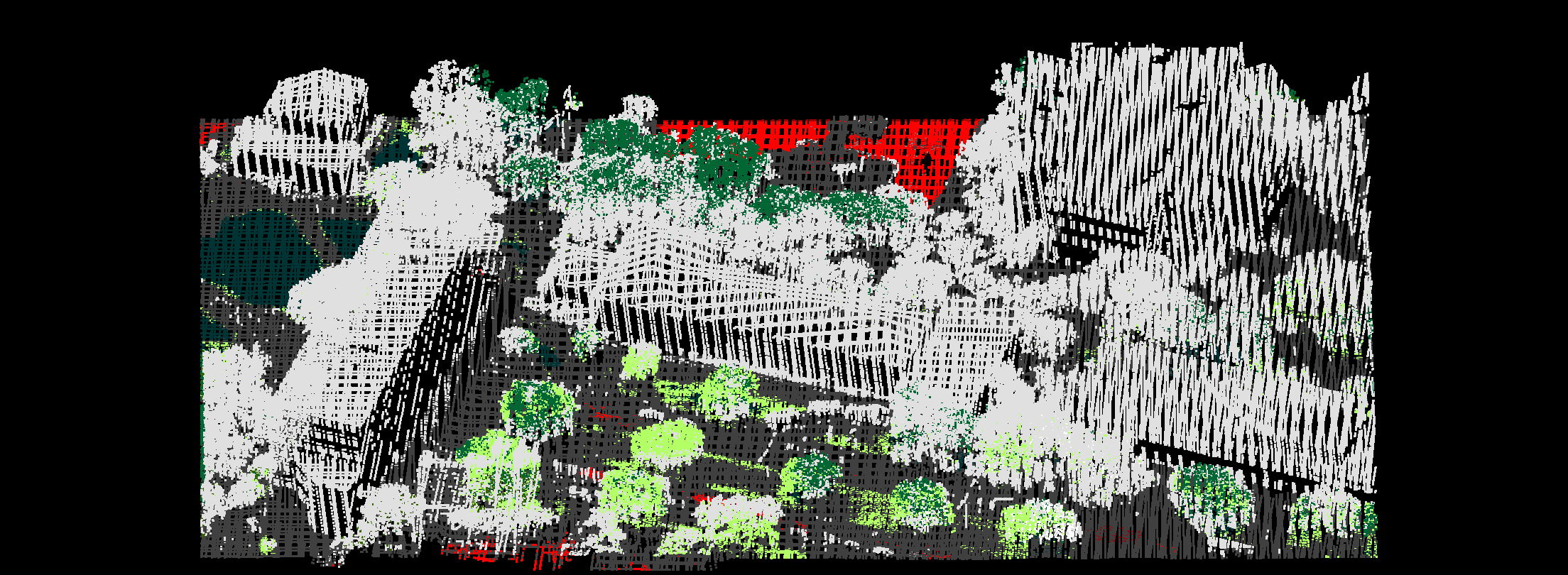}%trim - left, bottom, right, top
\label{pred2F_DFC2018_a}}
\hfil
\subfloat[]{\includegraphics[clip=true, trim = 250 0 250 50, width=0.3\linewidth]{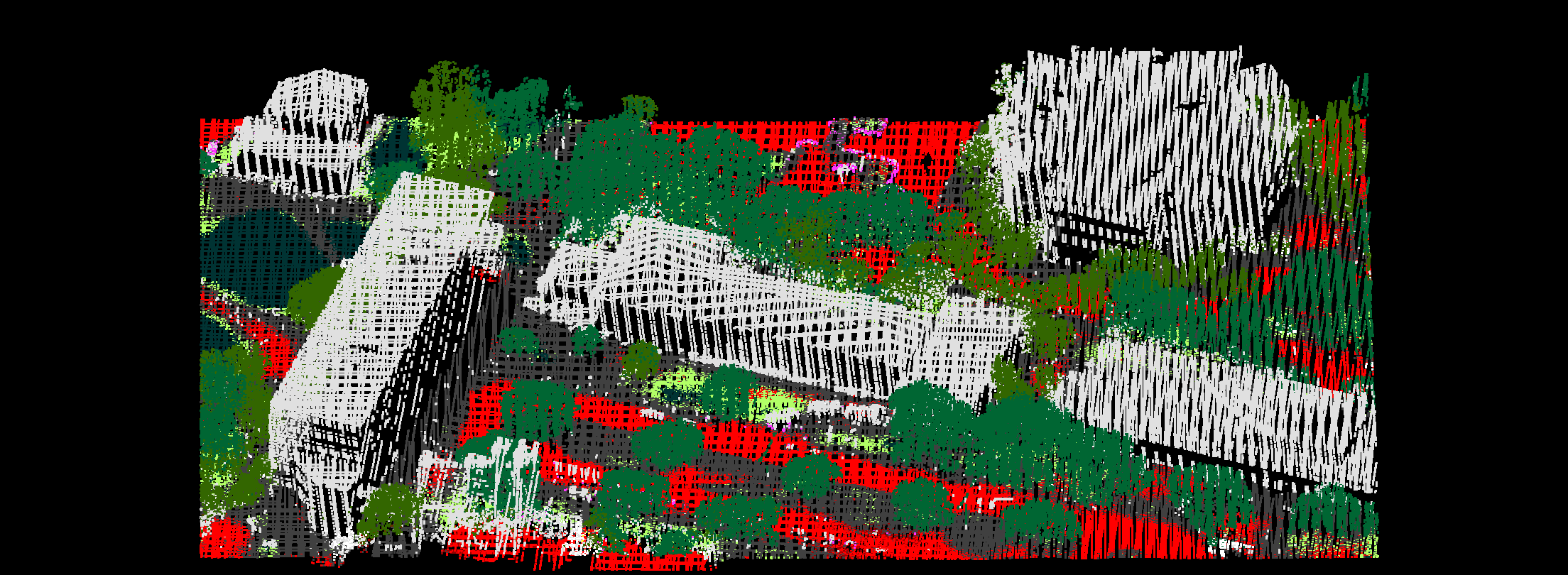}%trim - left, bottom, right, top
\label{pred2F_DFC2018_b}}
\hfil
\subfloat[]{\includegraphics[clip=true, trim = 250 0 250 50, width=0.3\linewidth]{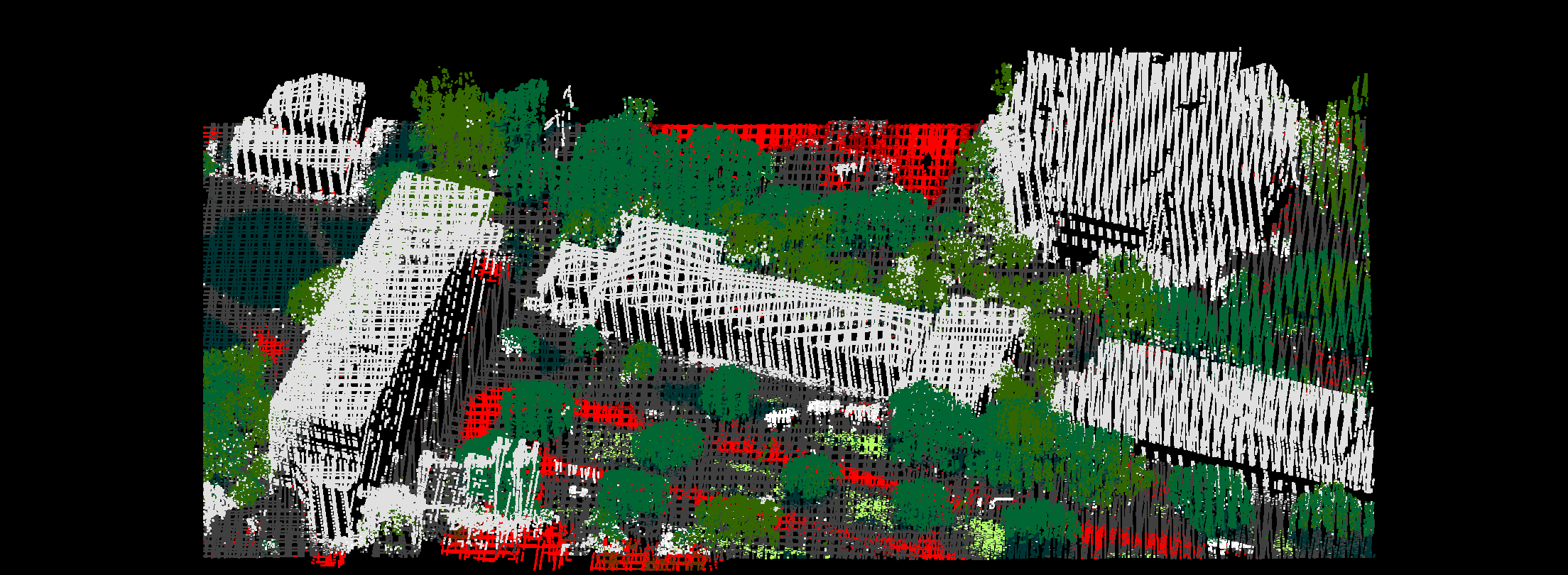}%trim - left, bottom, right, top
\label{pred2F_DFC2018_c}}
\hfil
\subfloat[]{\includegraphics[clip=true, trim = 250 0 250 50, width=0.3\linewidth]{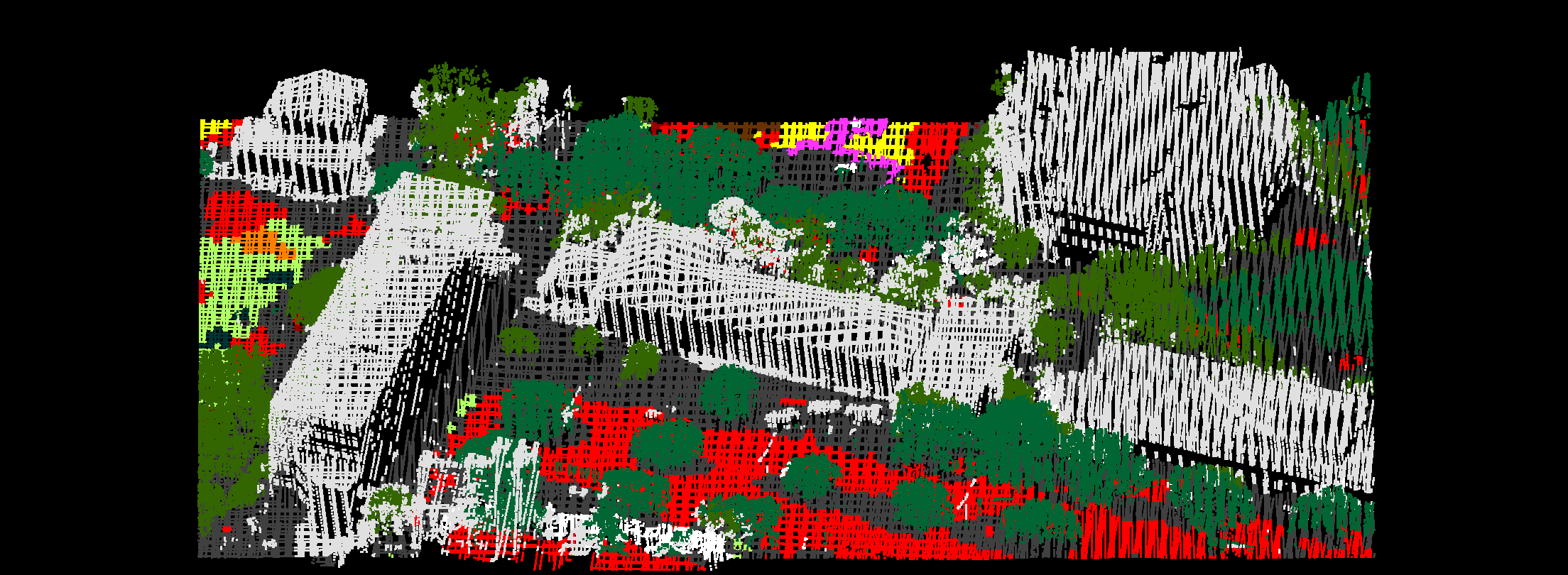}%trim - left, bottom, right, top
\label{pred2F_DFC2018_d}}
\hfil
\subfloat[]{\includegraphics[clip=true, trim = 250 0 250 50, width=0.3\linewidth]{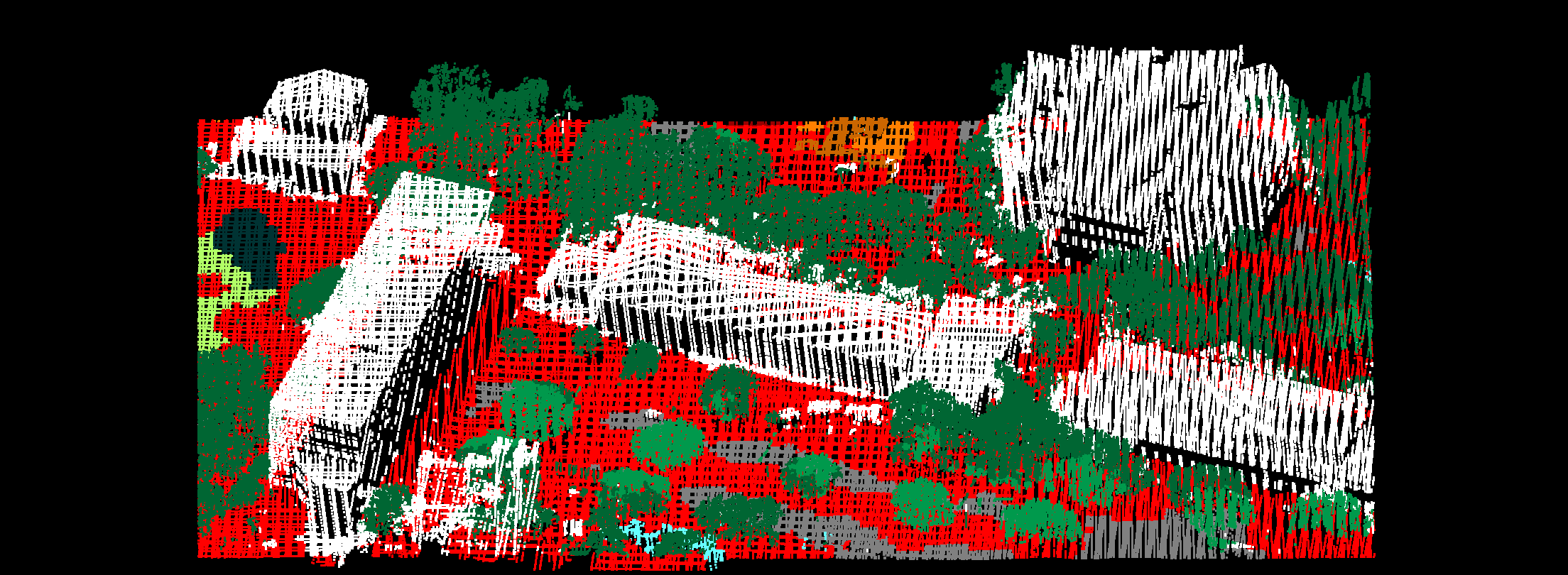}%trim - left, bottom, right, top
\label{pred2F_DFC2018_e}}
\hfil
\subfloat[]{\includegraphics[clip=true, trim = 250 0 250 50, width=0.3\linewidth]{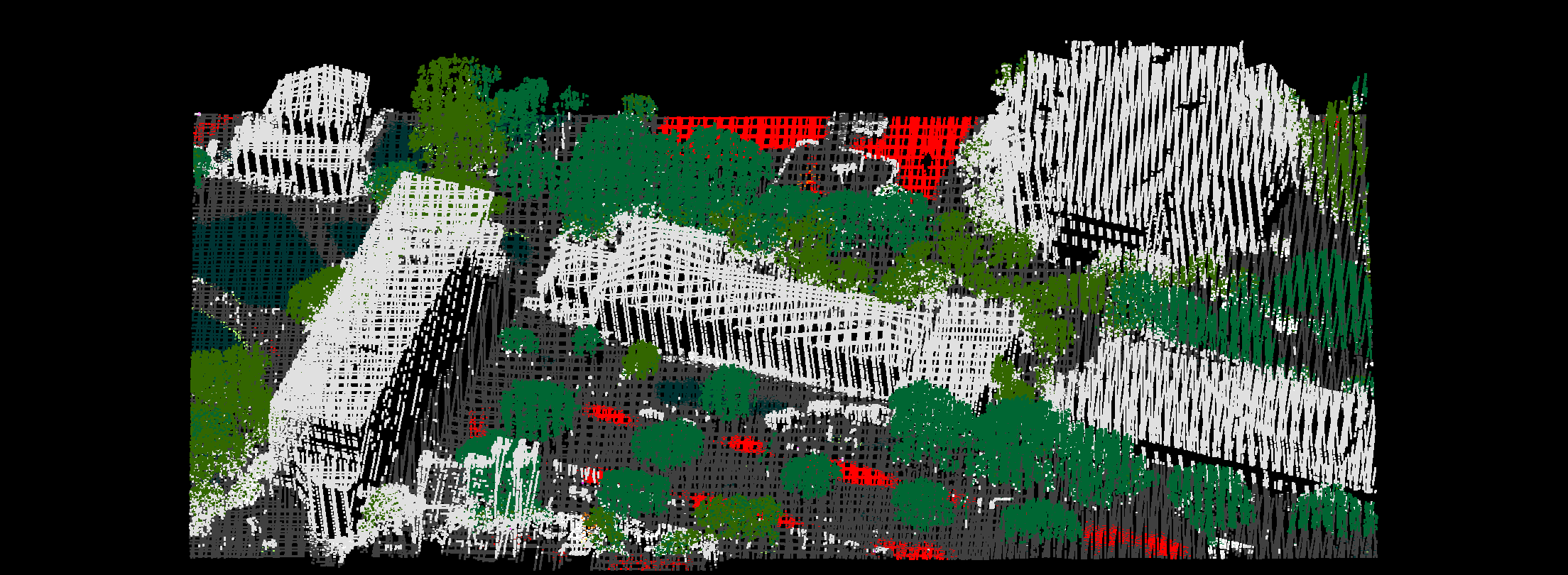}%trim - left, bottom, right, top
\label{pred2F_DFC2018_f}}
\hfil
\subfloat[]{\includegraphics[clip=true, trim = 250 0 250 50, width=0.3\linewidth]{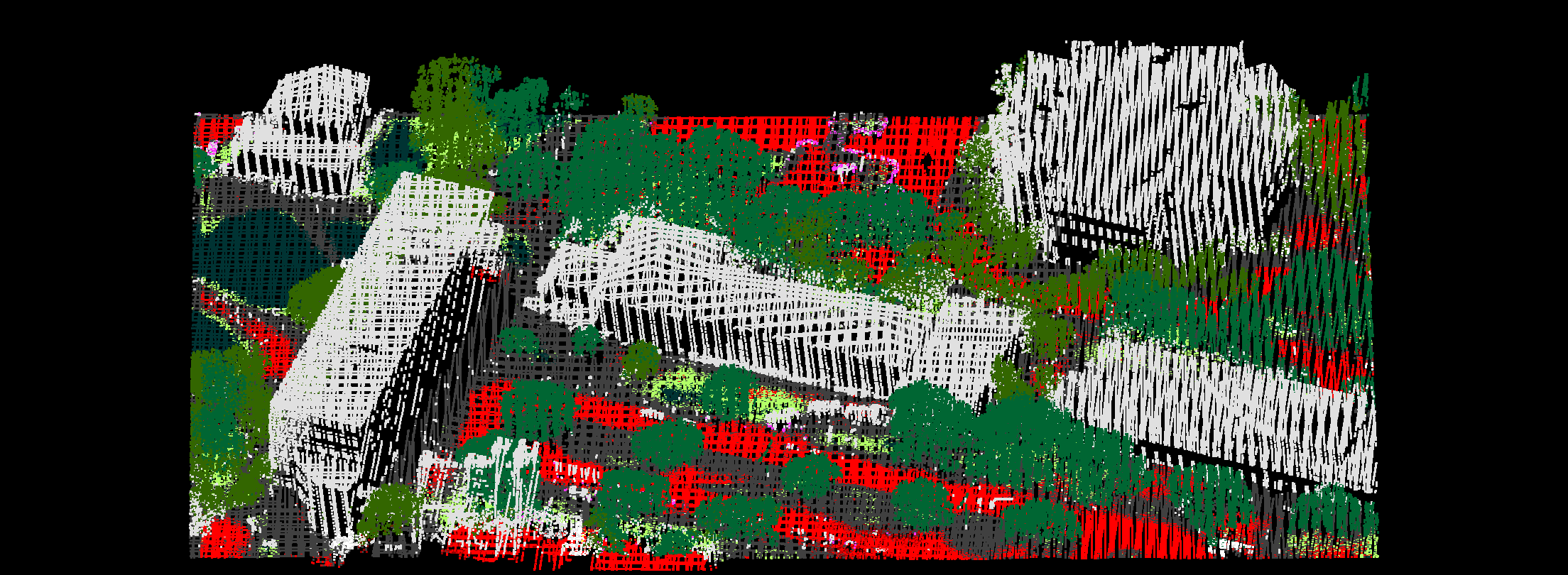}%trim - left, bottom, right, top
\label{pred2F_DFC2018_g}}
\hfil
\subfloat[]{\includegraphics[clip=true, trim = 250 0 250 50, width=0.3\linewidth]{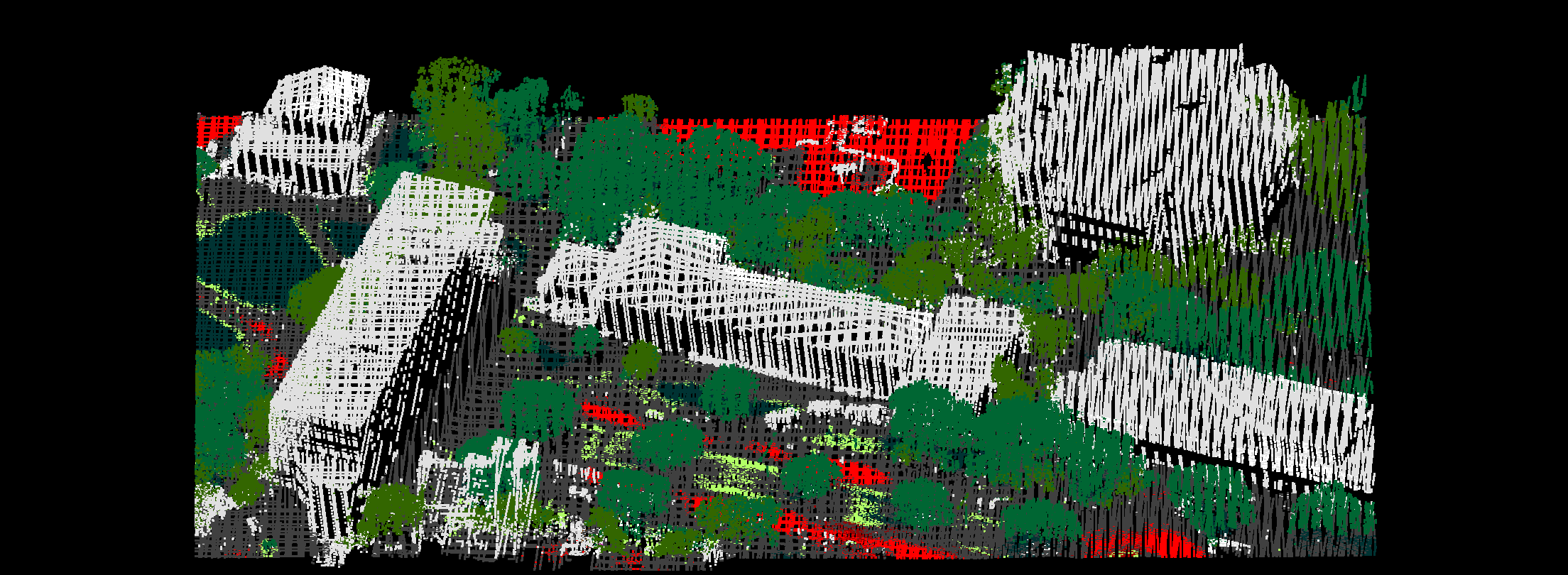}%trim - left, bottom, right, top
\label{pred2F_DFC2018_h}}
\hfil
\subfloat[]{\includegraphics[clip=true, trim = 250 0 250 50, width=0.3\linewidth]{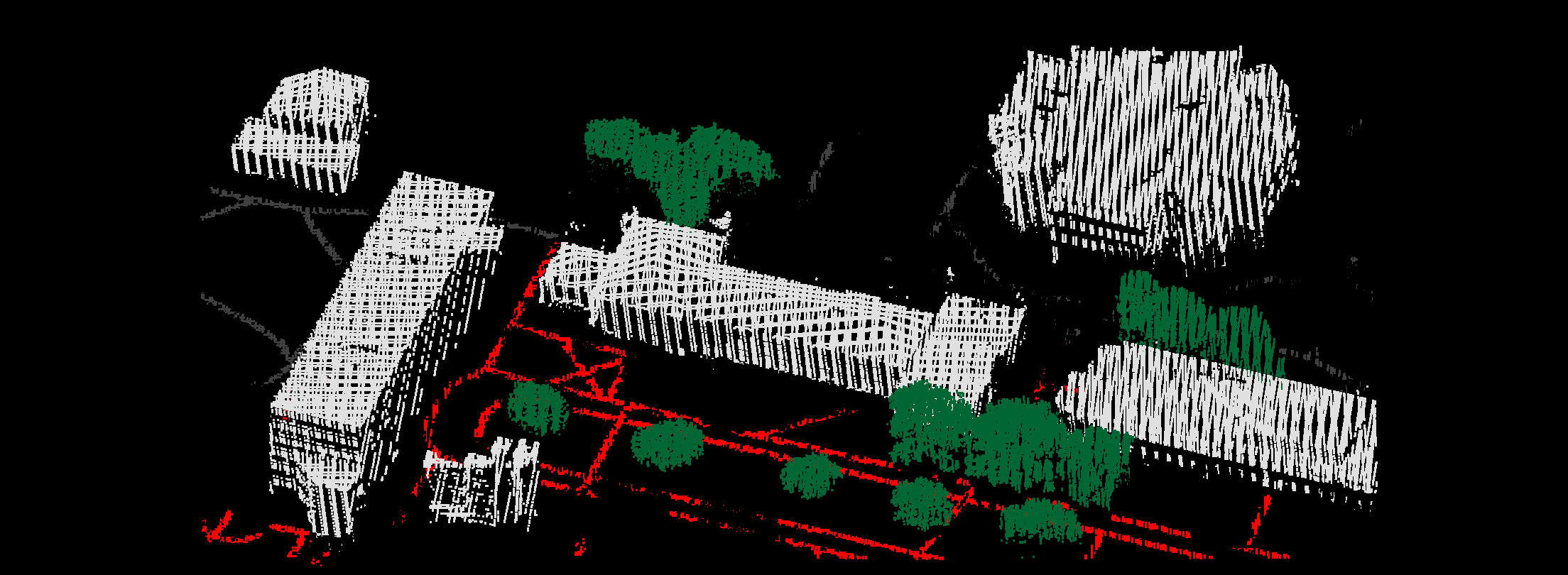}%trim - left, bottom, right, top
\label{pred2F_DFC2018_i}}
\hfil
\subfloat[]{\includegraphics[clip=true, trim = 250 0 250 50, width=0.3\linewidth]{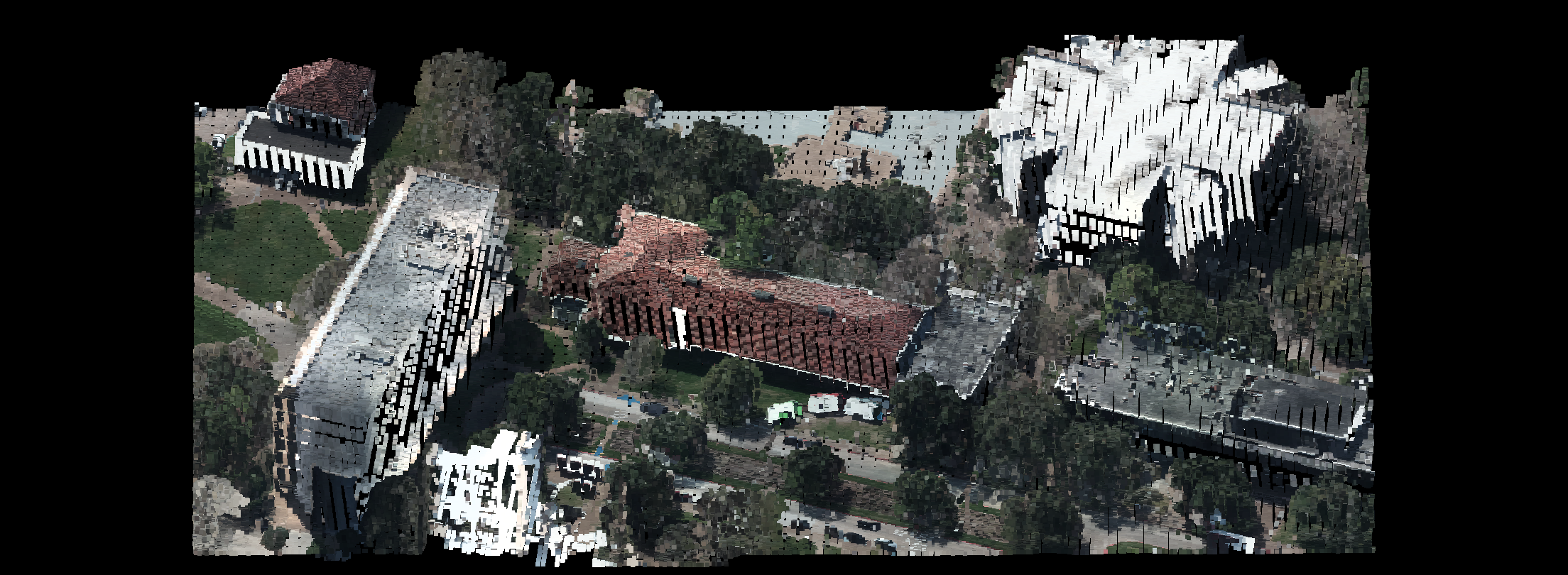}%trim - left, bottom, right, top
\label{pred2F_DFC2018_j}}
\caption{Visual comparison of 3D predictions from different models on the DFC2018 dataset. (a) PointNet. (b) PointNet++. (c) DGCNN. (d) KPConv. (e) L3D. (f) Point Transformer. (g) Point Transformer V3. (h) HyperPointFormer. (i) Ground truth. (j) RGB point clouds.}
\label{fig:pred2F_DFC2018}
\end{figure*}

Figure \ref{fig:pred2F_DFC2018} presents a comparison of 3D predictions from various models. Although a comprehensive qualitative analysis is somewhat challenging due to some points being unlabeled (Class 0), it is evident that Transformer-based models generally yield more realistic predictions. However, some points belonging to trees are misclassified as buildings in the top-right area of the results from Point Transformer and in the central area of the results from Point Transformer V3.

\textbf{2D quantitative and qualitative results. } While our 3D evaluation yielded satisfactory results, it is essential to compare our 3D framework with 2D fusion methods. To showcase its performance against 2D competitors, we assessed our approach using the actual Test split of the DFC2018 dataset. This enabled us to compare our method not only with the winners of the DFC2018 contest (CNN-NN and Fusion-FCN) but also with recent models specifically designed for remote sensing data fusion, such as DGConv \cite{9761218} and MFT \cite{10153685}. 

In Table \ref{tab:3D_2D_accuracy}, we showcase the competitive performance of our framework in generating 2D prediction maps, despite its original design for 3D point clouds. Accuracy figures for CNN-NN and Fusion-FCN, without post-classification for end-to-end learning comparison, were cited from \cite{8727489}, while Unet and Unet-DGConv from \cite{9761218}. 

To strengthen our comparisons and provide more comprehensive evaluation, we included experiments with recent Transformer-based fusion methods, such as  MFT \cite{10153685}, Cross-HL \cite{10462184}, SoftFormer \cite{LIU2024277}, and NCGLF$^{2}$ \cite{TU2024102192}. MFT and Cross-HL are recent Transformer-based models for hyperspectral and lidar data fusion with cross-attention. Their approach is similar to ours, but they preprocessed all modalities in a 2D format rather than 3D approach used in our manuscript. NCGLF$^{2}$ combines local and global attention mechanisms extracted from multi-scale features. Softformer, originally designed for fusing SAR and optical satellite images using a combination of CNNs and Transformers, can also be adapted for hyperspectral and lidar data fusion, making it relevant to our study.

From Table \ref{tab:3D_2D_accuracy}, it can be concluded that our method is highly competitive compared to other 2D-based Transformer approaches, particularly MFT,  Cross-HL, and SoftFormer. Our method outperformed all comparing methods in terms of OA and Kappa values. Note that the number of samples for the 2D evaluation is nearly perfectly balanced, with only a few exceptions, as shown in Table \ref{tab:dfc2018}. This indicates that the OA effectively reflects the balanced accuracy across all classes. Furthermore, our method demonstrated superior performance in predicting specific classes, such as highways and trains, highlighting its strength in handling certain structured features. However, it showed weaker predictions for ground-level classes, such as artificial turfs and grass, indicating a potential area of improvement.

%Additionally, we re-implemented MFT \cite{10153685} using its official GitHub repository due to the unavailability of accuracy numbers for the DFC2018 dataset.

\begin{figure*}[!t]
\centering
\subfloat[]{\includegraphics[clip=true, trim = 10 10 10 10, width=0.3\linewidth]{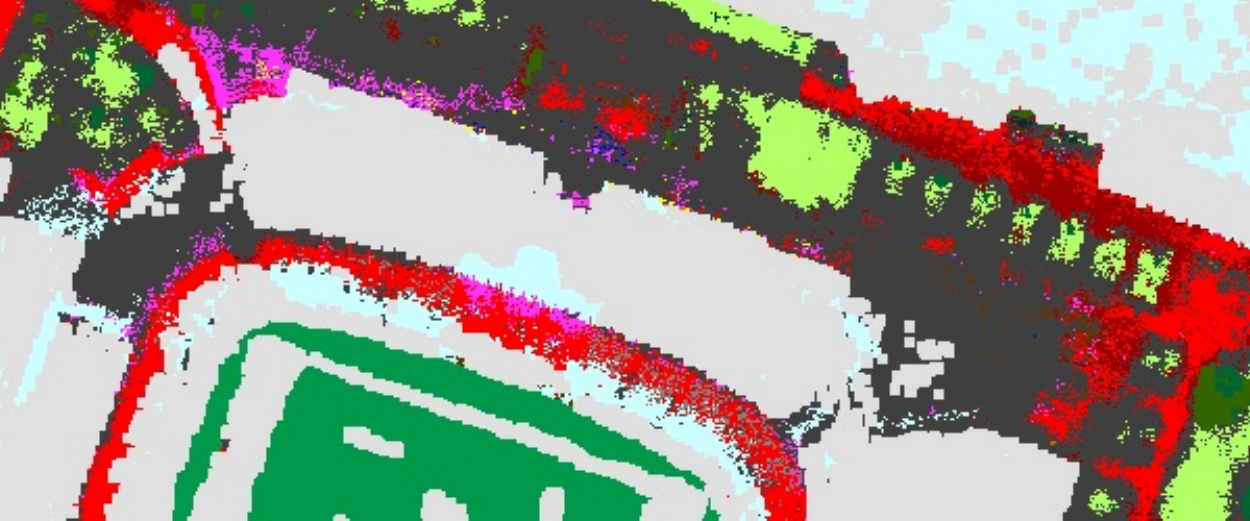}%trim - left, bottom, right, top
\label{fig:pred2D_DFC2018_a}}
\hfil
\subfloat[]{\includegraphics[clip=true, trim = 10 10 10 10, width=0.3\linewidth]{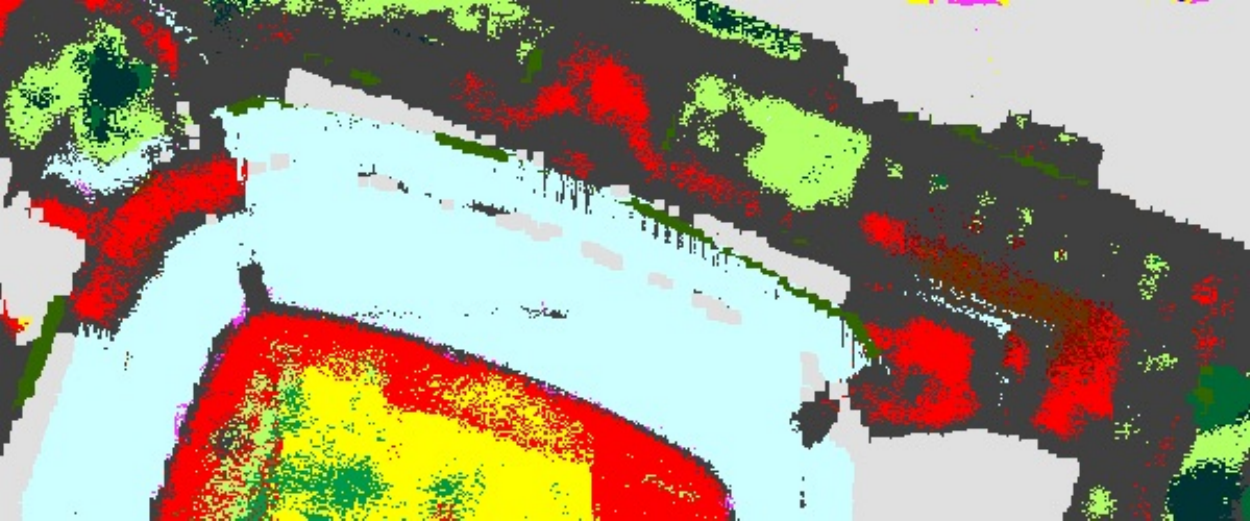}%trim - left, bottom, right, top
\label{fig:pred2D_DFC2018_b}}
\hfil
\subfloat[]{\includegraphics[clip=true, trim = 10 10 10 10, width=0.3\linewidth]{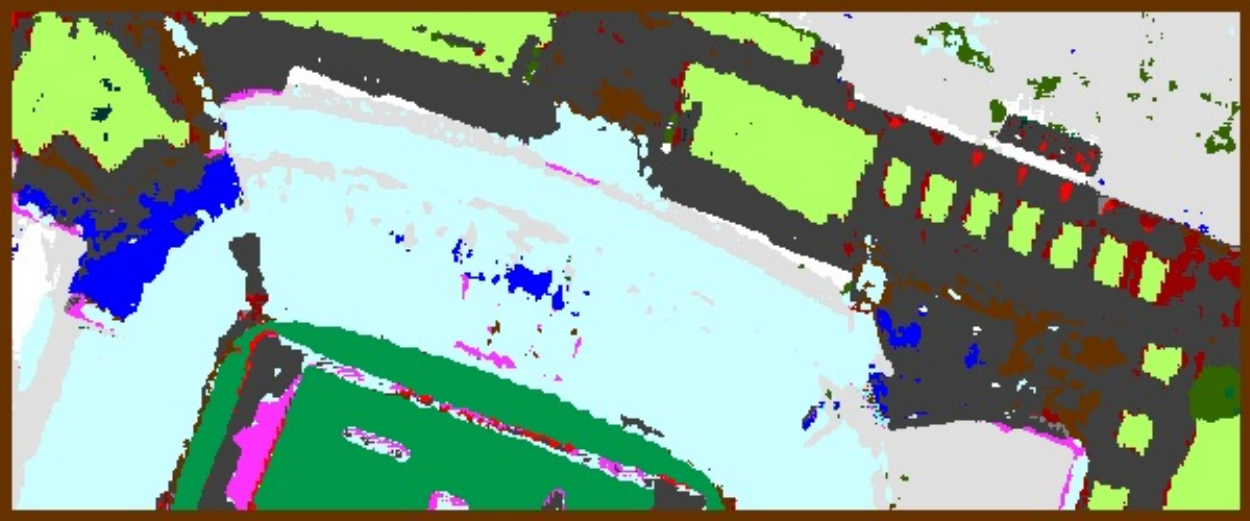}%trim - left, bottom, right, top
\label{fig:pred2D_DFC2018_c}}
\hfil
\subfloat[]{\includegraphics[clip=true, trim = 10 10 10 10, width=0.3\linewidth]{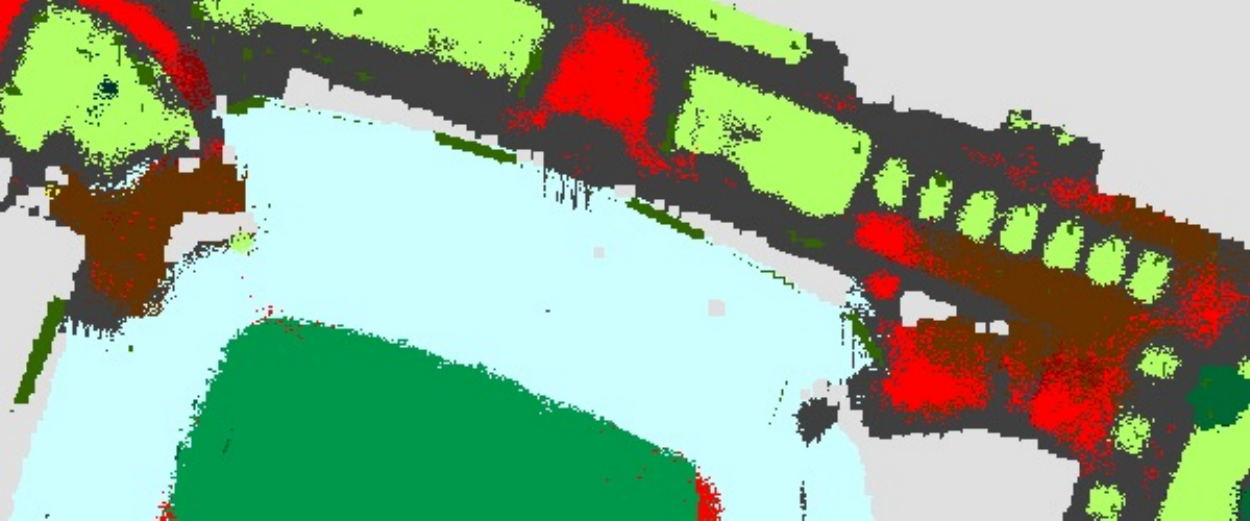}%trim - left, bottom, right, top
\label{fig:pred2D_DFC2018_d}}
\hfil
\subfloat[]{\includegraphics[clip=true, trim = 10 10 10 10, width=0.3\linewidth]{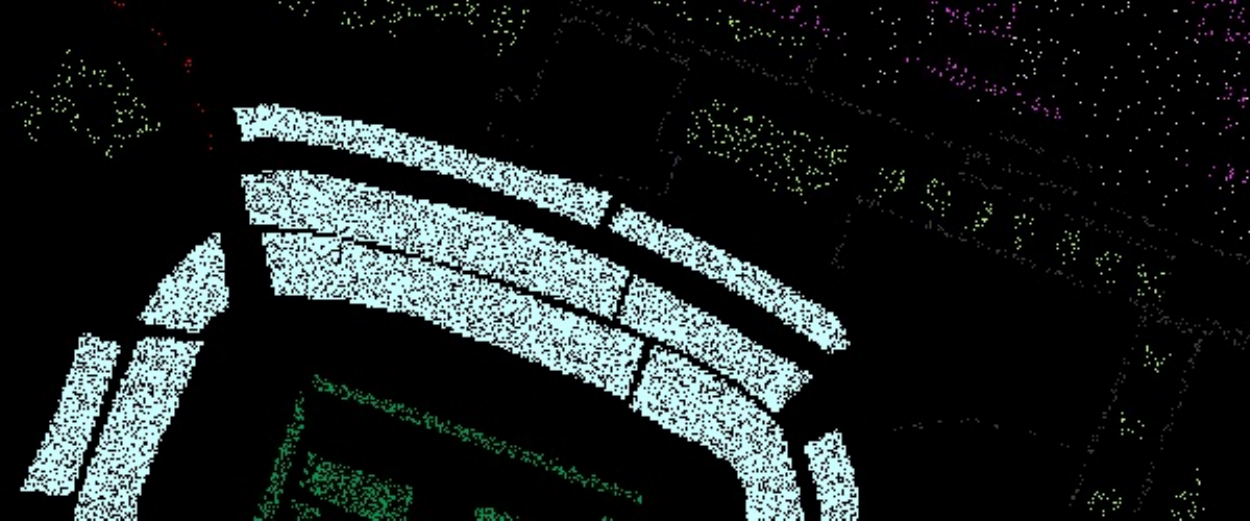}%trim - left, bottom, right, top
\label{fig:pred2D_DFC2018_e}}
\hfil
\subfloat[]{\includegraphics[clip=true, trim = 10 10 10 10, width=0.3\linewidth]{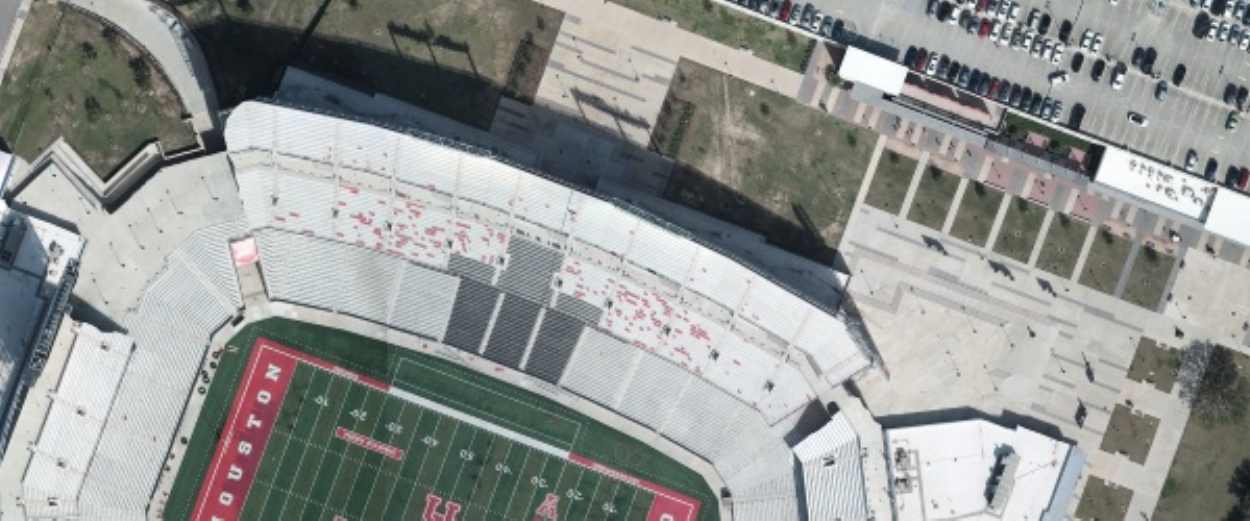}%trim - left, bottom, right, top
\label{fig:pred2D_DFC2018_f}}
\caption{Visual comparison of 2D predictions from different models on the actual Test split of the DFC2018 dataset. (a) DGCNN (3D-based method). (b) Point Transformer (3D-based method). (c) MFT (2D-based method. (d) HyperPointFormer (3D-based method). (e) Ground truth. (f) RGB image.}
\label{fig:pred2D_DFC2018}
\end{figure*}

Fig. \ref{fig:pred2D_DFC2018} shows 2D prediction maps from 3D and 2D approaches in a particularly challenging area. The problematic class, stadium seat, is not detected by DGCNN, but it is successfully identified by the other models. However, Point Transformer fails to classify the grass inside the stadium. Conversely, MFT incorrectly labels some pixels of the building as grass in the top right area. While our 3D approach correctly detects the building, it fails to identify the cars in the parking lot on top of the building.

%At this stage, we postulate that our approach offers greater flexibility compared to dedicated 2D methods, potentially enabling the acquisition of precise semantic labels in both 3D and 2D formats. To illustrate the advantage of the 3D approach, we present a case where predictions are only achievable with the 3D method. Fig. \ref{fig:2D_3D_pred} displays examples where the 2D prediction maps fail to accurately represent real-world conditions. Specifically, the 2D approach cannot depict scenarios where trees, cars, and ground surfaces occupy the same location but at different elevations. The nadir view of the 2D approach can only represent the highest objects at each location in the image.

\begin{table*}[]
    \centering
    \caption{Quantitative Results of 2D Evaluation on the Actual Test Split of The DFC2018 Dataset. The best results are highlighted in bold and the second-best results are underlined.}
    \label{tab:3D_2D_accuracy}
    %\begin{tabular}{p{0.3cm}p{2.2cm}p{1cm}p{1.15cm}p{0.7cm}p{1cm}p{0.6cm}p{1cm}p{0.85cm}p{1.3cm}|p{0.8cm}p{0.6cm}p{0.8cm}}
    \begin{tabular}{p{0.3cm}p{2.2cm}>{\raggedleft\arraybackslash}p{1cm}>{\raggedleft\arraybackslash}p{1.15cm}>{\raggedleft\arraybackslash}p{0.7cm}>{\raggedleft\arraybackslash}p{1cm}>{\raggedleft\arraybackslash}p{0.6cm}>{\raggedleft\arraybackslash}p{1cm}>{\raggedleft\arraybackslash}p{0.85cm}>{\raggedleft\arraybackslash}p{1.3cm}|>{\raggedleft\arraybackslash}p{0.8cm}>{\raggedleft\arraybackslash}p{0.6cm}>{\raggedleft\arraybackslash}p{0.8cm}}
        \hline
        \textbf{No} & \textbf{Class} & \textbf{CNN-NN} \cite{8517699} & \textbf{Fusion-FCN} \cite{8518295} & \textbf{Unet} \cite{9761218} & \textbf{DGConv} \cite{9761218} &  \textbf{MFT} \cite{10153685} & \textbf{NCGLF$^2$} \cite{TU2024102192} & \textbf{Cross-HL} \cite{10462184} & \textbf{SoftFormer} \cite{LIU2024277} & \textbf{DGCNN} \cite{dgcnn} & \textbf{PT} \cite{point_transformer} & \textbf{Ours} \\
        % & & 2D Input & 2D Input & 2D Input & 2D Input & 2D Input & 2D Input & 2D Input & 3D Input & 3D Input &3D Input \\
        \hline
        \textbf{1} & Healthy grass & 99.70 & 88.70 & 83.67 & 82.68 & \underline{98.09} & 96.81 & \textbf{98.31} & 92.82 & 87.51 & 95.90 & 85.26 \\
        \textbf{2} & Stressed grass & 0.00 & \textbf{87.30} & 67.17 & 69.11 & \underline{83.58} & 81.50 & 85.64 & 76.11 & 76.83 & 50.77 & 57.63 \\
        \textbf{3} & Artificial turf & \textbf{94.60} & 64.14 & 39.85 & 56.46 & 59.62 & 4.77 & 50.19 & \underline{83.45} & 4.77 & 17.45 & 9.69 \\
        \textbf{4} & Evergreen trees & \underline{98.10} & 97.05 & 82.20 & 84.00 & 93.83 & 96.51 & 97.15 & 97.38 & \textbf{98.35} & 98.00 & 78.23 \\
        \textbf{5} & Deciduous trees & \textbf{83.00} & 73.02 & 63.25 & 67.78 & \underline{81.16} & 51.20 & 79.68 & 71.76 & 56.14 & 62.14 & 77.24 \\
        \textbf{6} & Bare earth & 45.40 & 27.64 & 52.05 & 48.05 & 36.98 & \textbf{90.90} & \underline{58.38} & 20.00 & 31.70 & 8.37 & 33.38 \\
        \textbf{7} & Water & \textbf{84.50} & 9.15 & 24.43 & 16.83 & 69.16 & 0.00 & \underline{71.62} & 67.20 & 0.05 & 0.00 & 0.18 \\
        \textbf{8} & Res. buildings & 0.00 & \underline{75.03} & 71.72 & 70.50 & \textbf{77.50} & 56.19 & 64.18 & 59.58 & 35.12 & 69.16 & 64.69 \\
        \textbf{9} & Non-res. buildings & \textbf{93.60} & \underline{93.55} & 64.73 & 61.87 & 72.43 & 75.65 & 79.44 & 77.11 & 85.38 & 84.92 & 87.58 \\
        \textbf{10} & Roads & 59.90 & 62.44 & 52.25 & 52.13 & 48.86 & 46.45 & 58.36 & 66.35 & \underline{70.49} & \textbf{84.67} & 42.64 \\
        \textbf{11} & Sidewalks & \underline{66.00} & 68.52 & 62.68 & 63.56 & 63.14 & 64.16 & 51.34 & 44.98 & \textbf{69.12} & 64.54 & 49.27 \\
        \textbf{12} & Crosswalks & 0.00 & 7.46 & \underline{26.35} & \textbf{38.02} & 2.78 & 6.44 & 14.24 & 5.20 & 0.07 & 0.00 & 7.09 \\
        \textbf{13} & Major thorough. & 30.70 & \underline{59.94} & 35.38 & 31.53 & \textbf{68.14} & 30.34 & 46.25 & 17.11 & 22.87 & 18.29 & 31.34 \\
        \textbf{14} & Highways & 30.30 & 17.95 & 43.94 & 33.16 & 23.36 & 36.92 & 18.64 & \underline{54.77} & 14.65 & 49.96 & \textbf{78.95} \\
        \textbf{15} & Railways & \underline{77.60} & \textbf{80.46} & 36.23 & 48.44 & 7.10 & 1.96 & 7.53 & 9.14 & 1.18 & 11.94 & 53.80 \\
        \textbf{16} & Paved parking lots & 72.30 & 60.80 & 71.44 & 72.55 & 57.34 & 61.61 & 58.40 & 75.29 & 58.38 & \textbf{84.16} & \underline{81.47} \\
        \textbf{17} & Unpaved parking l. & 0.00 & 0.00 & 0.00 & 0.00 & 0.00 & 0.00 & 0.00 & 0.00 & 0.00 & 0.00 & 0.00 \\
        \textbf{18} & Cars & 0.00 & 64.33 & \underline{71.45} & \textbf{78.47} & 43.42 & 50.14 & 45.92 & 74.26 & 41.39 & 62.00 & 69.86 \\
        \textbf{19} & Trains & 89.60 & 50.94 & 92.44 & 88.30 & 83.04 & 91.50 & 82.37 & 90.46 & 69.77 & \underline{94.88} & \textbf{96.02} \\
        \textbf{20} & Stadium seats & 86.90 & 41.97 & 75.46 & 71.06 & 95.76 & 32.66 & 93.84 & 64.49 & 77.56 & \textbf{96.90} & \underline{95.81} \\
        \hline
         & OA (\%) & 58.60 & 63.28 & 63.66 & 63.74 & \underline{64.19} & 56.80 & 63.34 & 63.10 & 52.73 & 61.39 & \textbf{64.41}\\
         & Kappa (\%) & 56.00 & 61.00 & 61.00 & \underline{62.00} & 61.98 & 54.12 & 61.16 & 60.84 & 49.79 & 58.99 & \textbf{62.23}\\
         %& Kappa (\%) & 0.56 & 0.61 & 0.61 & \underline{0.62} & 0.619 & \textcolor{red}{54.12} & \textcolor{red}{61.16} & \textcolor{red}{60.84} & 49.79 & 0.590 & \textbf{0.622}\\
        \hline
    \end{tabular}
\end{table*}

\subsection{ISPRS Vaihingen 3D Semantic Segmentation dataset}
\label{sec:vaihingen3D}

\textbf{Description.} To evaluate the generalization ability of our method, we tested it on different datasets containing various spectral attributes. The ISPRS Vaihingen 3D semantic segmentation (Vaihingen3D) dataset \cite{NIEMEYER2014152} is a classical benchmark that includes 3D lidar point cloud data and 2D images. Unlike typical RGB images, the 2D image data comprises Near Infrared (NIR), Red, and Green channels. The images are provided in true orthoimage format, ensuring spatial consistency when integrated with the lidar data.

\begin{figure}[]
\centering
\subfloat[]{\includegraphics[clip=true, trim = 0 0 0 0, width=0.5\linewidth]{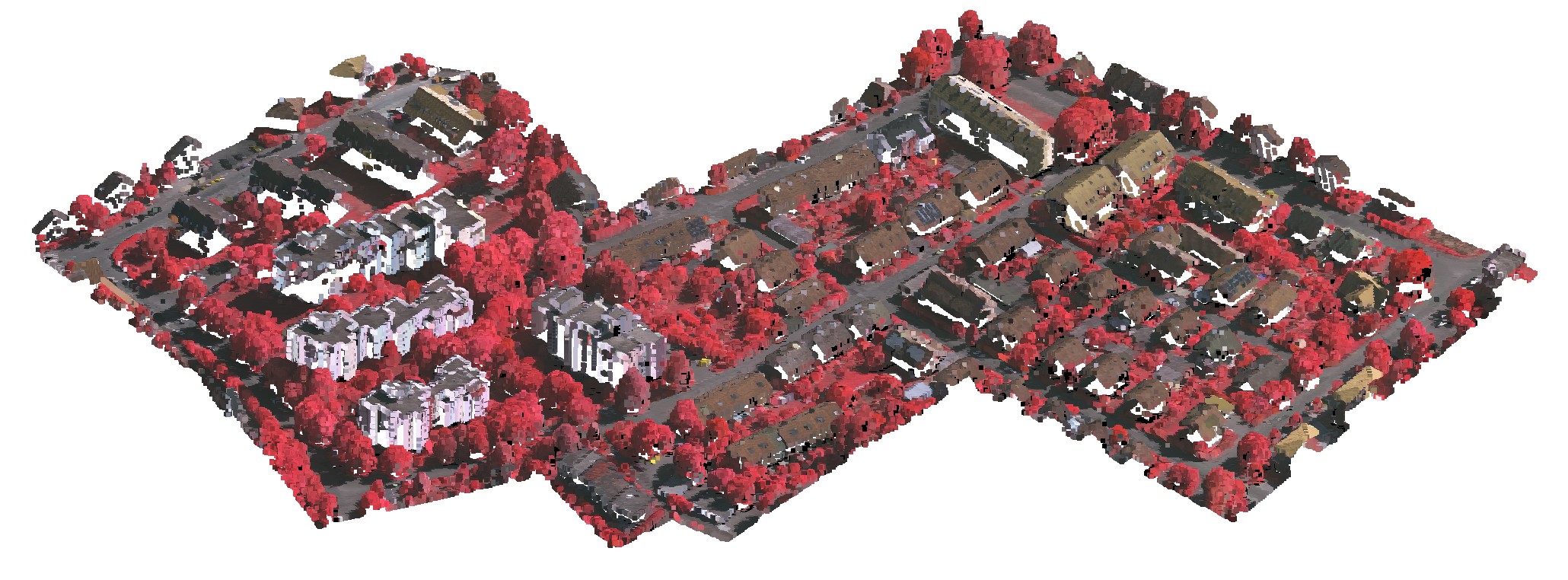}%trim - left, bottom, right, top
\label{fig:v3d_RGBPC_a}}
\hfil
\subfloat[]{\includegraphics[clip=true, trim = 0 0 0 0, width=0.5\linewidth]{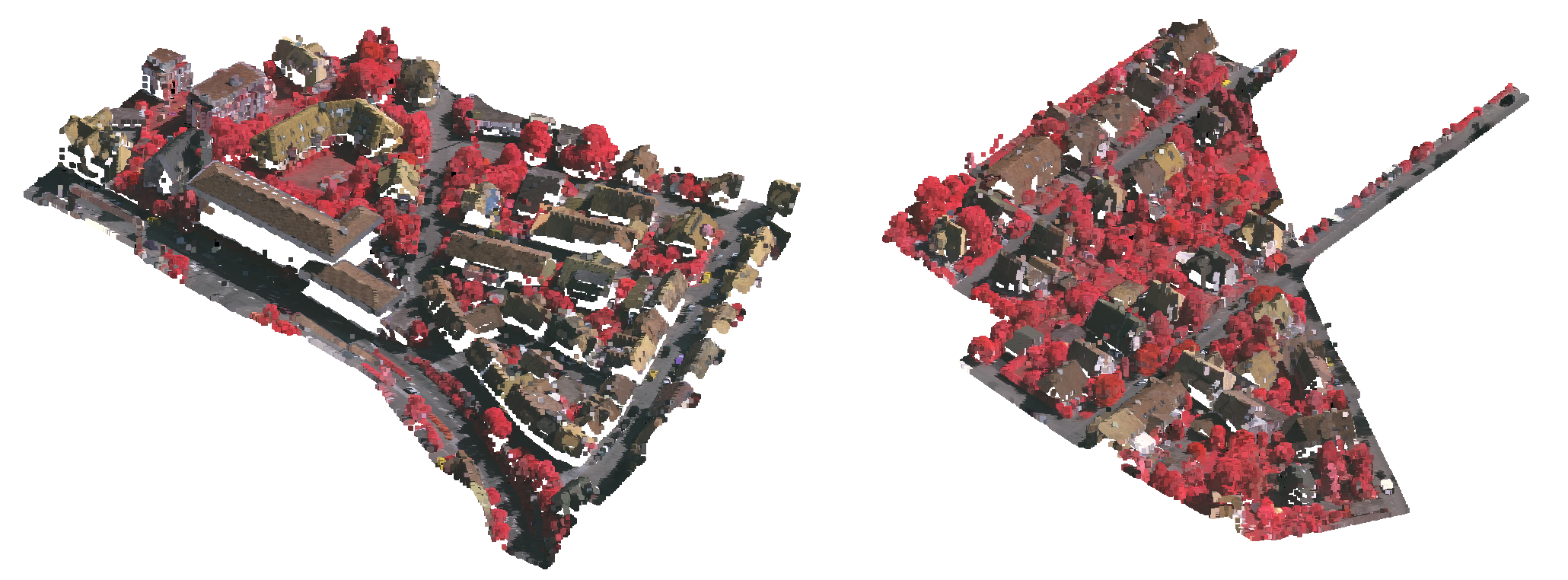}%trim - left, bottom, right, top
\label{fig:v3d_RGBPC_b}}
\caption{The Vaihingen3D dataset rendered with NIR, Red, and Green bands. (a) Train area. (b) Test area.}
\label{fig:v3d_RGBPC}
\end{figure}

The dataset contains 9 semantic classes, which are listed in Table \ref{tab:vaihingen3d}. The dataset is divided into Train and Test areas, as shown in Fig. \ref{fig:v3d_RGBPC}.

Similar to the DFC2018 dataset, we combine lidar and image data by projecting image features (NIR-R-G) in nadir view onto the point cloud using the nearest neighbor algorithm. Fig. \ref{fig:v3d_RGBPC} illustrates the results of our multimodal 3D point clouds.

\begin{table}[]
    \centering
    \caption{Statistics of Vaihingen3D Dataset}
    \label{tab:vaihingen3d}
    \begin{tabular}{llll}
        \hline
        \textbf{No} & \textbf{Class} & \multicolumn{2}{c}{\textbf{Number of samples}} \\
        & & \textbf{Train} & \textbf{Test} \\
        \hline
        \textbf{1} & Powerline & 546 & 600 \\
        \textbf{2} & Low vegetation & 180850 & 98690 \\
        \textbf{3} & Impervious surfaces & 193723 & 101986 \\
        \textbf{4} & Car & 4616 & 3708 \\
        \textbf{5} & Fence/hedge & 12070 & 7422 \\
        \textbf{6} & Roof & 152045 & 109048 \\
        \textbf{7} & Facade & 27250 & 11224 \\
        \textbf{8} & Shrub & 47605 & 24818 \\
        \textbf{9} & Tree & 135173 & 54226 \\
        \hline
    \end{tabular}
\end{table}

\textbf{Experimental setup. } As detailed in \cite{9868041}, we divided the dataset into blocks of 30x30 meters. Points within these blocks, which were assigned to the training split, were randomly sampled to create subsets of 4096 points. The blocks were then split, with 90\% allocated to the training set and 10\% to the validation set. Our models were trained for 200 epochs until convergence, and the model with the highest mean IoU on the validation set was selected for predicting the Test split.

\textbf{Quantitative and qualitative results.} Table \ref{tab:Vaihingen3D_accuracy} presents the accuracy of our method in comparison to general models for point cloud data \cite{dgcnn, pointcnn, point_transformer} and models specifically designed for Airborne Laser Scanning (ALS) data \cite{YOUSEFHUSSIEN2018191, Winiwarter2019, WEN202050}. For evaluation purposes, we re-implemented DGCNN, PointCNN, SpUnet, Point Transformer, and Point Transformer V3, incorporating spectral features (NIR-R-G), which were concatenated with lidar data for data-level fusion. In contrast, we directly provide the accuracy numbers for RIT \cite{YOUSEFHUSSIEN2018191}, alsNet \cite{Winiwarter2019}, and D-FCN \cite{WEN202050} from their respective papers.

Our approach outperforms the baseline model, Point Transformer, with an increase of 1.7 percentage points for OA and 2.9 percentage points for the mean F1 score. Furthermore, our method demonstrates superior performance compared to both general deep learning models for point cloud classification and models specifically designed for ALS data.

\begin{figure*}[]
\centering
\subfloat[]{\includegraphics[clip=true, trim = 300 0 300 0, width=0.24\linewidth]{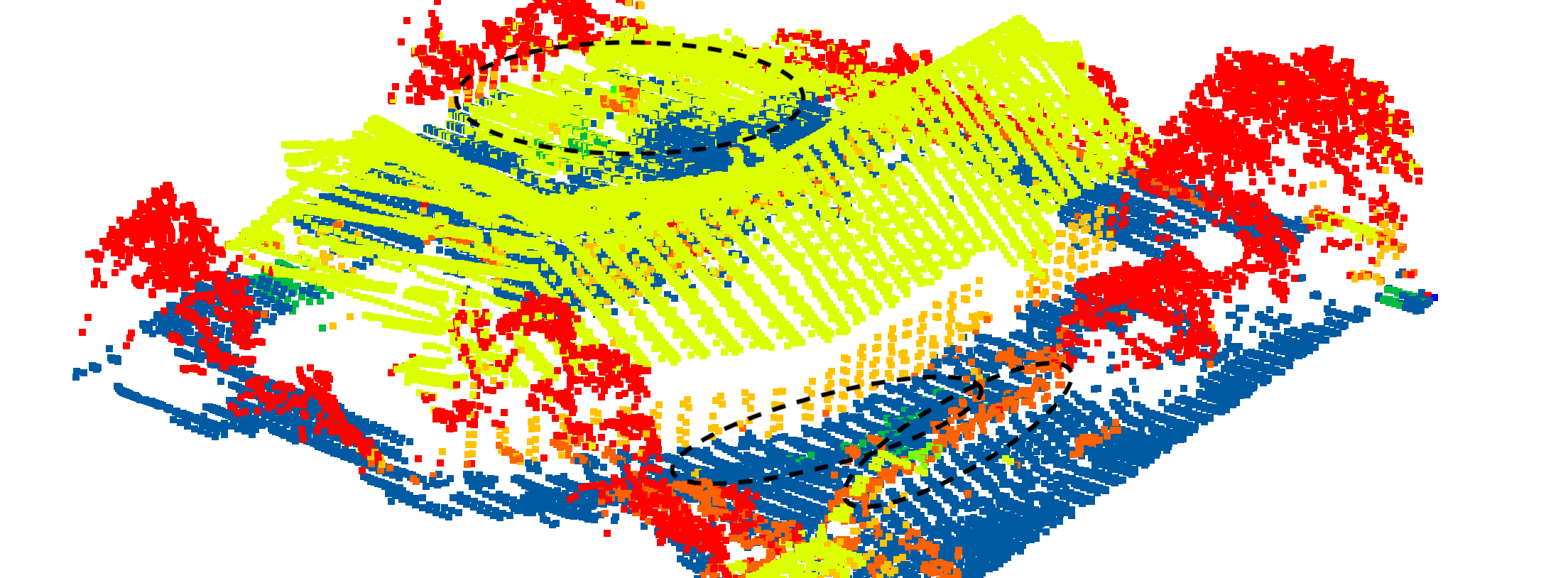}%trim - left, bottom, right, top
\label{fig:predV3D_a}}
\hfil
\subfloat[]{\includegraphics[clip=true, trim = 300 0 300 0, width=0.24\linewidth]{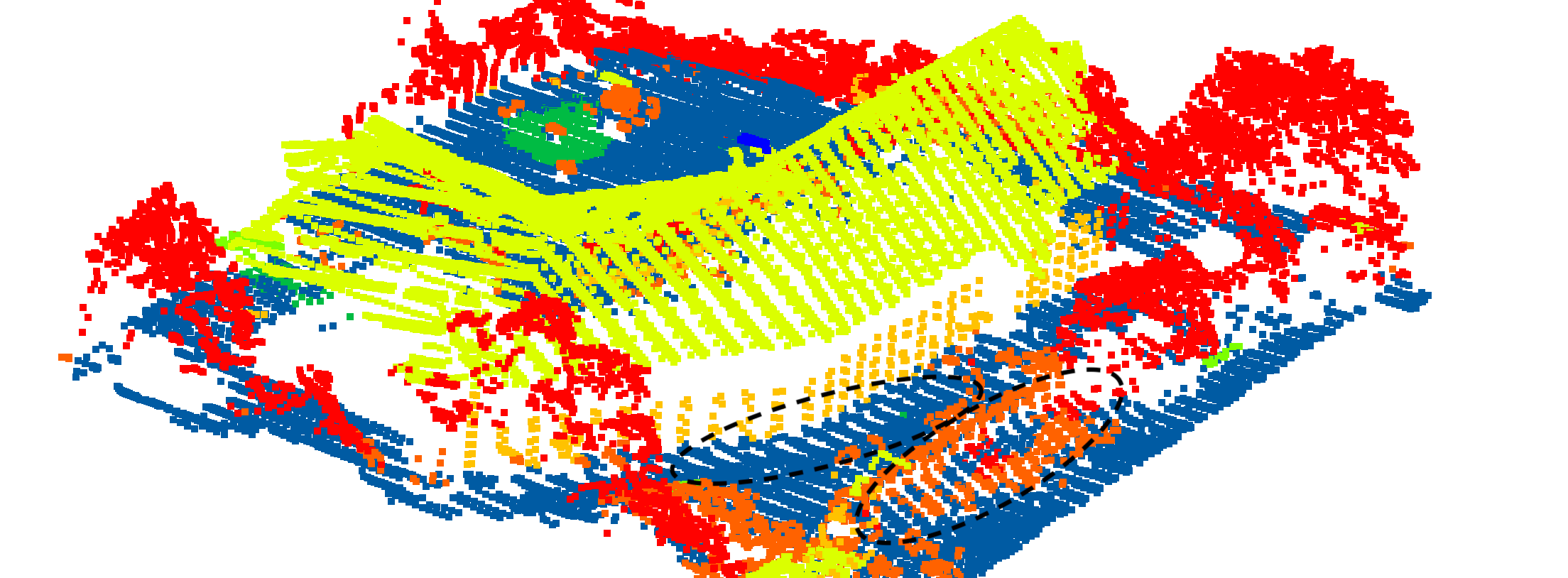}%trim - left, bottom, right, top
\label{fig:predV3D_b}}
\hfil
\subfloat[]{\includegraphics[clip=true, trim = 300 0 300 0, width=0.24\linewidth]{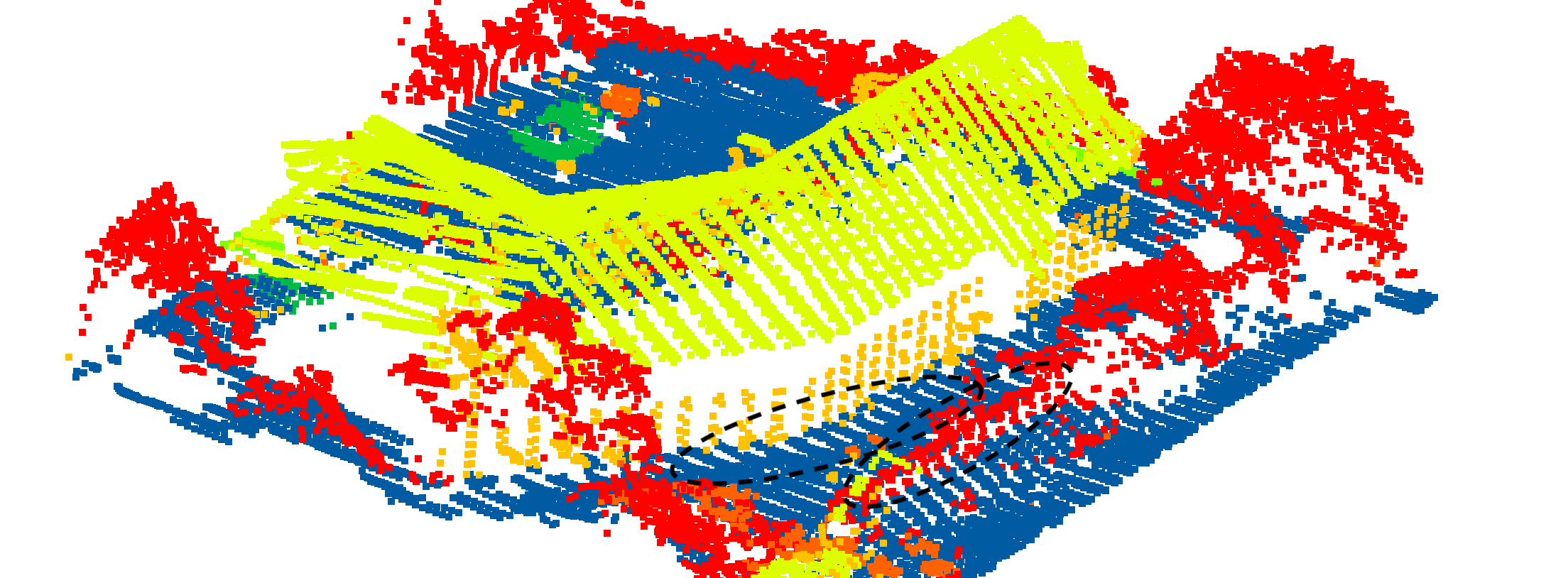}%trim - left, bottom, right, top
\label{fig:predV3D_c}}
\hfil
\subfloat[]{\includegraphics[clip=true, trim = 300 0 300 0, width=0.24\linewidth]{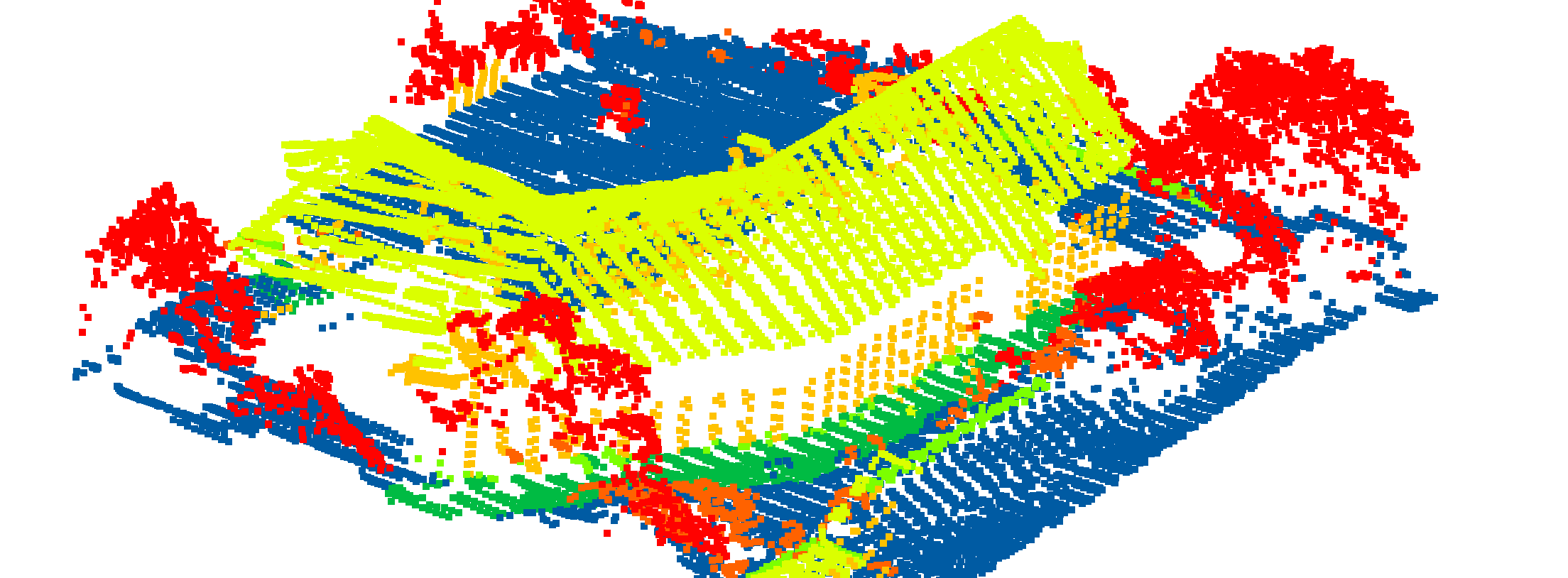}%trim - left, bottom, right, top
\label{fig:predV3D_d}}
\caption{Visual comparison of 3D predictions from different models on the Vaihingen3D dataset. (a) DGCNN. (b) Point Transformer. (c) HyperPointFormer. (d) Ground truth.}
\label{fig:predV3D}
\end{figure*}

Fig. \ref{fig:predV3D} presents a visual comparison of 3D predictions from DGCNN, Point Transformer, and our proposed method. All models accurately predicted points on building roofs and facades. However, DGCNN misclassified large areas at the top of the scene as building roofs. While Point Transformer and our method provided more accurate predictions in this area, none of the models successfully classified points on fences or impervious surfaces adjacent to the buildings. Additionally, Point Transformer made more errors in classifying low vegetation as shrubs.

\begin{table*}[h]
    \centering
    \caption{Quantitative Results of Vaihingen3D Dataset. The best results are highlighted in bold and the second-best results are underlined.}
    \label{tab:Vaihingen3D_accuracy}
    \begin{tabular}{p{0.3cm}p{2.3cm}p{1.1cm}p{1.3cm}p{1.3cm}p{1.3cm}p{1.5cm}p{1cm}p{1cm}p{1cm}p{1cm}}
        \hline
        \textbf{No} & \textbf{Class} & \textbf{DGCNN} \cite{dgcnn}& \textbf{PointCNN} \cite{pointcnn} & \textbf{alsNet} \cite{Winiwarter2019} & \textbf{RIT} \cite{YOUSEFHUSSIEN2018191} & \textbf{D-FCN} \cite{WEN202050} & \textbf{SpUnet} \cite{8579059} & \textbf{PT} \cite{point_transformer} & \textbf{PTv3} \cite{10658198} & \textbf{Ours} \\
        \hline
        \textbf{1} & Powerline & 57.9 & 55.7 & \underline{70.1} & 37.5 & \textbf{70.4} & 20.3 & 65.6 & 63.9 & 65.5 \\
        \textbf{2} & Low vegetation & 79.6 & \underline{80.7} & 80.5 & 77.9 & 80.2 & 80.5 & 78.3 & \textbf{81.9} & \textbf{81.9} \\
        \textbf{3} & Impervious surfaces  & 90.6 & 90.9 & 90.2 & \textbf{91.5} & \underline{91.4} & 89.6 & 88.8 & 90.0 & 90.9 \\
        \textbf{4} & Car & 66.1 & \underline{77.8} & 45.7 & 73.4 & \textbf{78.1} & 75.1 & 72.9 & 73.4 & \underline{77.8} \\
        \textbf{5} & Fence & 31.5 & 30.5 & 0.76 & 18.0 & 37.0 & \underline{38.9} & 34.3 & \textbf{46.7} & 38.0 \\
        \textbf{6} & Roof & 91.6 & 92.5 & 93.1 & \underline{94.0} & 93.0 & \textbf{94.6} & 92.2 & \textbf{94.6} & 93.3 \\
        \textbf{7} & Facade & 54.3 & 56.9 & 47.3 & 49.3 & 60.5 & 60.9 & 62.4 & \underline{63.5} & \textbf{64.0} \\
        \textbf{8} & Shrub & 41.6 & 44.4 & 34.7 & 45.9 & \underline{46.0} & \textbf{47.4} & 43.8 & 28.7 & 50.0 \\
        \textbf{9} & Tree & 77.0 & 79.6 & 74.5 & \textbf{82.5} & 79.4 & \underline{82.3} & 79.7 & 80.3 & 77.4 \\
        \hline
         & OA (\%) & 81.2 & 82.2 & 80.6 & 81.6 & 82.2 & \underline{83.4} & 81.3 & \textbf{83.5} & 83.0 \\
         & mean F1 (\%) & 65.0 & 67.7 & 60.4 & 63.3 & \underline{70.7} & 65.5 & 68.7 & 69.2 & \textbf{71.0} \\
        \hline
    \end{tabular}
\end{table*}

\subsection{2019 IEEE Data Fusion Contest dataset}
\label{sec:dfc2019}

\textbf{Description.} The 2019 IEEE GRSS Data Fusion Contest (DFC2019) dataset \cite{9246669} was designed for four different tasks. The first task involves point cloud classification, while the others focus on 3D reconstruction from images. For our study, we used the point cloud data from the classification task to evaluate the generalization ability of our approach. The dataset consists of 500 x 500 meter segments with four classes: ground, trees (or high vegetation), buildings, water, and elevated roads or bridges.

\textbf{Experimental setup. } We utilized 110 point clouds for training and 10 point clouds for testing. Following the same approach as in the Vaihingen3D experiment, we divided the point clouds into 30 x 30 meter blocks. Models were trained using 90\% of the blocks from the training set, with the remaining 10\% reserved for validation. The model that achieved the highest accuracy on the validation set was then used for inference on the testing set to compute the final accuracy. Since the dataset does not include corresponding hyperspectral or RGB image data, our implementation for this experiment relied solely on the echo information from the lidar data as the spectral information for data fusion.

\textbf{Quantitative and qualitative results.} Table \ref{tab:DFC2019_accuracy} compares the accuracy of our proposed method with existing studies as reported in \cite{9246669}. DPNet, a variant of PointNet++, incorporates additional dense connections. We have cited the accuracy results for DPNet, PointNet++, and PointSIFT from \cite{9246669}, while we re-implemented PointNet, DGCNN, and Point Transformer for a more comprehensive comparison.

Our approach provides a modest improvement over the baseline model, Point Transformer, with an increase of 0.1 percentage points in overall accuracy (OA) and 0.4 percentage points in mean IoU. Although the improvement is relatively small, this is expected given that the dataset only contains lidar point cloud data with no additional spectral information and includes a limited number of classes. In contrast, our previous experiments with the DFC2018 dataset showed more significant improvements, as that dataset combined lidar point clouds with hyperspectral and RGB images, which provide richer semantic information for distinguishing challenging classes.

\begin{figure*}[]
\centering
\subfloat[]{\includegraphics[clip=true, trim = 300 0 300 0, width=0.24\linewidth]{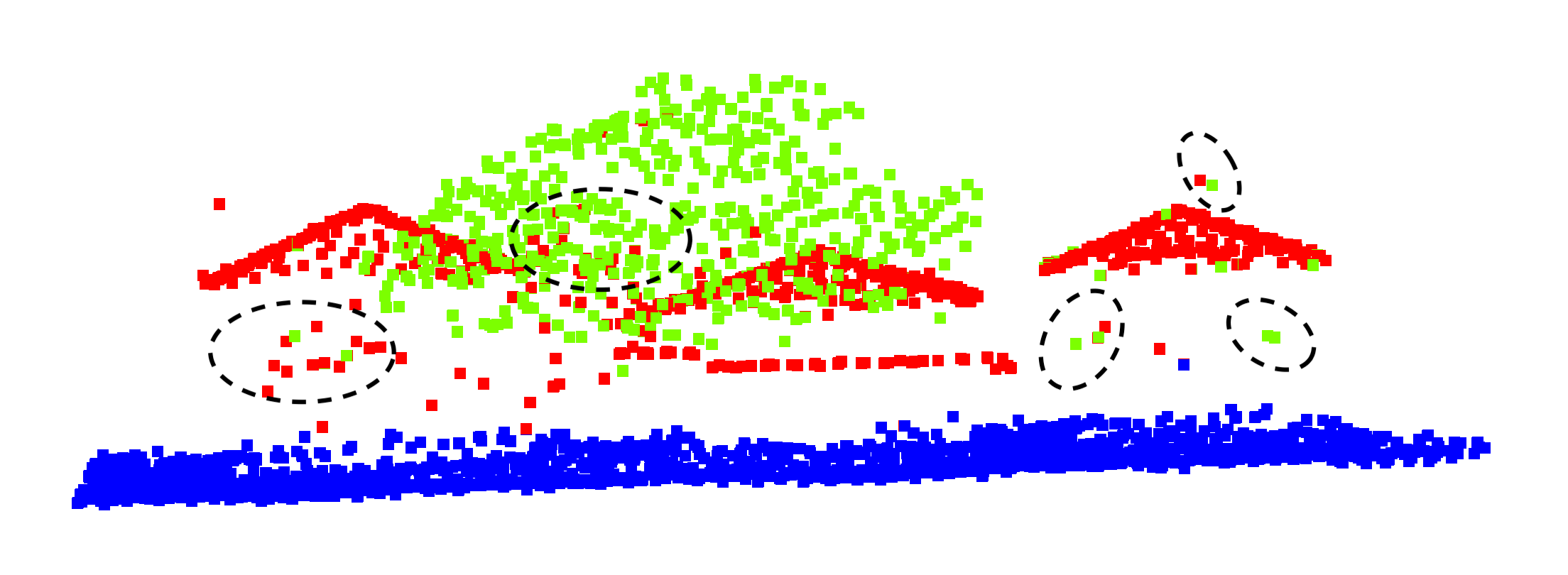}%trim - left, bottom, right, top
\label{fig:predDFC2019_a}}
\hfil
\subfloat[]{\includegraphics[clip=true, trim = 300 0 300 0, width=0.24\linewidth]{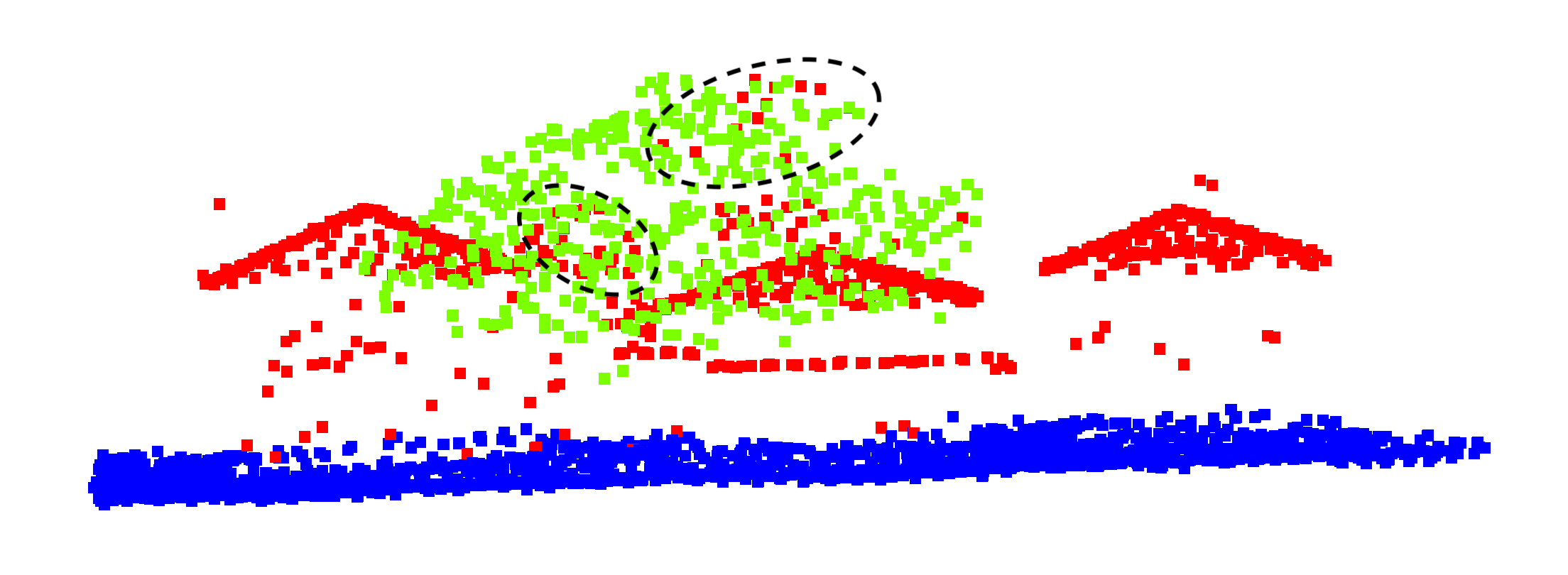}%trim - left, bottom, right, top
\label{fig:predDFC2019_b}}
\hfil
\subfloat[]{\includegraphics[clip=true, trim = 300 0 300 0, width=0.24\linewidth]{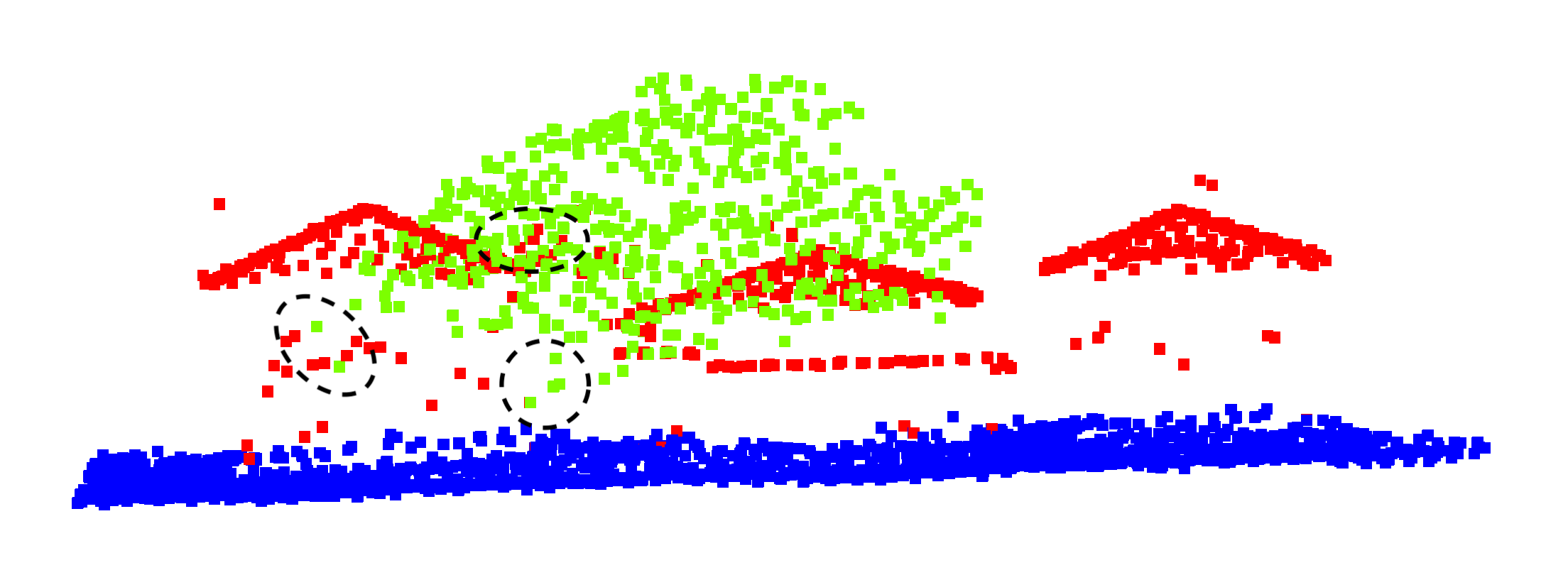}%trim - left, bottom, right, top
\label{fig:predDFC2019_c}}
\hfil
\subfloat[]{\includegraphics[clip=true, trim = 300 0 300 0, width=0.24\linewidth]{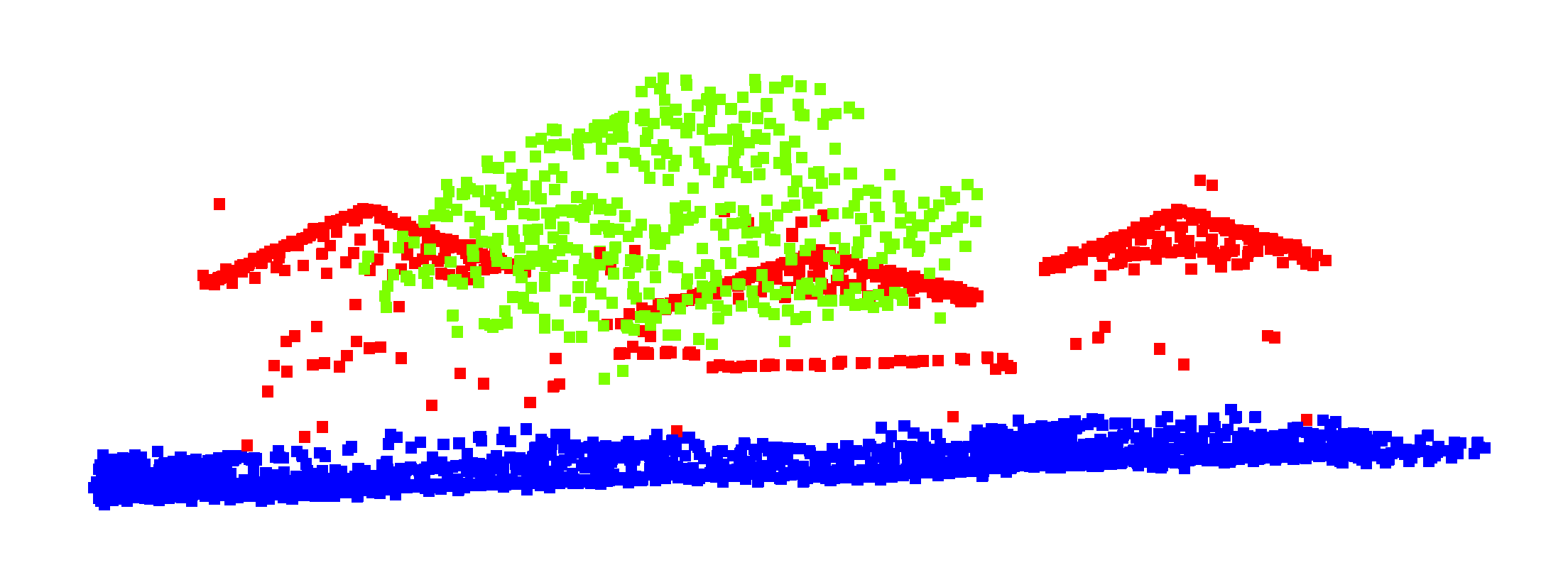}%trim - left, bottom, right, top
\label{fig:predDFC2019_d}}
\caption{Visual comparison of 3D predictions from various models on the DFC2019 dataset. (a) DGCNN. (b) Point Transformer. (c) HyperPointFormer. (d) Ground truth.}
\label{fig:predDFC2019}
\end{figure*}

Fig. \ref{fig:predDFC2019} compares the predictions made by DGCNN, Point Transformer, and our proposed method. DGCNN incorrectly classified some building points as trees, while Point Transformer misclassified tree points as buildings. Our method produced more accurate predictions for this scene, although some points on buildings were still misclassified as trees due to their lower density compared to roof points.

\begin{table*}[]
    \centering
    \caption{Quantitative Results of DFC2019 Dataset. The best results are highlighted in bold and the second-best results are underlined.}
    \label{tab:DFC2019_accuracy}
    \begin{tabular}{p{0.3cm}p{2.3cm}p{1.3cm}p{1.3cm}p{1.3cm}p{1.3cm}p{1.3cm}p{1.3cm}p{1.3cm}}
        \hline
        \textbf{No} & \textbf{Class} & \textbf{PointNet} \cite{pointnet} & \textbf{PointNet++} \cite{pointnet2} & \textbf{DGCNN} \cite{dgcnn} & \textbf{PointSIFT} \cite{9246669} & \textbf{DPNet} \cite{9246669} & \textbf{PT} \cite{point_transformer} & \textbf{Ours} \\
        \hline
        \textbf{1} & Ground & 96.0 & 96.5 & 97.0 & 97.5 & \underline{98.0} & 98.3 & \textbf{98.4} \\
        \textbf{2} & Trees & 74.4 & 93.8 & 89.5 & 94.4 & 94.2 & \textbf{95.7} & \underline{95.2} \\
        \textbf{3} & Buildings & 73.5 & 88.5 & 86.7 & 90.8 & 91.4 & \underline{92.5} & \textbf{93.4} \\
        \textbf{4} & Water & 85.4 & 72.0 & 86.3 & 86.6 & \underline{88.6} & \textbf{92.3} & 85.9 \\
        \textbf{5} & Bridges & 54.8 & 69.7 & 68.1 & 78.5 & \underline{83.5} & 77.1 & \textbf{84.7} \\
        \hline
         & OA (\%) & 93.1 & 96.7 & 96.4 & 97.6 & 97.8 & \underline{98.1} & \textbf{98.2} \\
         & mIoU (\%) & 76.9 & 84.1 & 85.5 & 89.6 & 91.2 & \underline{91.2} & \textbf{91.5} \\
         \hline
    \end{tabular}
\end{table*}

\section{Discussion}
\label{sec:discussion}

\subsection{Ablation study}
%\subsection{Ablation study and discussion}

To demonstrate the capability of our method, we conducted ablation studies by replacing the architectures and the fusion modules with typical fusion methods. 

\textbf{Comparison with early and late fusion architectures.} We compare our method with early and late fusion strategies using Point Transformer as the backbone model. In the case of early fusion, we combine all modalities by stacking them as input features before feeding them into the network. For late fusion, we divide the network into two separate sub-networks, each with its own encoder and decoder sections. After encoding the respective modalities, we stack the features from the last decoder layers and pass them through the final fully-connected layers for classification. 

Table \ref{tab:fusion_func} shows that mid-level fusion, incorporating multi-scale feature fusion, outperforms both early and late fusion approaches. This demonstrates the advantage of integrating features at intermediate stages of the network, allowing for better exploitation of the complementary information provided by different modalities.

\begin{table}[h]
    \centering
    \caption{Comparison of Early, Late, and Mid Fusion}
    \label{tab:fusion_arch}
    \begin{tabular}{llll}
        \hline
        \textbf{Architecture} & \textbf{Precision (\%)} & \textbf{Recall (\%)} & \textbf{F1 (\%)} \\ 
        \hline
        Early (base) & 64.18 & 50.64 & 52.31 \\
        Late & 60.34 & 48.18 & 49.57 \\
        Mid & 66.16 & 52.89 & 54.54 \\
        \hline
    \end{tabular}
\end{table}

%\textbf{Comparison with different fusion functions.} Continuing with mid-fusion as the selected architecture, we explore different fusion functions. Four fusion functions were considered: concatenation, average-pooling, max-pooling, and summation. As shown in Table \ref{tab:fusion_func}, our proposed fusion module, \texttt{CPA}, outperforms these fusion strategies. We contend that the interaction of features at different scales aids the module in learning the relevance of features from each modality.

\textbf{The effectiveness of \texttt{CPA} module.} To evaluate the effectiveness of cross-attention, we conduct an experiment by removing the cross-attention mechanism from the dual-branch architecture. Table \ref{tab:fusion_func} compares the model utilizing the \texttt{CPA} module during the fusion and the model without \texttt{CPA}, while keeping the rest of the architecture unchanged. The results show a significant performance drop when \texttt{CPA} is removed, highlighting the importance of enabling one branch to assess relevance of features from another during fusion.

\begin{table}[h]
    \centering
    \caption{The effectiveness of \texttt{CPA} module}
    \label{tab:fusion_func}
    \begin{tabular}{llll}
        \hline
        \textbf{Fusion} & \textbf{Precision (\%)} & \textbf{Recall (\%)} & \textbf{F1 (\%)} \\
        \hline
        With \texttt{CPA} & 69.05 & 53.40 & 55.54 \\
        Without \texttt{CPA} & 63.96 & 49.93 & 51.26 \\
        \hline
    \end{tabular}
\end{table}

\textbf{Different backbone models.} We further investigated the importance of the dual-branch network with the \texttt{CPA} module by substituting the backbone model with classical models for point cloud classification, namely PointNet and DGCNN. PointNet is a pioneer in the field, renowned for its capacity to learn directly from raw 3D point clouds. On the other hand, DGCNN excels in learning local 3D features by leveraging k-nearest neighbor graphs.

Table \ref{tab:backbone} demonstrates a similar trend, wherein the models exhibit improved performance when individual modalities are processed separately and subsequently combined using the \texttt{CPA} module. This underscores the adaptability of our approach, paving the way for future research avenues.

\begin{table}[h]
    \centering
    \caption{Comparison of Different Backbone Models}
    \label{tab:backbone}
    \begin{tabular}{llll}
        \hline
        \textbf{Backbone} & \textbf{Precision (\%)} & \textbf{Recall (\%)} & \textbf{F1 (\%)} \\
        \hline
        PointNet & 50.95 & 33.51 & 35.75 \\
        PointNet with \texttt{CPA} & 53.70 & 35.89 & 38.66 \\
        DGCNN & 61.23 & 46.68 & 48.08 \\
        DGCNN with \texttt{CPA} & 63.35 & 49.35 & 50.53 \\
        \hline
    \end{tabular}
\end{table}

\textbf{Significance of input features.} We investigated the importance of multi-source data by training models with different input features using the DFC2018 dataset, which provides lidar, RGB images, and hyperspectral images. The labels in the DFC2018 dataset are unique due to their diverse shapes and materials, making it essential to combine geometrical and spectral features for accurate detection across various classes.

We trained the Point Transformer model with three different input configurations. In the first experiment, we used only the XYZ coordinates from the lidar data. In the second experiment, we used only the spectral features from RGB and hyperspectral images. While we used image features, the model was trained in a point-based manner, meaning that only the point features were used, excluding the XYZ coordinates as input. The XYZ coordinates were still employed for the k-NN points computation and for performing the FPS operation during the downsampling phase.

In the final experiment, we concatenated all the data (lidar and image-based features) as the input features. Table \ref{tab:input_feat} summarizes the accuracy results for the different input configurations on the DFC2018 dataset, where we trained on Area 1 and evaluated on Area 2. From these results, it is evident that combining lidar and image-based data (such as RGB and hyperspectral images) significantly improved the accuracy. A similar result can be seen in the Vaihingen3D dataset from Table \ref{tab:input_feat_V3D}.

\begin{table}[h]
    \centering
    \caption{Comparison of Different Input Modalities on DFC2018 Dataset}
    \label{tab:input_feat}
    \begin{tabular}{llll}
        \hline
        \textbf{Features} & \textbf{Precision (\%)} & \textbf{Recall (\%)} & \textbf{F1 (\%)} \\
        \hline
        Lidar & 43.12 & 37.46 & 36.83 \\
        RGB + HSI & 45.54 & 42.54 & 37.11 \\
        Lidar + RGB + HSI & 64.18 & 50.64 & 52.31 \\
        \hline
    \end{tabular}
\end{table}

\begin{table}[h]
    \centering
    \caption{Comparison of Different Input Modalities on Vaihingen3D Dataset}
    \label{tab:input_feat_V3D}
    \begin{tabular}{llll}
        \hline
        \textbf{Features} & \textbf{Precision (\%)} & \textbf{Recall (\%)} & \textbf{F1 (\%)} \\
        \hline
        Lidar & 68.47 & 66.20 & 66.70 \\
        Lidar + NIR-R-G & 70.45 & 68.07 & 68.67 \\
        \hline
    \end{tabular}
\end{table}

\subsection{Results interpretation}

\textbf{Importance of modality-specific processing.} We found that splitting the network into two dedicated branches, i.e., one for processing lidar and the other processing hyperspectral data, is crucial. Each modality provides fundamentally different information, and separate processing allows each branch to specialize in extracting complementary features.

\textbf{Benefits of cross-attention over hard fusion. } Unlike hard fusion methods such as concatenation or summation, our \texttt{CPA} mechanism learns dynamic fusion weights between modalities. This allows the network to adaptively emphasize either geometric or spectral features depending on the context.

\begin{figure}[h!]
\centering
\subfloat[]{\includegraphics[clip=true, trim = 0 25 25 60, width=0.5\linewidth]{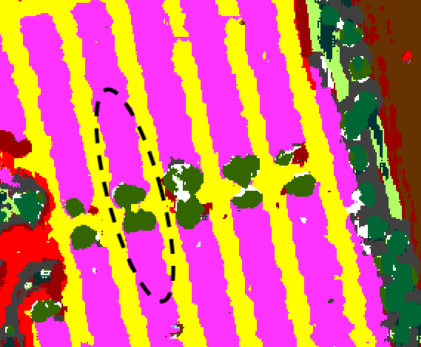}%trim - left, bottom, right, top
\label{fig:2D_3D_pred_a}}
\hfil
\subfloat[]{\includegraphics[clip=true, trim = 0 0 0 0, width=0.98\linewidth]{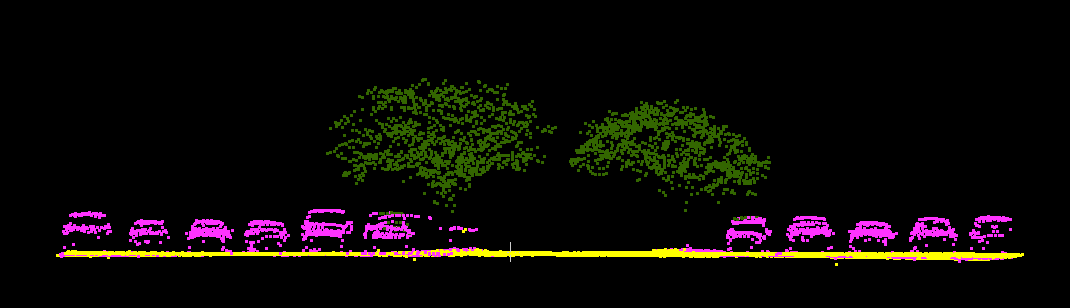}%trim - left, bottom, right, top
\label{fig:2D_3D_pred_b}}
\caption{Comparison of 2D and 3D predictions on the same area. (a) 2D prediction map using MFT \cite{10153685}. (b) Cross section from 3D prediction using HyperPointFormer.}
\label{fig:2D_3D_pred}
\end{figure}

\textbf{Advantages of 3D prediction for spatial analysis. } At this stage, we postulate that our approach offers greater flexibility compared to dedicated 2D methods, potentially enabling the acquisition of precise semantic labels in both 3D and 2D formats. To illustrate the advantage of the 3D approach, we present a case where predictions are only achievable with the 3D method. Fig. \ref{fig:2D_3D_pred} displays examples where the 2D prediction maps fail to accurately represent real-world conditions. Specifically, the 2D approach cannot depict scenarios where trees, cars, and ground surfaces occupy the same location but at different elevations. The nadir view of the 2D approach can only represent the highest objects at each location in the image.

\subsection{Features visualization}

We utilized t-SNE (t-Distributed Stochastic Neighbor Embedding) \cite{JMLR:v9:vandermaaten08a} to visualize the high-dimensional features learned by our neural network, specifically to assess the effectiveness of our classification model. By applying t-SNE to the output features of the last hidden layer before the final classification layer, we aimed to gain insight into the feature space and the separability of different classes. The resulting t-SNE plots allowed us to observe the distinct clusters formed by different classes, demonstrating the network's ability to learn meaningful and discriminative representations.

\begin{figure}[!t]
\centering
\subfloat[]{\includegraphics[clip=true, trim = 50 120 50 50, width=0.47\linewidth]{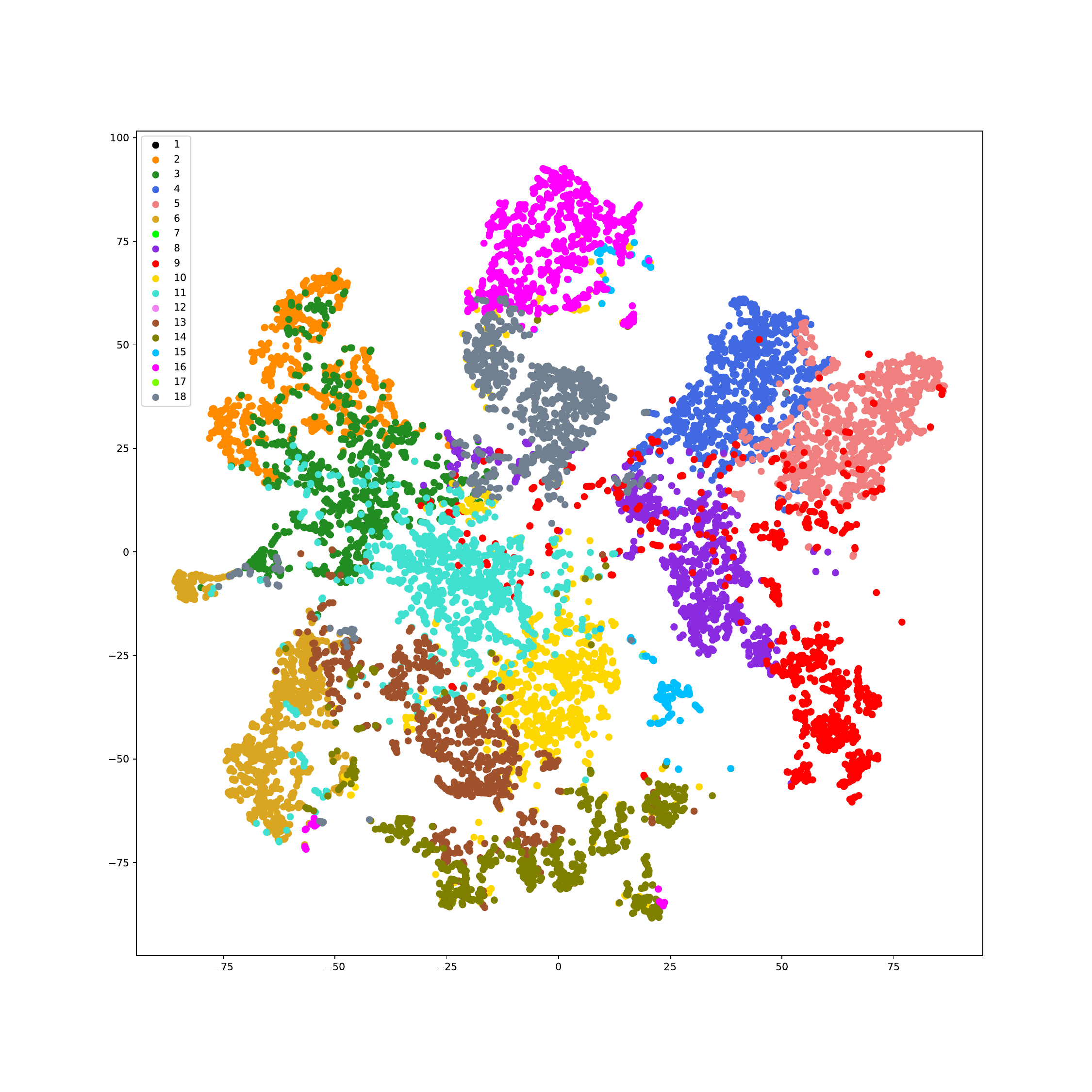}%trim - left, bottom, right, top
\label{fig:tsne_a}}
\hfil
\subfloat[]{\includegraphics[clip=true, trim = 50 120 50 50, width=0.47\linewidth]{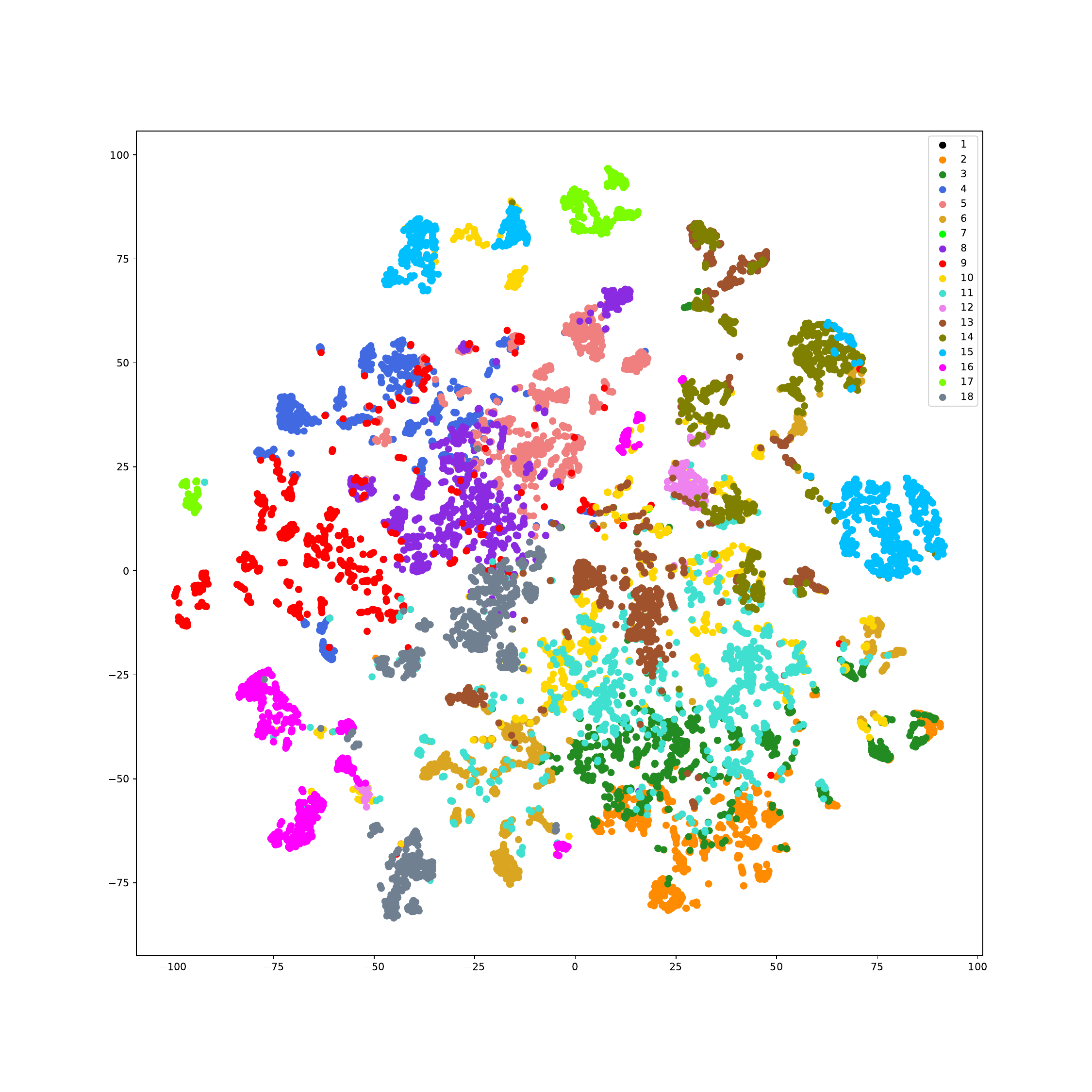}%trim - left, bottom, right, top
\label{fig:tsne_b}}
\hfil
\subfloat[]{\includegraphics[clip=true, trim = 50 120 50 50, width=0.47\linewidth]{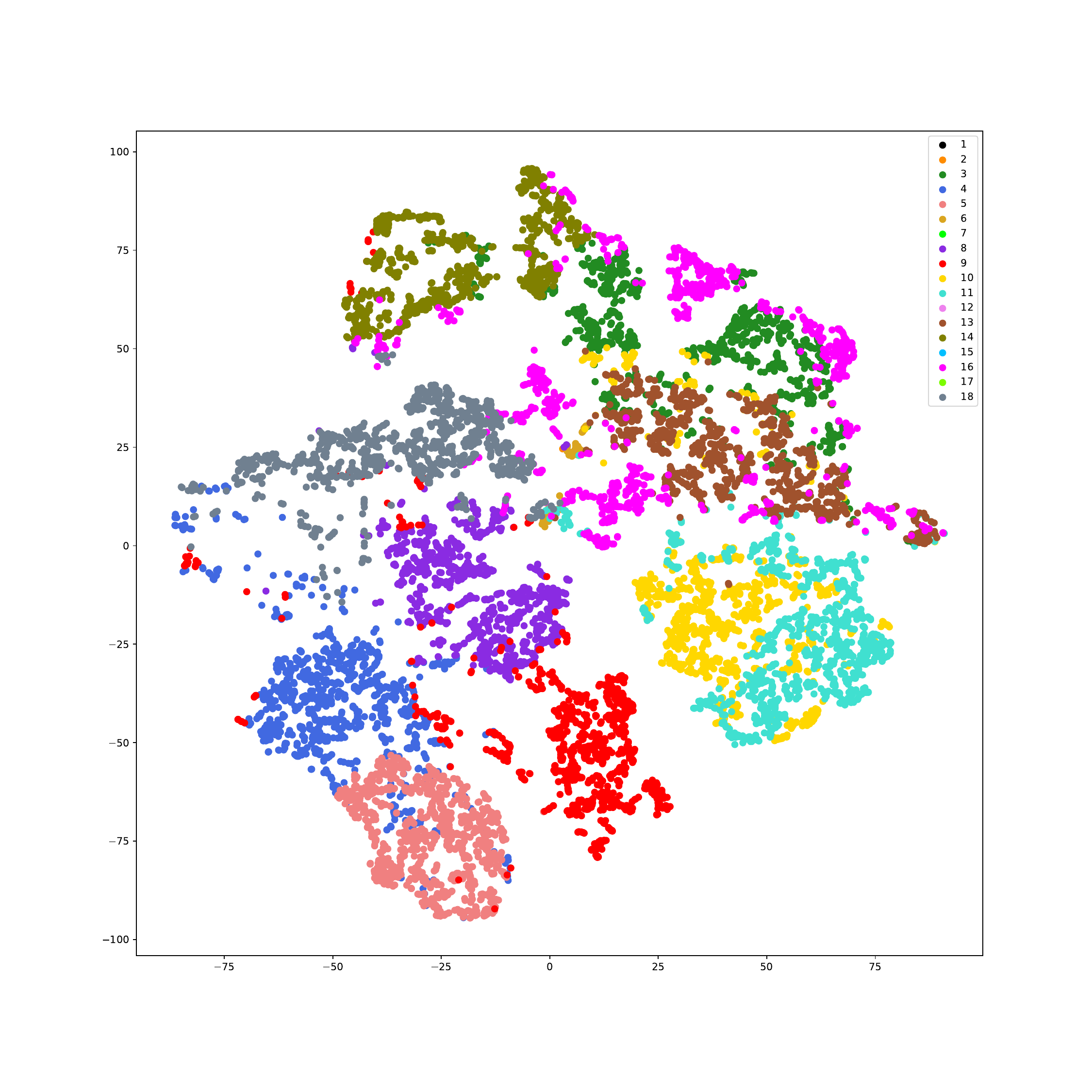}%trim - left, bottom, right, top
\label{fig:tsne_c}}
\hfil
\subfloat[]{\includegraphics[clip=true, trim = 50 120 50 50, width=0.47\linewidth]{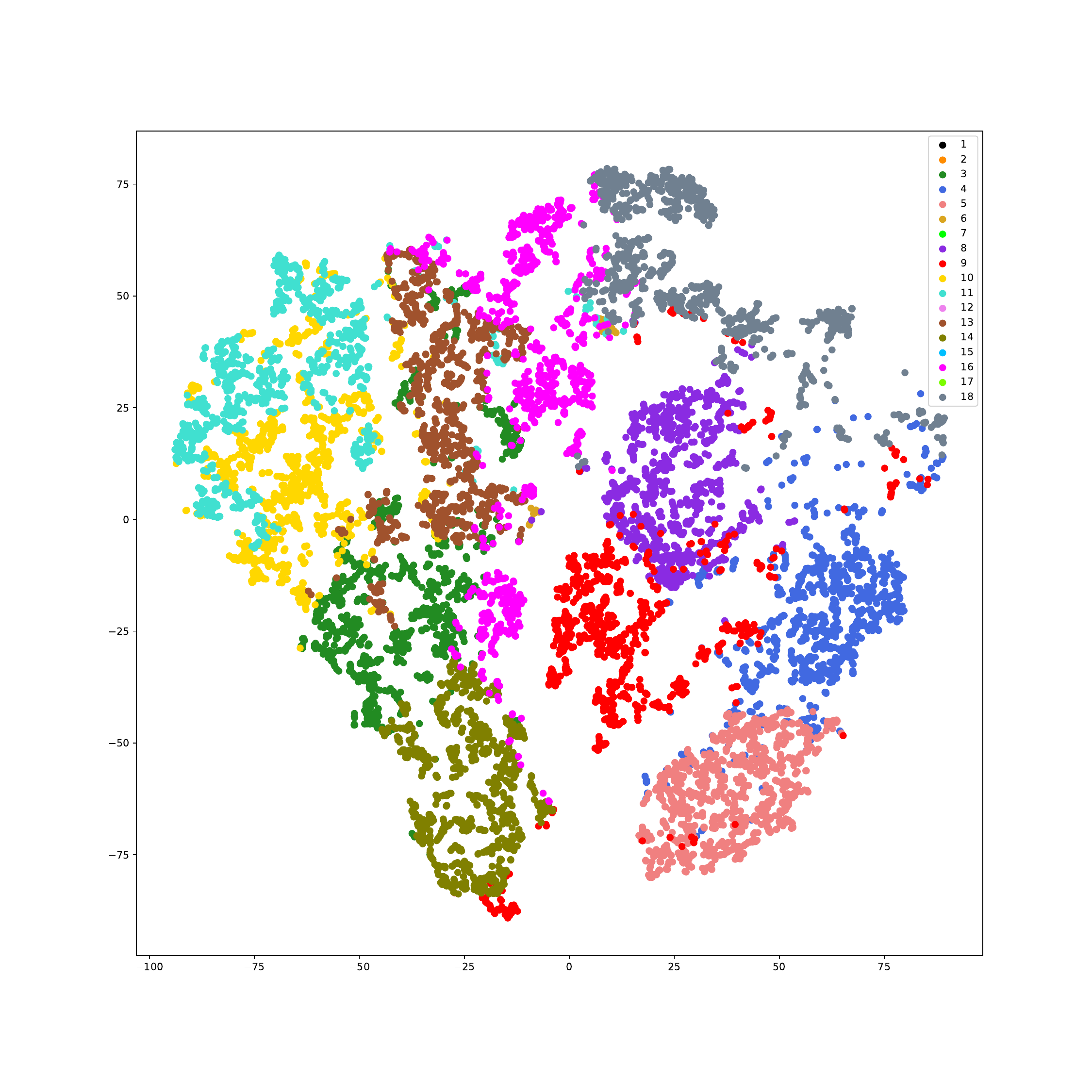}%trim - left, bottom, right, top
\label{fig:tsne_d}}
\caption{Visualization of t-SNE. (a) HyperPointFormer. (b) DGCNN. (c) Point Transformer (d) Point TransformerV3} 
\label{fig:tsne}
\end{figure}

As shown in Fig. \ref{fig:tsne}, the t-SNE visualization of our method reveals clearer and more distinct clusters compared to DGCNN. This indicates that our method provides better separation and discrimination of the classes in the feature space, reflecting its superior performance in learning meaningful representations from the data. Compared to Point Transformer and Point Transformer V3, all models demonstrated a strong ability to separate geometrically well-defined classes such as trees and buildings, as expected, due to their distinct 3D structural characteristics. However, our method achieved noticeably better separation of classes such as grass, bare earth, and paved parking lot. These classes lack strong geometric features, making them more challenging to distinguish using structure alone. We hypothesize that the spectral branch in our model effectively learns class-specific semantics from hyperspectral data, resulting in improved discrimination in cases where geometry is ambiguous.

\subsection{Challenges and Best Practices.} 
Aligned with the ethical considerations and responsible practices outlined in \cite{responsible_ai}, we acknowledge several limitations in our study.

One primary limitation is the inherent mismatch during spectral projection from 2D images to 3D point clouds. In remote sensing, images typically capture objects from above, while lidar data provides a 3D representation of the environment. For instance, lidar can capture areas below the canopy in the vegetated regions, while images only represent the top layer of the canopy. As a result, 3D points below the canopy do not have direct corresponding pixels in the 2D images. To address this limitation, one possible solution is to estimate the spectral values by interpolating from nearby pixels or within the same semantic class. Additionally, contextual information, such as the likelihood of finding terrain beneath a canopy, can help improve the accuracy of the fusion process.

Another limitation arises from potential geometric mismatches between lidar and image data, which can impact the performance of models. While we assume that the benchmark datasets have been carefully processed and georeferenced to minimize these issues, real-world data may experience mismatches due to several factors \cite{10466310}. To address this, geometric mismatches should be corrected prior to fusion to improve model accuracy. Various methods have been proposed to tackle this issue \cite{10466310,CHEN2014659,An:20,9090401,10655611}.

% \textcolor{red}{We also acknowledge the limitation of relying solely on local attention. While the use of downsampling layers helps achieve broader spatial coverage, it is worth considering a hybrid approach that combines both local and global attention. Such an approach could leverage the strengths of local attention for capturing fine-grained, per-point details while utilizing global attention to capture overarching relationships across the entire point cloud. This hybrid strategy may provide a more balanced trade-off between computational efficiency and contextual understanding, making it a promising direction for future exploration.}

The reliance on local attention mechanisms also presents a limitation, as it may restrict the model's ability to capture broader relationships across the entire point cloud. Combining local attention with global attention could offer a better balance between capturing fine-grained details and understanding overall structure of the point cloud. This hybrid approach presents a promising direction for future research.

While the fusion of geometric and spectral features forms the core contribution of our work, we acknowledge that both the DFC2018 and Vaihingen3D datasets have significant data imbalance. We observe that our method underperforms compared to baselines on classes with few samples such as water and crosswalks (DFC2018), and powerline (Vaihingen3D), despite using class weighting during training. This limitation highlights that explicitly handling class imbalance is an important area for future research but falls outside the current scope. More targeted strategies, such as class-specific data augmentation, may be necessary to improve performance on rare classes.

% \textcolor{red}{Future research should aim to address the challenges encountered during the development of our method, particularly by improving 2D-to-3D spectral projection techniques to create more accurate multimodal 3D point clouds. Additionally, models should concentrate on simultaneously resolving geometry mismatches across different modalities while learning semantic features. Lastly, identifying an efficient way to combine local and global attention mechanisms in a 3D context could substantially enhance performance.}

In conclusion, addressing these limitations, particularly 2D-to-3D spectral projection techniques, improving geometric mismatch correction, and exploring hybrid attention mechanisms, will help enhance the robustness, accuracy, and applicability of the method in real-world scenarios.

\section{Conclusion}
\label{sec:con}
In this work, we addressed the limitations of traditional 2D methods for integrating multimodal remote sensing data by proposing a 3D fusion approach. We developed a dual-attention neural network that simultaneously learns geometric and spectral features from hyperspectral and lidar data. By leveraging a cross-attention-based fusion mechanism, we effectively combined different modalities at multiple scales, allowing each modality to inform the relevance of the other.

Our evaluation, conducted using the IEEE 2018 Data Fusion Contest (DFC2018) dataset, demonstrated that our 3D fusion method achieves competitive results compared to existing 2D approaches. Moreover, our method offers enhanced flexibility by providing 3D predictions that can be projected onto 2D maps, a capability that reverse approaches do not possess.

We explored various self-attention mechanisms and fusion architectures, including scalar attention, vector attention, and different fusion strategies. Our ablation studies provided valuable insights into the effectiveness of these different fusion strategies.

To further advance research in this area, we will release our multimodal 3D point clouds, an extended version of the DFC2018 dataset, to the public. We believe this will encourage the community to explore data fusion in a 3D context. Our code will also be made available to support replication and further development of our work.

In conclusion, our 3D fusion approach not only enhances classification performance but also opens up new possibilities for 3D predictions in remote sensing applications, paving the way for more advanced and accurate urban scene analysis.

\section*{Acknowledgment}
This work was supported by the European Regional Development Fund and the Land of Saxony by providing the high specification Nvidia A100 GPU server that we used in our experiments. We would like to thank NCALM at the University of Houston and IADF TC for providing the data, through the 2018 Data Fusion Contest.

\ifCLASSOPTIONcaptionsoff
  \newpage
\fi

\bibliographystyle{IEEEtran}
\bibliography{references}

\begin{IEEEbiography}[{\includegraphics[width=1in,height=1.25in,clip,keepaspectratio]{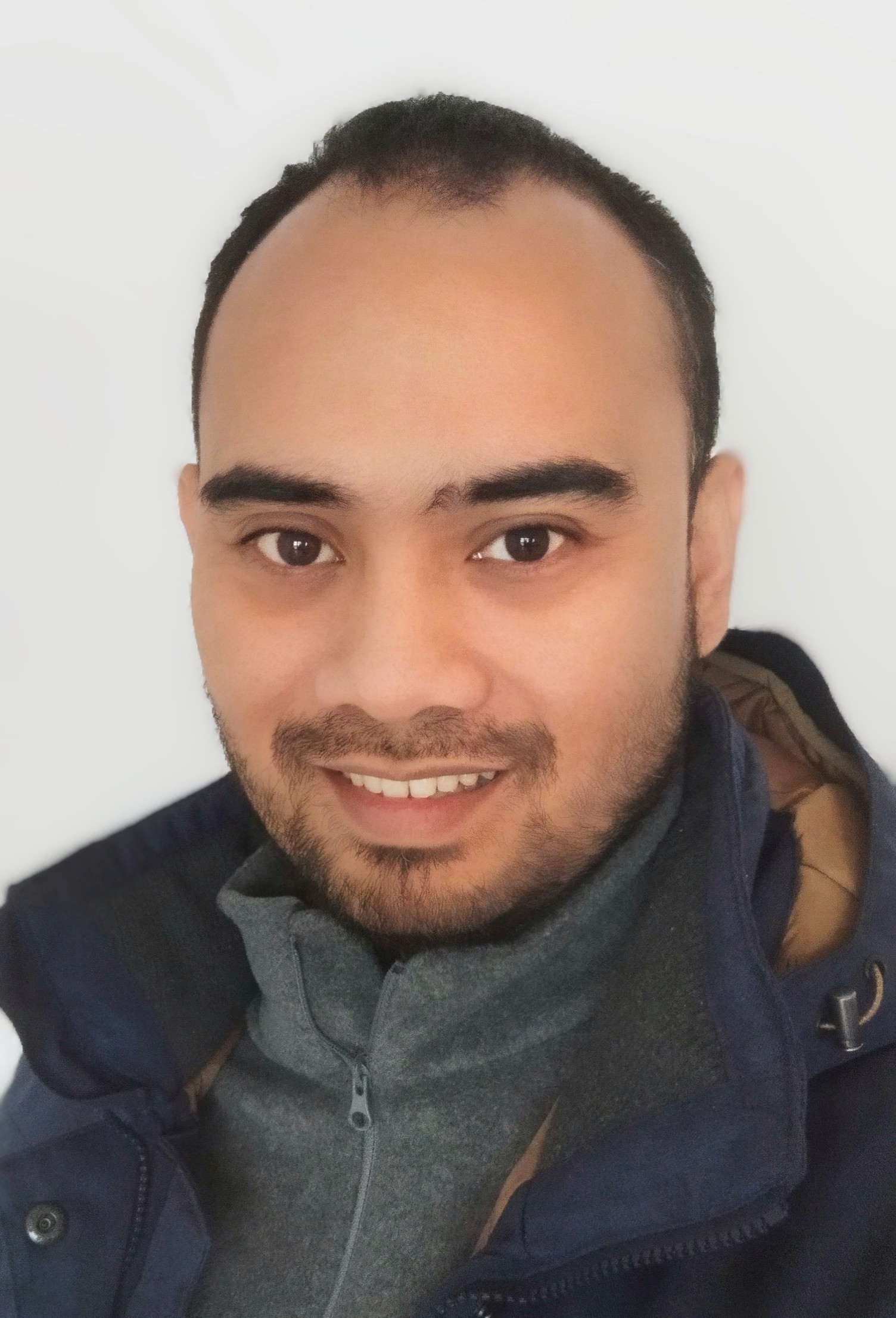}}]{Aldino Rizaldy} received his M.Sc. degree in Geoinformation from Faculty ITC, University of Twente, The Netherlands. Currently, he is a Ph.D. student at the Helmholtz-Institute Freiberg for Resource Technology, Helmholtz-Zentrum Dresden-Rossendorf and Remote Sensing and Geoinformatics, Freie Universität Berlin. His main research interests are deep learning, computer vision, 3D point cloud, lidar and photogrammetry.
\end{IEEEbiography}

\begin{IEEEbiography}[{\includegraphics[width=1in,height=1.25in,clip,keepaspectratio]{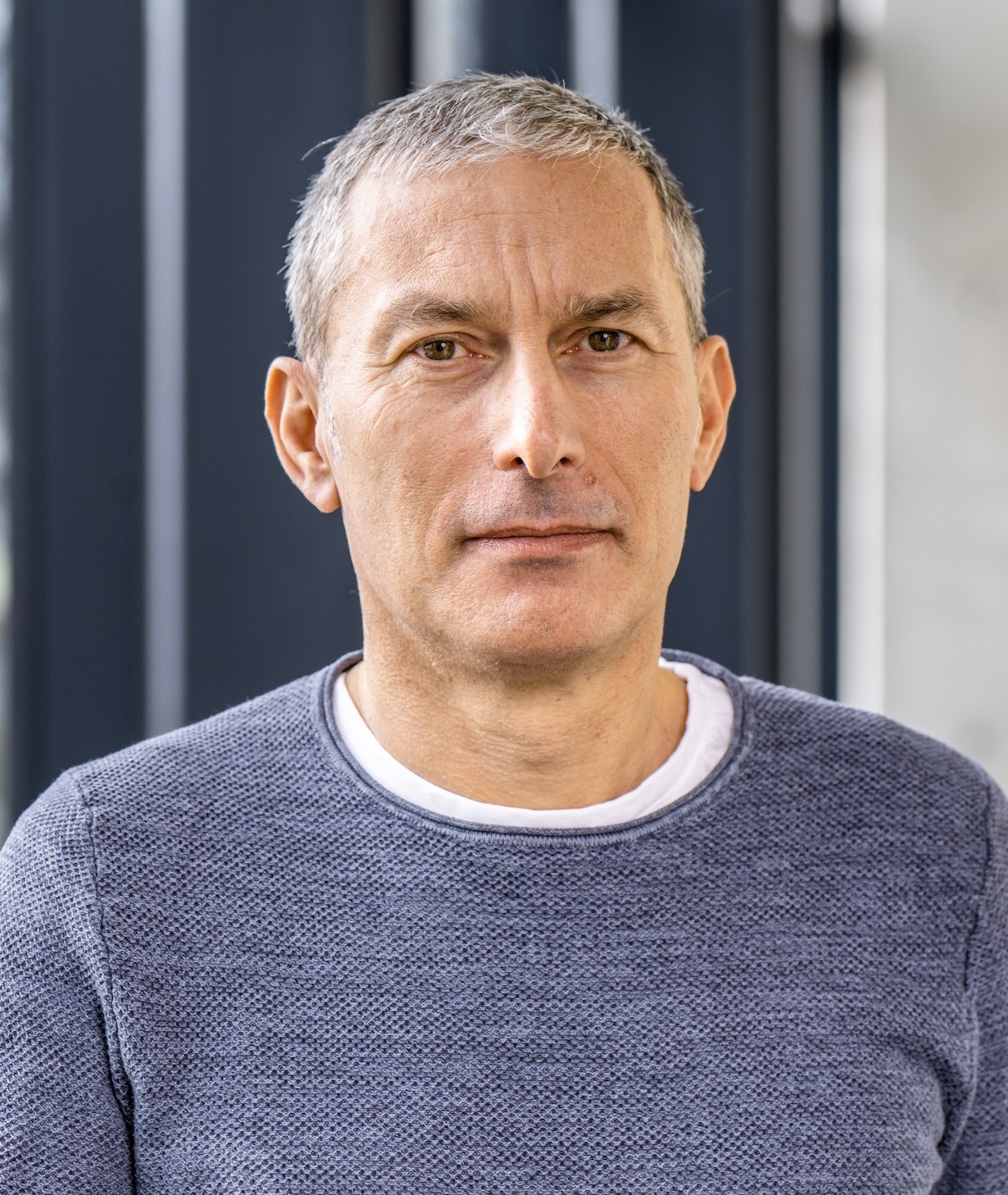}}]{Richard Gloaguen}
Head of the Exploration Department at the Helmholtz-Institute Freiberg for Resource Technology. Richard Gloaguen received the Ph.D. degree “Doctor Communitatis Europae” in marine geosciences from the University of Western Brittany, Brest, France, in collaboration with the Royal Holloway University of London, U.K., and Göttingen University, Germany, in 2000. He was a Marie Curie Post-Doctoral Research Associate at the Royal Holloway University of London from 2000 to 2003. He led the Remote Sensing Group at University Bergakademie Freiberg, Freiberg, Germany, from 2003 to 2013. Since 2013, he has been leading the division “Exploration Technology” at the Helmholtz-Institute Freiberg for Resource Technology, Freiberg. He is currently involved in UAV-based multisource imaging, laser-induced fluorescence, and non-invasive exploration. His research interests focus on multisource and multiscale remote sensing integration using computer vision and machine learning.
\end{IEEEbiography}%

\begin{IEEEbiography}
[{\includegraphics[width=1in,height=1.25in,clip,keepaspectratio]{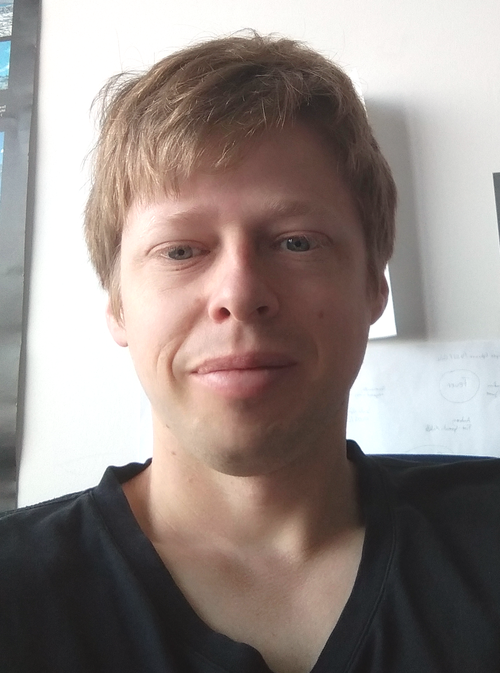}}]
    {Fabian Ewald Fassnacht} is trained as forester and has a PhD in Remote Sensing of Forests from the University of Freiburg, Germany. He has been working as PostDoc and Lecturer at the Karlsruhe Institute of Technology in the time between 2014 and 2022. Currently he is a full professor for Remote Sensing and Geoinformatics at Freie Universitaet Berlin, Germany. His research focuses on the application of passive and active optical remote sensing data for environmental application with a particular focus on vegetation and forests. 
\end{IEEEbiography}

\begin{IEEEbiography}[{\includegraphics[width=1in,height=1.25in,clip,keepaspectratio]{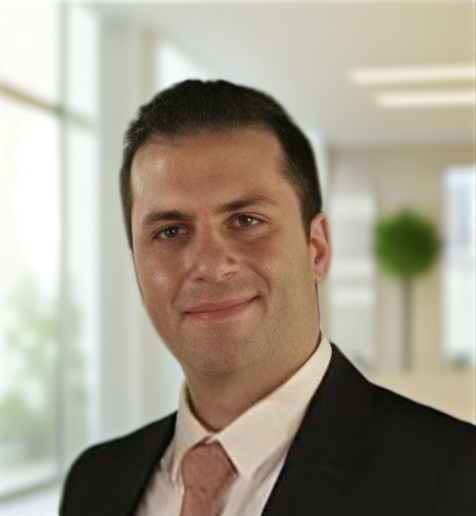}}]{Pedram Ghamisi} (Senior Member, IEEE) earned his Ph.D. in Electrical and Computer Engineering from the University of Iceland in 2015. He currently holds two positions: (1) Head of the Machine Learning Group at Helmholtz-Zentrum Dresden-Rossendorf (HZDR), Germany, and (2) Visiting Full Professor at Lancaster University, UK. Previously, he was Senior Principal Investigator, Research Professor, and Group Leader of AI4RS at the Institute of Advanced Research in Artificial Intelligence (IARAI), Austria. Additionally, he co-founded and served as the CTO of VasogNosis, a startup company based in the U.S. with branches in Milwaukee and California.

Prof. Ghamisi has received over 10 distinguished international awards and recognitions, including the IEEE-GRSS Highest Impact Paper Award (2020 and 2024), the IEEE GRSS Data Fusion Contest Winner (2017), the IEEE Mikio Takagi Prize (2013), and the IEEE GRSS Best Reviewer Prize (2017). Since 2021, he has been consistently ranked among the top 1\% of most-cited researchers by Clarivate. His research focuses primarily on deep learning for remote sensing applications, with particular emphasis on Open Science, AI for Good, Responsible AI, and AI Security, playing a key role in advancing the AI4EO era. Dr. Ghamisi has also been awarded multiple prestigious scholarships and grants, including the Helmholtz Foundation Models Initiatives (2024), the High Potential Program (HPP) Team Leadership (2018), and the Alexander von Humboldt Fellowship (2015). For more information, please visit www.ai4rs.com.
\end{IEEEbiography}

\end{document}